\documentclass[conference]{IEEEtran}

\usepackage{dcolumn}
\usepackage{times}
\usepackage{float}
\usepackage{stfloats}
\usepackage{tabularx,booktabs}
\newcolumntype{Y}{>{\centering\arraybackslash}X}
\usepackage{siunitx}
\usepackage[table]{xcolor}
\usepackage{tikz}
\usepackage{bbm}
\newcolumntype{e}[1]{D{+}{\,\pm\,}{#1}} 

\usepackage{relsize}    
\usepackage{titletoc}

\usepackage{enumitem}
\setlist[enumerate]{leftmargin=*}
\setlist[itemize]{leftmargin=*}

\newcommand{\bl}[1]{{\flushleft\textbf{#1}}}

\newcommand{\simtoreal}{sim-to-real}
\newcommand{\SimtoReal}{Sim-to-Real}

\newcommand{\name}{SLIM}
\newcommand{\code}{\texttt{{https://github.com/placeholder\_url}}}
\newcommand{\demo}{supplementary file}

\newcommand{\Search}{{\ttfamily Search}}
\newcommand{\MoveTo}{{\ttfamily MoveTo}}

\newcommand{\Grasp}{{\ttfamily Grasp}}
\newcommand{\SearchWithObj}{{\ttfamily SearchWithObj}}
\newcommand{\MoveToWithObj}{{\ttfamily MoveToWithObj}}
\newcommand{\MoveGripperToWithObj}{{\ttfamily MoveGripperToWithObj}}
\newcommand{\DropInto}{{\ttfamily DropInto}}
\newcommand{\Idle}{{\ttfamily Idle}}

\newcommand{\NoArmRetract}{{\textbf{No Arm Retract}}}
\newcommand{\NoPerturb}{{\textbf{No Perturbation}}}
\newcommand{\NoVisualAug}{{\textbf{No Visual Aug}}}
\newcommand{\DistillationOnly}{{\textbf{Distillation Only}}}
\newcommand{\HumanTeleop}{{\textbf{Human Teleop}}}
\newcommand{\NoDistillation}{{\textbf{No Distillation}}}

\newcommand{\bkcolor}[1]{\mathbf{#1}}

\newcommand{\tikzxmark}{%
\tikz[scale=0.18] {
    \draw[line width=0.7,line cap=round] (0,0) to [bend left=6] (1,1);
    \draw[line width=0.7,line cap=round] (0.2,0.95) to [bend right=3] (0.8,0.05);
}}
\newcommand{\tikzcmark}{%
\tikz[scale=0.18] {
    \draw[line width=0.7,line cap=round] (0.25,0) to [bend left=10] (1,1);
    \draw[line width=0.8,line cap=round] (0,0.35) to [bend right=1] (0.23,0);
}}

\newcommand*\circled[1]{\raisebox{.5pt}{\textcircled{\raisebox{-.9pt} {#1}}}}
	
\definecolor{NavyBlue}{rgb}{0.0, 0.0, 0.5}

\newcommand{\piagent}{\pi_{\rm stu}}
\newcommand{\piexpert}{\pi_{\rm tea}}

\newcommand{\Figure}{Fig.}

\newcommand{\instr}[1]{\textit{\textbf{#1}}}

\usepackage[numbers]{natbib}
\usepackage{multicol}
\usepackage{caption} 
\usepackage[bookmarks=true]{hyperref}

\usepackage{microtype}
\usepackage{graphicx}
\usepackage{subfigure}
\usepackage{booktabs}
\usepackage[utf8]{inputenc}
\usepackage{natbib}
\usepackage{hyperref}

\usepackage{tabularx}
\usepackage{algorithm}
\usepackage{algorithmic}
\usepackage{amsmath}
\usepackage{amssymb}
\usepackage{mathtools}
\usepackage{amsthm}
\usepackage{multirow}
\usepackage{multicol}
\usepackage{overpic}
\usepackage[capitalize,noabbrev]{cleveref}

\theoremstyle{plain}

\theoremstyle{definition}

\theoremstyle{remark}

\usepackage[textsize=tiny]{todonotes}

\usepackage{pifont}

\begin{document}

\title{\name: \SimtoReal{} Legged Instructive Manipulation \\ via
Long-Horizon Visuomotor Learning}

\author{\authorblockN{Haichao Zhang, \hspace{0.1mm}  Haonan Yu, \hspace{0.1mm} Le Zhao, \hspace{0.1mm}  Andrew Choi, \hspace{0.1mm} Qinxun Bai, \hspace{0.1mm} Break Yang, \hspace{0.1mm}   Wei Xu}
	\authorblockA{Horizon Robotics\\
		\texttt{\{first\_name.last\_name\}@horizon.auto}
	}
}

\twocolumn[{%
\renewcommand\twocolumn[1][]{#1}%
\maketitle
\includegraphics[width=\textwidth]{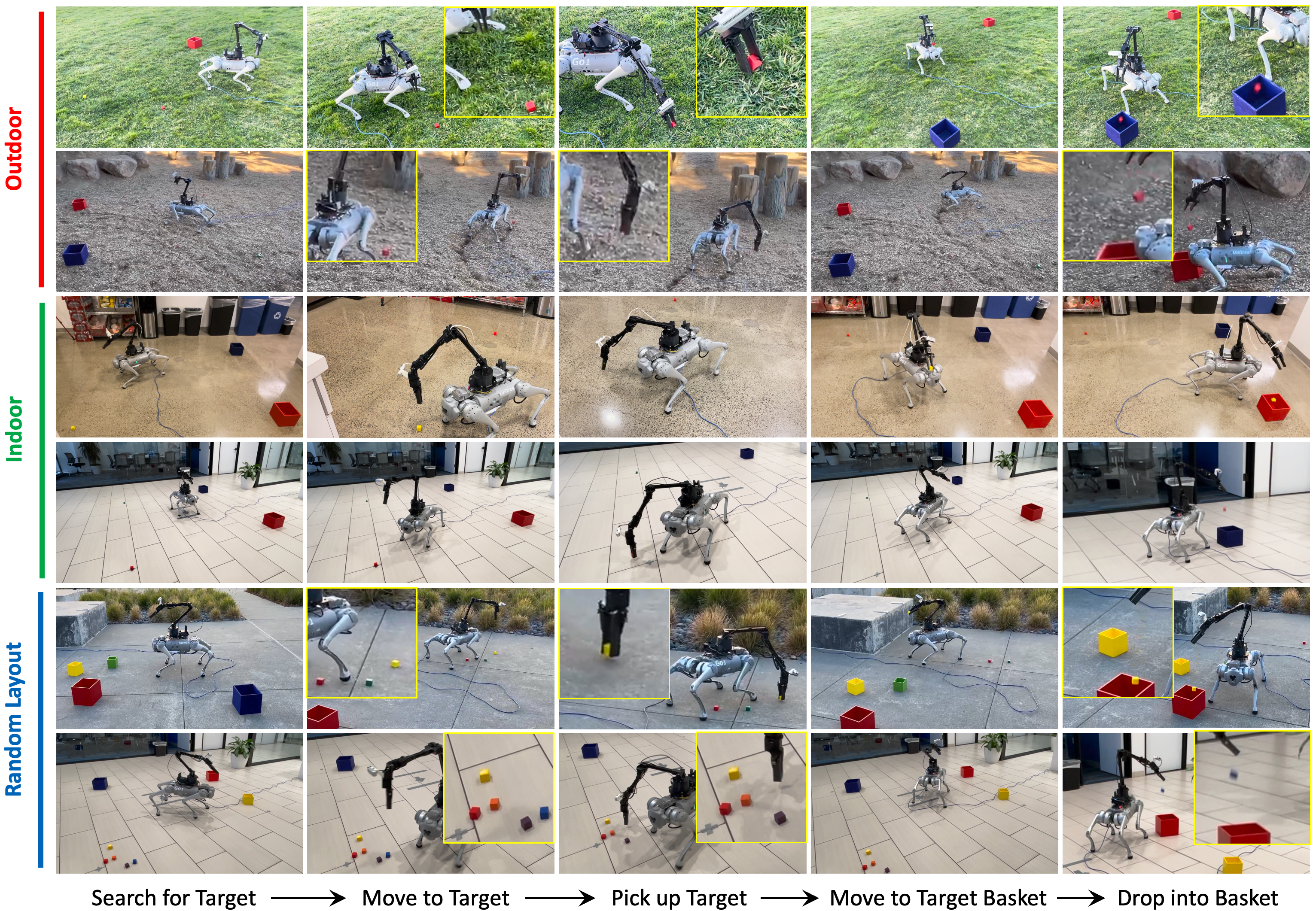}
\captionof{figure}{\textbf{\name{} in Real.} Snapshots of the \emph{same} \name{} policy deployed in diverse real-world scenes, featuring significant variations in terrain, background, distractors, and other environmental factors.
These scenes are not available in simulation training.
The subtask annotations at the bottom are added for understanding task progress and are not part of the input to the system.
}
\vspace{0.1in}
\label{fig:slim_in_real}
}]

\begin{abstract}
We present a low-cost legged mobile manipulation system that solves long-horizon real-world tasks, trained by reinforcement learning purely in simulation.
This system is made possible by 1)~a hierarchical design of a high-level policy for visual-mobile manipulation following task instructions, and a low-level quadruped locomotion policy, 2)~a teacher and student training pipeline for the high level, which trains a teacher to tackle long-horizon tasks using privileged task decomposition and target object information, and further trains a student for visual-mobile manipulation via RL guided by the teacher's behavior, and 3)~a suite of techniques for minimizing the \simtoreal{} gap.
\newline\hspace*{1em}In contrast to many previous works that use high-end equipments, our system demonstrates effective performance with more accessible hardware -- specifically, a Unitree Go1 quadruped, a WidowX-250S arm, and a single wrist-mounted RGB camera -- despite the increased challenges of \simtoreal{} transfer.
Trained fully in simulation, a single policy autonomously solves long-horizon tasks involving search, move to, grasp, transport, and drop into, achieving nearly 80\% real-world success.
This performance is comparable to that of expert human teleoperation on the same tasks while the robot is more efficient, operating at about 1.5$\times$ the speed of the teleoperation.
Finally, we perform extensive ablations on key techniques for efficient RL training and effective \simtoreal{} transfer, and demonstrate effective deployment across diverse indoor and outdoor scenes under various lighting conditions.
\end{abstract}

\IEEEpeerreviewmaketitle

\section{Introduction}

Legged mobile manipulation combines a robotic manipulator with a legged mobile platform, enabling robots to perform a wide variety of complex tasks in diverse environments~\cite{deep_wholebody,ASC, pan2024roboduet,VBC,GEFF, GAMMA}. Unlike stationary or wheeled manipulators, legged systems can adapt to uneven terrains, significantly expanding the scope of potential applications. Recent advancements in hardware and algorithms~\cite{deep_wholebody, walk_in_minutes, walk_these_ways} have also made legged manipulators more accessible and cost-effective, encouraging their adoption in research and industry.


Despite its promise, legged mobile manipulation presents unique challenges.
The expanded scope of the tasks, the long task horizon, the diverse scenes and terrains, and unstable legged base all compound the difficulty of achieving reliable performance.

One popular approach to tackle mobile manipulation is Imitation Learning (IL) from expert demonstrations~\citep{rt1, mobilealoha, wu2024tidybot, pi0}.
It is straightforward, leveraging human expertise to produce reasonable behaviors quickly with supervised training. However, IL relies heavily on large and comprehensive datasets to ensure generalization and robustness, making data collection resource-intensive, both in terms of human labor and hardware wear-and-tear. Moreover, its performance is upper-bounded by the quality of human demonstrations.
These limitations become even more pronounced when addressing the complexity of legged manipulation.


Reinforcement learning (RL) in simulation with \simtoreal{} transfer is another paradigm that avoids some of these limitations. 
Simulation provides virtually infinite data, ensuring robust policy training through diverse and comprehensive data generation ~\citep{tobin2017,imai2022,james2019sim,walk_in_minutes,VBC,playfuldoggybot}, and allows safe exploration without physical wear-and-tear.  In addition, RL, unlike IL, is not constrained by the quality of demonstrations.

However, \simtoreal{} RL
presents its own set of challenges: (1) effective RL training for long-horizon tasks is non-trivial, and (2) the \simtoreal{} gap -- the discrepancies between simulated and real-world environments in both 
dynamics and vision -- can significantly degrade policy performance in deployment.
These challenges are further exacerbated by the added complexities of legged manipulation.


To address these challenges, we present \name{} (\textbf{S}im-to-Real \textbf{L}egged \textbf{I}nstructive \textbf{M}anipulation), a standalone system for training robotic policies entirely in simulation and deploying them zero-shot in the real world. \name{} employs a hierarchical policy structure that separates high-level visual mobile manipulation following task instructions from low-level quadruped locomotion control (Section~\ref{sec:low-high}).
In high-level policy training, we utilize a teacher-student learning framework to enhance training efficiency, where a teacher policy leverages privileged information to guide the student.
We train the teacher policy
to solve long-horizon tasks via task decomposition and progressive learning (Section~\ref{sec:teacher_learning}).
We then train a student policy, conditioned on language and visual inputs, with RL guided by the teacher's behavior (Section~\ref{sec:student}).
Together with a suite of carefully designed \simtoreal{} techniques (Section~\ref{sec:sim2real_gap_redu}), including visual and dynamics randomization and low-level controller tuning, \name{} produces visuomotor policies that can maintain sim performance when transferred to real.

In this work, we implement \name{} on a Unitree Go1 quadruped with a top-mounted WidowX-250S manipulator and a wrist-mounted Intel RealSense D435 camera (only using the RGB stream), resulting in a large 19 degree-of-freedom (DOF) system.
We focus on the long-horizon task consisting of multi-stage search, move to, grasp, transport, and drop into, achieving nearly 80\% real-world success (Section~\ref{sec:result}).

Our key contributions are as follows:
\begin{enumerate}
    \item We develop a low-cost legged manipulation system \name{}.
    To the best of our knowledge, \name{} is the first end-to-end robotic system for solving long-horizon real-world legged manipulation tasks from \simtoreal{} RL alone.
    %
    %
    \item We conduct an extensive set of real-world experiments with 400 episodes across various indoor and outdoor scenes. Our model achieves $\sim$80\% success, comparable to that of expert human teleoperation, but significantly more efficient, operating at about 1.5$\times$ the speed.
    \item We identify crucial techniques for achieving successful \simtoreal{} transfer with inexpensive hardware, validated through extensive real-world ablations.
\end{enumerate}

{\flushleft We plan to open source the code to facilitate future efforts on related research.}
\section{Related Work}
\subsection{Learning-based Quadruped Locomotion}
Traditionally, quadruped locomotion has been tackled through classical control methods designed to follow hand-tuned gaits on flat ground~\citep{control_dynamic_gaits, dynamic_quad_locomotion}, dynamic rigid platforms~\citep{stable_quad_control}, discrete terrain~\citep{agrawal2022vision}, and rough terrain~\citep{robust_quad_walking}.
Though impressive, such methods require significant human engineering efforts and can be brittle to environmental changes.

More recently, there has been explosive progress in using learning-based approaches for achieving quadruped locomotion.
In particular, \simtoreal{} reinforcement learning (RL) has arisen as a robust solution showcasing impressive feats such as parkour~\citep{robot_parkour, extreme_parkour, caluwaerts2023barkour} and high speed running~\citep{margolisyang2022rapid}.
Other works have also focused on exploiting the strengths of legged locomotion over their wheeled counterparts through utilizing the strong inherent coupling of proprioception with egocentric vision~\citep{fu2021coupling, agarwal2022ego} and active estimation of the environment~\citep{margolis2023active, kumar2021rma, HIM}.
Furthermore, 
necessary tuning can be minimal
compared to classical approaches~\citep{fu2021minimizing}. 
For instance,
massively parallelized simulation has been 
shown capable of learning 
gaits in just minutes~\citep{walk_in_minutes}.
Finally, data-driven legged locomotion has shown great promise in its ability to zero-shot generalize to new morphologies~\citep{feng2022genloco}, surfaces~\citep{walk_these_ways, kumar2021rma, HIM} and agile skills~\citep{yang2023generalized, ji2023dribble}.

\subsection{Legged Mobile Manipulation}

Significant advancements in robust locomotion have further enabled researchers to push the boundaries of legged mobile manipulation. 
Building upon data-driven approaches, quadrupeds have been demonstrated pushing objects with their body~\citep{wholebody},  
dribbling
balls around~\citep{ji2023dribble}, and 
manipulating
objects using egocentric~\citep{wu2024helpfuldoggybot} or calf mounted grippers~\citep{locoman}.
One of the the most popular setups has been the traditional top-mounted 6DOF manipulator design~\citep{deep_wholebody, ha2024umionlegs, VBC, GAMMA, ASC, ma2022mpc, pan2024roboduet, sleiman2023wholebody_control}.
In addition to significant workspace expansion, manipulators offer other benefits, such as assisting with balance~\citep{huang2024manipulatortail} and serving as an intuitive interface for collecting human demonstrations~\citep{ha2024umionlegs}.
Classical approaches to solving top-mounted legged mobile manipulation typically design wholebody controllers~\citep{sleiman2023wholebody_control} or combine locomotion learning with model-based manipulation control~\citep{ma2022mpc}.
Others simply call high-level APIs provided by the quadruped manufacturers to achieve graspability-aware policies~\citep{GAMMA}, navigational pick-and-place~\citep{ASC}, and language-conditioned mobile manipulation~\citep{GEFF}.
Complex tasks have also been achieved by leveraging teacher-student training setups in simulation for grasping~\citep{VBC} and door opening~\citep{zhang2024learning}.
Finally, to expand the robot's workspace even further, wholebody loco-manipulation has arisen where leg joints are actuated in a way to assist manipulation~\citep{deep_wholebody, VBC, pan2024roboduet}.

Most 
closely related
to our work, \citet{VBC} introduce a visual whole-body
control approach (VBC), which trains a \simtoreal{} visuomotor whole-body loco-manipulation policy using a high-low hierarchical model and a teacher-student training setup for the object pickup task, similar to our design.
The system can grasp an impressive set of diverse objects by leveraging whole-body manipulation with capable hardware. 
VBC takes a few shortcuts to simplify the task. 
VBC requires the user to manually click on the target object to perform initial segmentation using a third-party vision model. This segmentation is then used to track the object through a third-party tracking model, which requires the object to stay within the camera view the whole time.
Additionally, VBC uses a scripted policy to put the object into the basket on the back of the quadruped, once the object is lifted, making the task scope more limited and the task horizon much shorter.

In comparison, \name{} offers the follow advantages:
\begin{enumerate}
\item \name{} is a \textbf{complete and self-contained system}. 
All modules--vision, high-level visual policy, low-level motor control--are trained in our simulation environment integrated within a single framework without any third party modules, resulting in a lean 19 million parameter neural network system, allowing both minimal latency and full autonomy.
\item \name{} is \textbf{fluid and intuitive to use}. Given a language instruction, \name{} is able to rapidly accomplish tasks in a wide variety of scenes.
\item Finally, \name{} is 
\textbf{practical}.
\name{} can accomplish complex long-horizon real-world tasks autonomously from the beginning to the end, as apposed to some of the existing methods that require human involvement (\emph{e.g.} manually clicking for segmentation) and can only automate part of the task with restrictions on the feasible starting state distribution (\emph{e.g.} requiring full visibility of the target object from the beginning)~\cite{VBC}.
\end{enumerate}

\section{Architecture Overview}
\label{sec:architecture}

\begin{figure*}[t]
    \centering
    \begin{overpic}[width=\textwidth]{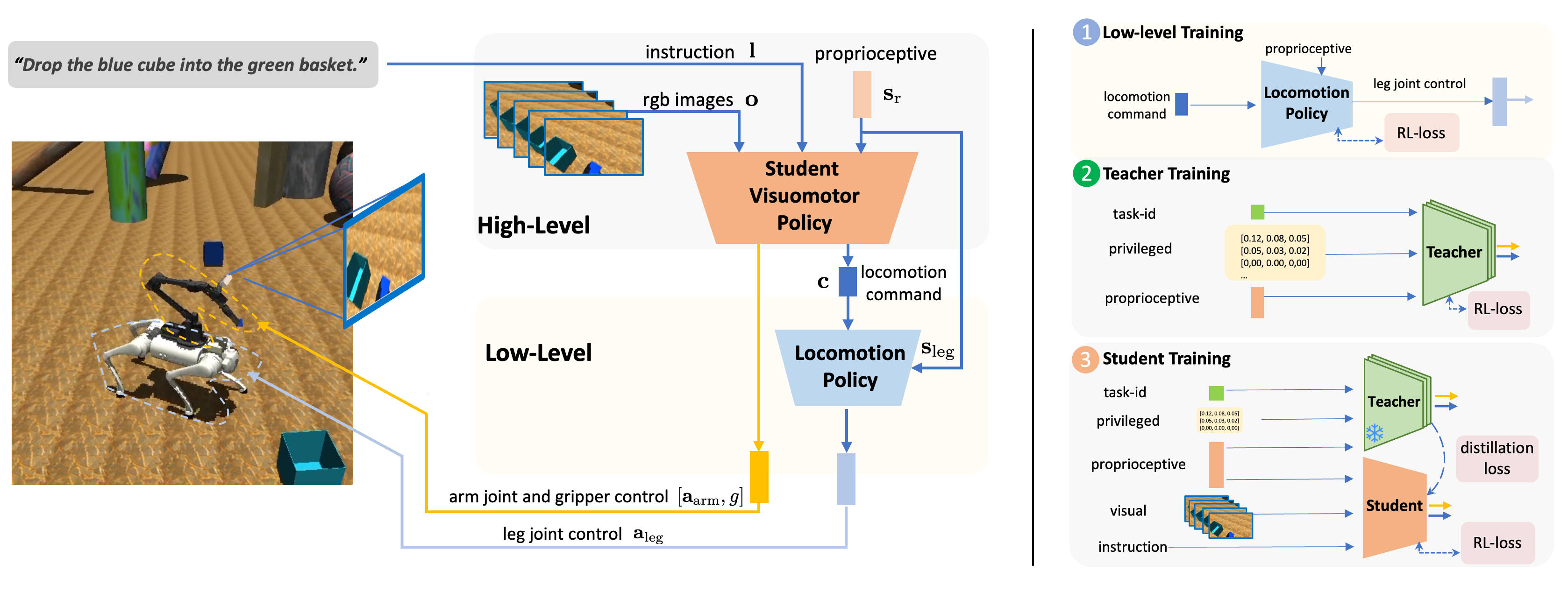}
    \end{overpic}
    \vspace{-0.2in}
    \caption{\textbf{Hierarchical Framework and Visuomotor Policy Pipeline (Left).} Given language instruction and sensor inputs, the high-level policy generates a set of two control signals: 1) the arm and gripper control signals, and 2) the locomotion command. The arm control signals are directly passed to the arm driver, and the locomotion command is passed to the low-level policy to control the leg joints of the quadruped. 
    \textbf{Low-Level Training (Top Right), and High-Level Teacher and Student Training (Middle and Bottom Right).} \name{} training is divided into three sequential stages. First, the low-level locomotion policy is trained via RL to follow a sampled linear and angular velocity command. 
    Second, the high-level teacher policy is trained via RL with privileged low-dimensional state input to solve the long-horizon task. 
    Finally, the student visuomotor policy is trained by distilling the teacher behavior while maximizing task rewards, using visual, sensory, and language instruction as input. 
    Both the teacher and student policies command the same frozen low-level policy produced by the first stage.  The teacher is only run in simulation and the student can be deployed in real.
    }
    \label{fig:framework}
    \vspace{-0.1in}
\end{figure*}

\Figure~\ref{fig:framework} illustrates the hierarchical (high- and low-level policy) framework (left) and for the high level, the teacher-student training pipeline (right) used in \name{}.
Given language instruction and sensor inputs, the high-level policy generates an intermediate control command that will be used as an input to the low-level policy.
A low-level locomotion policy is trained to track the specified locomotion command via RL and is frozen after training. 
Then a high-level teacher policy is trained with privileged information via RL to solve the long-horizon task, and a student visuomotor policy is trained by distilling the teacher policy while maximizing task rewards, conditioned on visual inputs and language instruction.

\subsection{Observations and Actions}
\label{sec:obs_and_actions}
\bl{Observations.} The robot observation at each time step consists of three components:
\begin{itemize}[align=parleft, leftmargin=*, labelsep=2.5em]
  \item[$\mathbf l$] A tokenized instruction vector of length $L$.
  \item[$\mathbf o$] A temporal stack of RGB images from a single wrist-mounted camera with a shape of $(N,H,W,3)$, where $N$ is the stack size, and $H$ and $W$ are the image height and width, respectively.
    The environment is always partially observed by the robot, and most of the workspace will not be contained in this camera view during the task.
  \item[$\mathbf s_\textrm{r}$] A temporal stack of proprioceptive state vectors of the robot with a shape of $(N,D)$, where $D$ is the dimensionality of the concatenation of proprioceptive readings of all joints. 
  $\mathbf s_\textrm{r} \!=\! [\mathbf s_\textrm{leg}, \mathbf s_\textrm{arm}]$, where $\mathbf s_\textrm{leg}$ denotes the leg-related proprioceptive state (leg joint positions) and $s_\textrm{arm}$ the arm proprioceptive state (arm joint positions).
\end{itemize}

\bl{Privileged Observations (only available in simulation).} To achieve better learning efficiency, we first train the teacher policy using privileged observations, and then distill it to the student policy conditioned on the standard observations from the robot. The teacher policy has the subtask id $k$ as input in place of the language instruction $\mathbf{l}$, and lower-dimensional object state observation $\mathbf s_\textrm{p}$ in place of the visual input $\mathbf{o}$:
\begin{itemize}[align=parleft, leftmargin=*, labelsep=2.5em]
    \item[$k$] An integer in $[1,K]$ indicating which subtask the teacher is currently solving. $K$ is the maximum number of subtasks allowed by the system.
    \item[$\mathbf s_\textrm{p}$] A group of temporal stacks of privileged object features (\textit{e.g.}, positions, orientations, scales, categories, \textit{etc.}), where each stack has a shape of $(N,Q_m)$, with $0\le m < M$ denoting the object index, and $M$ the maximum number of task related objects in the scene. Note that object features are also partially observed and their visibility is always determined by the robot's camera field of view. The features are set to zero if the object is not in the view.
    This ensures that object visibility is consistent between the teacher policy and the student policy, and avoids any information gap during policy distillation.
\end{itemize}
The privileged information can be extracted or computed from the underlying simulator state.

\bl{Actions.} The robot policy outputs a triplet of actions:
\begin{itemize}[align=parleft, leftmargin=*, labelsep=2.5em]    
    \item[$\mathbf a_\textrm{arm}$] A vector in $[-z,z]^I$, indicating the delta changes to the arm joint positions. $z$ is the largest joint position change allowed for a control interval.
    $I$ is the number of actuated arm joints, excluding the gripper joint.
    \item[$g$] A target gripper position in $[0,1]$.
    \item[$\mathbf a_\textrm{leg}$] The target quadruped joint positions in $\mathbb R^{J}$.
    $J$ is the number of actuated leg joints.
\end{itemize}
In total, a complete action $\mathbf{a}$ has $(I+1+J)$ dimensions.

\subsection{Hierarchical Policy Structure}
\label{sec:low-high}
The overall system takes visual and instructional input and controls all joints of the quadruped and the arm, which involves a non-trivial training task with high-dimensional input and output.  
We  use a two-level hierarchical policy to divide the complexity of visual mobile manipulation training from those of legged locomotion training, illustrated on the left of \Figure~\ref{fig:framework},
similar to \cite{VBC}.

The high-level policy takes visual, instructional, and proprioceptive inputs, and outputs a locomotion speed command and arm control command. The low level policy tracks the locomotion command from the high level, using proprioceptive inputs to command the quadruped's leg joints.
The assumption is that given the locomotion command, quadruped locomotion control is largely independent of the high-level task semantics.
\bl{Intermediate Action.}
The following intermediate action separates the high and low-level policies,
\begin{itemize}[align=parleft, leftmargin=*, labelsep=2.5em]    
    \item[$\mathbf c$] A vector in $\mathbb{R}^2$, containing the target forward and angular velocities for the quadruped.
\end{itemize}
With this, a high-level policy is defined as 
\[(\mathbf{l}, \mathbf{o}, \mathbf{s_\textrm{r}}) \rightarrow \mathbf{a}_{\text{hi}}\triangleq(\underbrace{[\mathbf{a_\textrm{arm}},g]}_{\textrm{arm and gripper control}},\underbrace{{\mathbf{\textcolor{white}{[}c\textcolor{white}{]}}}}_{\textrm{locomotion command}}) \in \mathbb{R}^{I+3},\]
while the low-level policy is defined as
\[(\mathbf{c}, \mathbf{s_\textrm{leg}}) \rightarrow \mathbf{a_\textrm{leg}} \in \mathbb{R}^{J}.\]
\noindent Note that with this decomposition, the low-level policy no longer observes task specific inputs $(\mathbf{l},\mathbf{o})$ and thus, does not need to process high-dimensional image and language inputs.

\bl{Low Level.} The low-level policy is a quadruped controller that generates joint position targets for PD control to follow a task-agnostic 2D linear and angular-velocity command $\mathbf{c}$~\cite{walk_these_ways} and is trained in simulation using PPO~\citep{ppo}.
An illustration of the training can be found in the top row of the right column of \Figure~\ref{fig:framework}.
It is worth noting that during low-level training, the same embodiment as shown in \Figure~\ref{fig:framework} (arm mounted on top of the quadruped) is used in low-level training.
We randomly sample both command $\mathbf{c}$ and arm joint actions $\mathbf{a_\textrm{arm}}$ during training. 
Since the arm mounted on the quadruped can have varying poses during the task, this enforces the low-level command following ability to generalize to various arm configurations.
That is, the quadruped base has to learn to keep balance while achieving locomotion commands regardless of the arm's current joint positions and movements.  We intentionally remove arm joint state and joint command from low level input, see details in Appendix~\ref{app:low-level}.
Similar to most \simtoreal{} approaches, we also randomize the simulation environment during training. Besides widely-used domain randomization parameters as in~\cite{deep_wholebody}, we further randomize the simulated delay of each sensor and foot softness of the robot to better adapt to variations in real deployment.
Full details of the training setup and reward design can be found in Appendix~\ref{app:low_level}.

\bl{High Level.} The high-level policy is responsible for computing the locomotion command $\mathbf{c}$ and the arm controls $(\mathbf{a_\textrm{arm}}, g)$, given language instruction $\mathbf{l}$, proprioceptive state $\mathbf s_\textrm{r}$, and the current stack of RGB observations $\mathbf{o}$.
As shown
in \Figure~\ref{fig:framework}, the command $\mathbf{c}$ is forwarded to the low-level policy, which then follows the received command for a number of time steps, before the high-level policy outputs the next command.
For the manipulator, we opt to operate in joint space, as opposed to task space or end-effector pose, so we do not have to worry about inverse kinematics computation.
Once the low-level policy is trained, it is frozen and used as a base controller by the high-level (teacher or student) policies.
In the remainder of main text, we only talk about the \textit{high-level policies} unless otherwise stated.

\subsection{Teacher-Student Framework For High-Level Training}
With the low-level policy taking over the responsibilities of legged locomotion control, high level can focus on task dependent decision making.  However, the high level policy needs to process high dimensional visual and language input, and complete the long-horizon task.  This presents an enormous space for the policy to explore, and a challenge for efficient RL training. We adopt a teacher-student learning framework ~\citep{fan2021secant, VBC, zhang2024learning} for efficient high-level training.
The overall structure of the framework is shown in the middle and bottom right sections of \Figure~\ref{fig:framework}.
As shown, the 
teacher is trained purely with RL from privileged, structured, and low-dimensional inputs $(k,\mathbf{s_\textrm{p}},\mathbf{s_\textrm{r}})$ some of which can be obtained from the simulator but not easily in the real world.
When a teacher policy is successfully learned, it is frozen and used to guide the student via behavior distillation.
As the student will eventually be deployed in the real world, its inputs no longer contain privileged information.
Besides policy distillation, we also use an RL objective to allow the student's behavior to be shaped by the same set of task rewards used by the teacher, for a potentially more effective policy than using distillation alone (details in Section~\ref{sec:baselines}, \DistillationOnly{} baseline).
In the two sections below, we explain the teacher and the student policies in more detail.

\section{The Teacher: Long-Horizon  RL with Task Decomposition and Policy Expansion}
\label{sec:teacher_learning}

The Teacher works with privileged object state information, thus, offloading the complexities of visual representation learning to the student, and only needs to focus on the following challenges of long-horizon task learning,
\begin{enumerate}
    \item \textbf{\textit{Continual exploration}}: 
    For long-horizon tasks, there may be a number of intermediate milestones (bottleneck states, {\emph{c.f.} Appendix~\ref{app:bottlebeck_states}}) that must be sequentially achieved.
    Therefore, even after reaching an intermediate milestone, the teacher must continue exploring new frontiers in order to ultimately solve the entire task.
    Without carefully encouraging continual exploration, the teacher can stop exploration early and settle on a suboptimal solution.
    \item \textbf{\textit{Loss of capacity and catastrophic forgetting}}: 
    As the teacher progresses in training, it needs to cope with both the loss of capacity issue~\citep{capacity_loss, understanding_plasticity}, which hinders the network from continual learning,
    and the catastrophic forgetting issue, which could destroy skills that have already been acquired during the learning of new tasks.
\end{enumerate}
Addressing these challenges is the key for successful teacher policy learning.  We propose to use \emph{task decomposition} and further integrate with \emph{policy expansion}~\cite{pex} for this purpose.

We first decompose the long-horizon task $\mathcal{T}$ into $K$ subtasks $\{\tau^k\}_{k=1}^K$ with shorter horizons.
To incorporate the task decompositional structure in learning, we leverage the fact that privileged information 
is accessible to the teacher and thus include the subtask index $k$ as part of the privileged observation.

\begin{figure}[t]
    \centering
    \begin{overpic}[width=\linewidth]{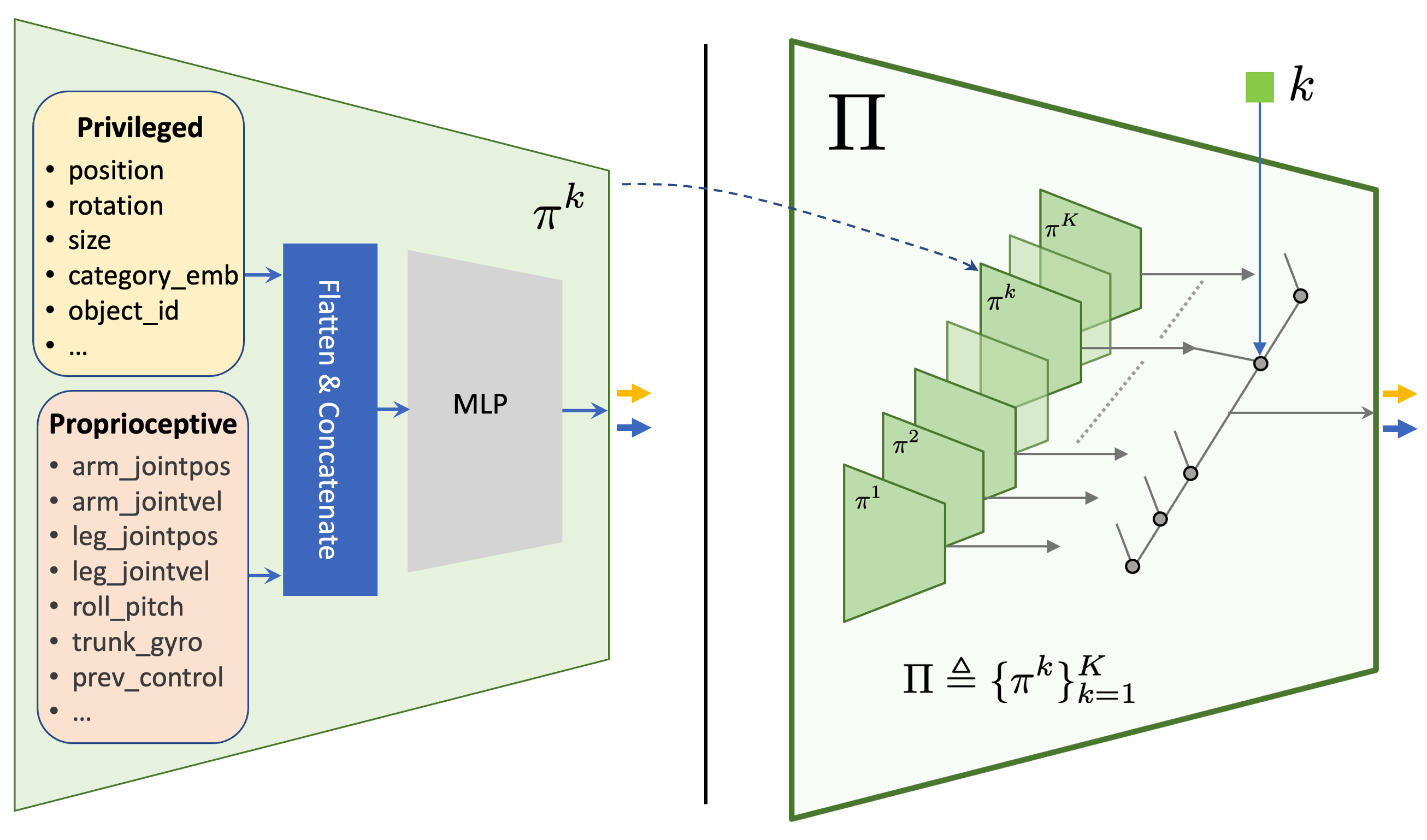}
    \end{overpic}
    \vspace{-0.2in}
    \caption{\textbf{Teacher Policy Network Structure.}
    The full teacher network $\Pi$ is a set of structurally identical networks $\{\pi^{k}\}_{k=1}^K$ gated by the subtask id $k$.
    On the left, each individual teacher policy $\pi^k$ takes a set of privileged and proprioceptive input, flattening and concatenating them before passing it through an MLP. 
    On the right, we show the full teacher network $\Pi$. Given an subtask id $k$, the $k$-th individual teacher policy is activated during computation, \emph{i.e.,} $\Pi[k] \equiv \pi^{k}$.
    }
    \vspace{-0.1in}
    \label{fig:expert_nn}
\end{figure}

Secondly, we can solve the long-horizon task by creating new network instances whenever a new subtask is encountered along the decomposed long-horizon task. 
This way, there will be dedicated policies for continual exploration and learning of new subtask without affecting any skills acquired in the previous subtasks.
Intuitively, it works as follows. 
We initiate the exploration and learning with a single policy network $\Pi=\{ \pi^1\}$ that is responsible for learning to solve the initial subtask.\footnote{We use the term policy network to represent all the networks that are required for learning a policy. For example, in the context of Actor-Critic formulation, it encompasses both the actor and critic networks.}  Whenever a new subtask is encountered a new policy is added into the policy set, \emph{i.e.,} $\Pi=\{\pi^1, \pi^2\}$.
By doing this progressively for the all the $K$ subtasks, we get  
\begin{align}\label{eq:expert}
    \Pi &\triangleq \{ \pi^k\}_{k=1}^K,
\end{align}
where $\pi^k$ denotes an individual teacher policy for subtask $k$, and $\Pi$ denotes the full teacher network comprised of a set of individual teacher policies,
as illustrated in \Figure~\ref{fig:expert_nn}.
This can be regarded as a progressive application of the policy expansion (PEX) scheme~\cite{pex} and we dub it as Progressive PEX.

Because of the policy expansion at each subtask transition, it addresses the two challenges brought by the long-horizon task learning naturally:
1) it can achieve multi-stage continual exploration and the full ability of  exploration is always ensured when entering a new stage; 2) since the policy for solving one subtask is encapsuled in a dedicated network, it mitigates the catastrophic forgetting issue for the already learned subtask policies. For a newly encountered subtask, the dedicated network that will be newly allocated addresses the issue of lost plasticity.

The left side of \Figure~\ref{fig:expert_nn} provides a graphical illustration of one individual teacher policy network $\pi^k$, taking low-dimensional proprioceptive and privileged observation as input and is responsible for learning to solve the corresponding subtask.
When necessary, we use $\piexpert^{k}(\mathbf a_{\text{hi}} | \mathbf{s})$ in place of $\pi^k$ to highlight its input and output and its role as a teacher network.
The right side of \Figure~\ref{fig:expert_nn} shows the full teacher network 
$\Pi$.  A forward pass through $\Pi$ is carried out by the proper individual network indexed from  $\Pi$ with subtask id $k$: $\Pi[k] \equiv \pi^{k}$.

Based on this implementation, the training of $\Pi$ is closely related to multi-task RL~\cite{metaworld, PaCo}.  For each training iteration, we get a batch of samples containing the subtask indices from the replay buffer and use each sample in the batch for training the sub-network associated with the subtask index of that sample.
In this work, we train the teacher policy using a multi-task variant of SAC algorithm~\cite{sac, metaworld, PaCo}.
The performance comparison between the method using a standard network structure and the Progressive PEX approach on solving long-horizon tasks is provided in Appendix~\ref{app:pex_results}, which clearly shows Progressive PEX has a much stronger ability in long-horizon task learning.

\section{The Student: Policy Distillation Guided RL}
\label{sec:student}

The student policy is the actual high-level policy that gets deployed in the real world.
It perceives the surrounding environment using the single RGB stream and motor sensors, and follows language instructions.
Since the student has to work in the real world, it does not have access to any privileged information \emph{e.g.} which subtask it is trying to solve currently.
Therefore, the student has to learn a single policy to solve the entire long-horizon task as a whole.

We train the student by distilling the teacher's multiple subtask policies into a single task policy $\piagent$.
Distillation alone could produce a reasonable policy, but due to state drifting / out-of-distribution (OOD) issues~\cite{imitation_learning_drift, dart}, imitation cannot reach the level of success of the teacher, especially due to error compounding over a long horizon.
Further complicating the distillation, the student receives unstructured high-dimensional visual input, whereas the teacher receives structured low-dimensional state input.
Additionally, differences in model architectures and inductive biases also impact distillation effectiveness.

\begin{figure}[t]
    \centering
    \begin{overpic}[width=8cm]{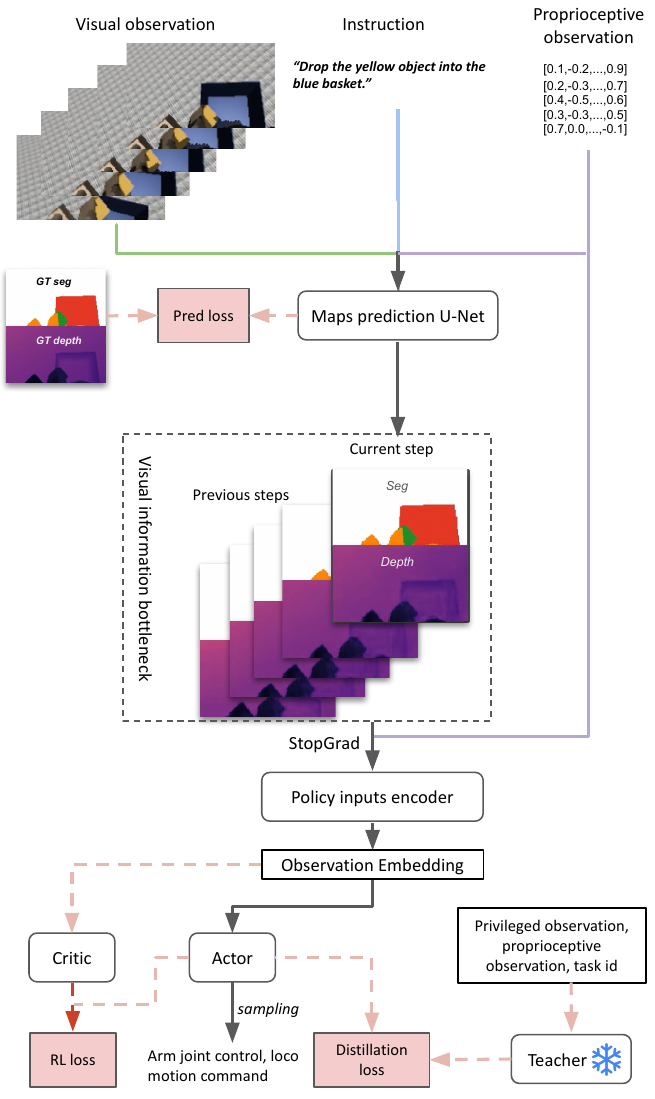}
    \end{overpic}
    \vspace{-0.05in}
    \caption{\textbf{Student Architecture Overview.}
    Red dashed lines represent data flows that only exist in simulation training.
    Solid lines denote the inference process of the student policy.
    }
    \vspace{-0.1in}
    \label{fig:agent}
\end{figure}

Given these considerations, the student can benefit from some adaptation of the teacher's skills. We choose to boost the student's policy with RL under the same set of task rewards used to train the teacher (\Figure~\ref{fig:agent}).

We train the student by modifying SAC~\citep{sac} to incorporate the distillation loss properly.
First, we use a mixed rollout strategy to generate replay data. 
At the beginning of a new episode, with a probability of $\beta$, we will sample actions for the entire episode from the student policy, otherwise from the teacher policy.
On the one hand, high-performing trajectories from the teacher along the whole task horizon facilitates coverage of future subtasks that the student cannot yet solve.
On the other hand, student rollout allows the student to explore and improve its behavior.
Second, we remove the entropy reward from policy evaluation following SACLite~\citep{yu2022you}, and we remove the entropy term from policy improvement, and replace it with the distillation loss with a fixed weight $\alpha$:
\begin{equation}
\begin{array}{l}
\displaystyle\max_{\piagent}\displaystyle\mathbb{E}_{(\mathbf s_\textrm{stu},\mathbf{s}_\textrm{p},k)\sim \mathcal{D}_{\text{replay}}}
\Big[\mathbb{E}_{\mathbf{a}_{\text{hi}}\sim\piagent(\cdot|\mathbf s_\textrm{stu})}Q\big(\mathbf s_\textrm{stu},\mathbf{a}_{\text{hi}}\big)\\
\ \ \ \ \ \ \ -\alpha {\rm KL}\!\left(\piexpert^k(\cdot|\mathbf{s}_\textrm{r},\mathbf{s}_\textrm{p}) \, || \, \piagent(\cdot| \mathbf s_\textrm{stu}) \right)\Big],\\
\end{array}
\label{eq:student}
\end{equation}
where $\mathbf s_\textrm{stu} = [\mathbf o, \mathbf s_\textrm{r}, \mathbf l]$.
To ensure that the KL term encourages the student policy $\piagent$ to explore, we modify the teacher policy $\piexpert$, keeping the mode of the action distribution unchanged but assign a fixed modal dispersion $\sigma$ (\emph{e.g.,} std. for Gaussian).
This results in a distillation loss that achieves two goals: imitating the mode while encouraging exploration.

\Figure~\ref{fig:agent} outlines the general architecture of the student.  More architecture details are in Appendix~\ref{app:model}.
When designing the student's representation model, we specifically take visual \simtoreal{} gap reduction into consideration, because once the student is trained, it will be directly deployed in real without any finetuning.
While the student could have a naive representation model that simply fuses multimodal inputs and generates a latent encoding for its policy to use, we choose to add an intermediate \emph{visual information bottleneck} (\Figure~\ref{fig:agent}) to reduce the visual \simtoreal{} gap while achieving better interpretability.

The idea is to let the policy depend on as minimal visual information as possible, namely, to limit the complexity of its input.
This can lead to potentially better generalization.
While there might be many choices for the visual bottleneck, from discrete latent codes~\citep{hafner2022deep} to canonical images~\citep{james2019sim}, we choose to use a pair of segmentation and depth maps.
There are mainly three reasons for this choice. 
First, we believe that they contain the minimal visual information required to achieve our task without incurring much task-relevant information loss.
Second, ground-truth segmentation and depth maps can be efficiently obtained from the simulator for high-quality supervised learning of such a vision bottleneck.
Thus when training the student, we add a supervision loss for predicting the segmentation mask and depth map from its inputs. 
Note that only the robot and task-related objects based on the instruction will be segmented.
Pixels from all other task-irrelevant objects will be classified as ``background''.
We use a U-Net~\citep{unet} to predict the segmentation and depth maps.
This is achieved by combining instruction encodings with image feature maps along the downsampling path of the U-Net using FiLM \citep{film}.
After this, the segmentation and depth maps are fed to the downstream policy network to generate actions.
Third, this design leads to interpretable visual representations, which is helpful for us to diagnose the visual \simtoreal{} gap.

\section{Task Implementation In Simulation}\label{sec:task_decompositions}

\begin{figure*}[t]
    \centering
    \includegraphics[width=\textwidth]{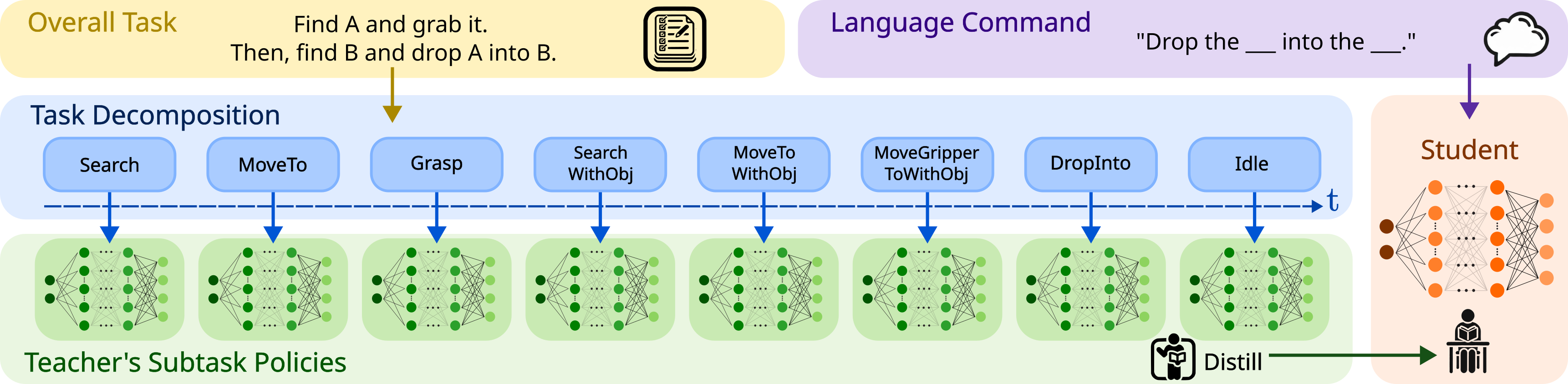}
    \caption{\textbf{Task Decomposition as Privileged Information (for teacher only).} Given a task, we decompose it into a sequence of subtasks. As shown above, the task of finding an object A, grasping it, and then dropping into an object B can be decomposed into a sequence of subtasks.
    In simulation, the subtask boundaries and transitions are well-defined, serving as a form of privileged information to the teacher. 
    The teacher can leverage this information, and use a separate policy to learn each subtask. 
    This expertise is then distilled into the student, which is conditioned on both visual camera feed and language commands.}
    \label{fig:task_decomposition}
\end{figure*}

As mentioned previously in Section~\ref{sec:teacher_learning}, \name{} employs task decomposition with progressive PEX to train the teacher policy to solve the long-horizon task.
In this work, we focus on a multi-stage search, grasp, transport and place task
(\Figure~\ref{fig:policy_visualization}).
This task will also serve as a concrete example to illustrate task decomposition. 
Given a language instruction (\emph{e.g.,} ``\instr{Drop the blue cube into the green basket.}"), the robot needs to locate an object A (the blue cube), pick it up, locate the container object B (the green basket), and then drop it into the container.
This task can be naturally broken down into the following sequential stages, where object A and B are specified by the language instruction:
\begin{enumerate}
\item \Search{}: The robot must first search for object A using the RGB images from the wrist-mounted ego-centric camera.
\item \MoveTo{}: After locating object A, the robot should approach it until it is within the reach of the arm.
\item \Grasp{}: The robot attempts to grasp and pick up object A. 
\item \SearchWithObj{}: After picking up object A, the robot should then search for container B. Within this stage and those after, there should be some coordination between the arm and the locomotion movement to expedite task success while avoiding dropping the grasped object.
\item \MoveToWithObj{}: After locating container B, the robot should approach it until it is within the reach of the arm.
\item \MoveGripperToWithObj{}: The robot then moves its gripper towards container B until reaching a target position right above the container.
\item \DropInto{}: Finally, the robot releases its gripper and drops object A into container B.
\end{enumerate}

\Figure~\ref{fig:task_decomposition} provides an illustration of the full task decomposition.
Note that in this figure, we append an auxiliary \Idle{} subtask, which encourages the robot to return its arm back to its starting position after finishing \DropInto{}.
This is an optional subtask inserted purely to expedite consecutive deployment sessions. 
We do not include \Idle{} in evaluations.

Task decomposition offers many more benefits, in addition to its usage in the progressive PEX formulation for Teacher training as discussed in Section~\ref{sec:teacher_learning}.
First, it offers a modular design of task rewards.
All subtasks contain a sparse subtask success reward.
For the subtasks that require moving the robot (or gripper) to a target region (\emph{e.g.,} \MoveTo{}, \Grasp{}, \emph{etc.}), we add a distance-based shaping reward to encourage exploration toward the target.
For \SearchWithObj{} and \MoveToWithObj{}, we add an arm-retract reward (another distance-based shaping reward), to encourage the arm to stay close to a predefined target pose shown in \Figure~\ref{fig:robot_system}. 
This reward serves to promote good vantage for the wrist camera and discourages other suboptimal behaviors after grasping such as looking at the ground during the \SearchWithObj{} subtask.
Its impact is investigated in the ablations in Section~\ref{sec:baselines} (\NoArmRetract{} baseline).

Second, task decomposition also allows various behavior priors to be conveniently integrated into the policy.
This enables us to encourage the agent to remain stationary during manipulation, or avoid walking off the workspace during search,  which are key for both safe and successful deployments (Section~\ref{sec:main_results}).

More details on the subtasks, rewards, and behavior priors are provided in Appendix~\ref{appendix:high_level_reward}.

\let\originalbigstar\bigstar
\renewcommand{\bigstar}{\textcolor{NavyBlue}{\ensuremath{\originalbigstar}}} 

\begin{table*}[t]
\centering
\renewcommand{\arraystretch}{1.2}
\caption{The suite of \simtoreal{} techniques adopted by \name{} with their importance ratings. 
$\bigstar\bigstar\bigstar$: critical; without it the whole system won't work at all.
$\bigstar\bigstar$: somewhat important; without it the system could still obtain some success in certain scenarios.
$\bigstar$: marginal; provides some boost to the system's performance.
}
\resizebox{\textwidth}{!}{
\begin{tabular}{l|l|l|l|l}
\hline
\multirow{2}{*}{\textbf{Category}} & \multirow{2}{*}{\textbf{Technique}} & \multirow{2}{*}{\textbf{Purpose}} & \multirow{2}{*}{\textbf{Reference}} & \textbf{Importance}\\[-0.3em]
& & & & \textbf{Rating (1-3)}\\
\hline 
\multirow{6}{*}{Dynamics} 
& \multirow{2}{*}{Arm PID Control} & Minimizes tracking errors and reduces arm \simtoreal{} gap  & & \multirow{2}{*}{$\bigstar\bigstar\bigstar$} \\[-0.3em]
& & without ad hoc system identification & & \\
\cline{2-3}\cline{5-5}
& Stationary Manipulation & Avoids arm tremors during grasping due to sim-to-real gap & \multirow{3}{*}{Appendix~\ref{app:dynamics_gap}} & $\bigstar\bigstar\bigstar$ \\
\cline{2-3}\cline{5-5}
& \multirow{2}{*}{Object Perturbations} & Increases robustness of grasping by avoiding memorizing & & \multirow{2}{*}{$\bigstar\bigstar\bigstar$} \\[-0.3em]
& &  a deterministic trajectory that fails in the real world  & & \\
\cline{2-3}\cline{5-5}
& Arm Mount Perturbation & Increases robustness to actual arm base height & & $\bigstar\bigstar$ \\
\cline{2-3}\cline{5-5}
& Arm Control Noise & Increases robustness to control noises in real & & $\bigstar$ \\
\hline
\hline
\multirow{7}{*}{Vision} & \multirow{2}{*}{Visual Information Bottleneck} & Makes the RL policy depend on minimal visual information & \multirow{2}{*}{Section~\ref{sec:student}} & \multirow{2}{*}{$\bigstar\bigstar\bigstar$}\\[-0.3em]
& &  for better generalization in real & & \\
\cline{2-5}
& Texture Randomization & Improves recognition of task objects by adding low-level visual distractions & \multirow{5}{*}{Appendix~\ref{app:visual_gap}} & $\bigstar\bigstar$ \\
\cline{2-3}\cline{5-5}
& Background Objects Randomization & Improves recognition of task objects by adding distractor objects & & $\bigstar\bigstar$ \\
\cline{2-3}\cline{5-5}
& Color Modeling & Increases robustness of color recognition under different lighting conditions & & $\bigstar\bigstar$\\
\cline{2-3}\cline{5-5}
& Spatial Augmentation & Image data augmentation for model robustness &  & $\bigstar$ \\
\cline{2-3}\cline{5-5}
& Image Domain Randomization & Reduces overfitting to synthetic images by pixel-level perturbations & & $\bigstar$ \\
\hline
\end{tabular}
}
\label{tab:sim2real_techniques}
\end{table*}

\section{Sim-to-Real Gap Reduction Techniques} 
\label{sec:sim2real_gap_redu}

In this section, we discuss crucial techniques for addressing the \simtoreal{} gap in both dynamics and vision, which are key for successful real-world deployment.

The dynamics \simtoreal{} gap is the mismatch between the transition function of the simulation versus that of the real world.  It can be due to the mismatch in the physical properties of objects and motors, or factors that are not properly simulated, such as friction and backlash.
This difference in dynamics can cause policies trained purely in simulation to fail when deployed in the real world, especially if the task requires accurate motor control, such as grasping a small object.

A second source of the \simtoreal{} gap for image-conditioned policies is the visual gap.
This gap arises from the inherently more complex and dynamic conditions of the real world (\emph{e.g.}, lighting, texture, camera noise), which are hard to be fully captured by simulators, leading to discrepancies in visual data.
Since we do not assume knowing the target scenes in advance, we have to ensure that the perception model is able to handle a wide range of vision scenarios.

For brevity, we list key techniques in Table~\ref{tab:sim2real_techniques}.
Comprehensive details can be found in Appendix~\ref{app:sim2real}.

\section{Physical System, Experiments and Results}
\label{sec:result}

\subsection{Physical Robotic System}

\begin{figure}[t]
    \centering
    \includegraphics[width=0.7\columnwidth]{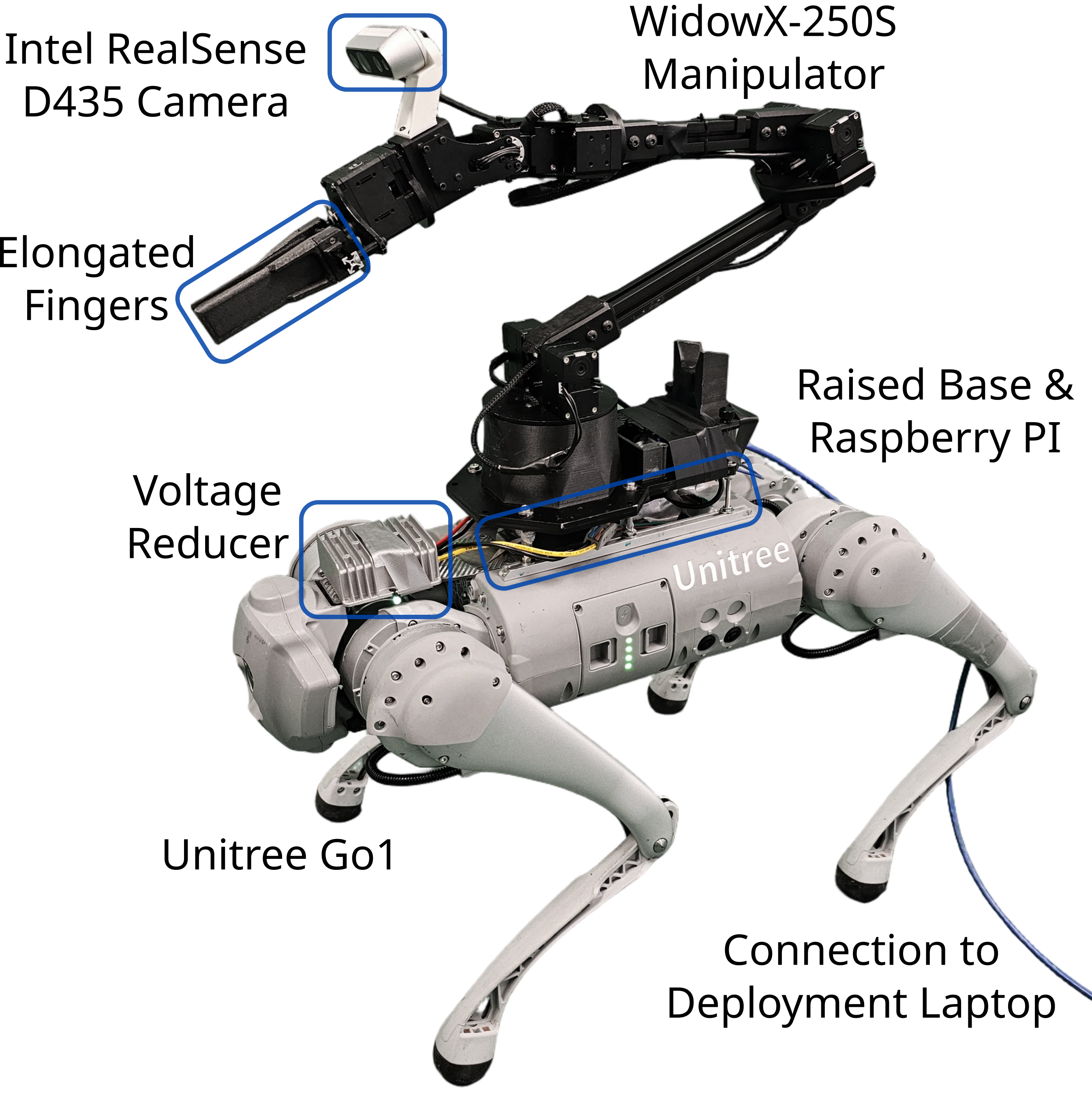}
    \vspace{-0.05in}
    \caption{\textbf{Qualitative Overview of the Robot System.}}
    \label{fig:robot_system}
\end{figure}

Our robotic system uses a Unitree Go1 with a top-mounted WidowX-250S manipulator. 
An Intel RealSense D435 camera is attached to the WidowX's wrist via a 3D-printed mount.
The RGB stream of D435 is used as the robot's sole visual feed (\Figure~\ref{fig:robot_system}).
Additionally, custom 3D-printed, elongated parallel fingers are fitted on the WidowX gripper to extend its reach. 
The inner surface of the fingers are padded with a thin layer of foam, which acts to increase friction for grasping.
Finally, the base of the WidowX is raised 50mm to allow for easy access to the onboard Raspberry PI's USB port and the ethernet port.
Overall, our entire robot system is relatively low-cost\footnote{The total cost of the development version of the quadraped (Go1-edu) plus the arm is around \$12K USD.
With the non-edu version, total cost is even lower ($\sim$6K USD) and reflects more closely the cost of mass produced parts.}.

\begin{table}[t]
    \renewcommand{\arraystretch}{1.2}
    \centering
    \caption{The dimensionality values for the observations and actions of \name{}. 
    For the meanings of these notations, we refer the reader to Section~\ref{sec:architecture}.
    }
    \resizebox{\columnwidth}{!}{
    \begin{tabular}{c|c|c|c|c|c|c|c|c|c|c}
        \hline
        $L$ & $H$ & $W$ & $N$ & $D$ & $K$ & $Q_m$ & $M$ & $z$ & $I$ & $J$  \\
        \hline
        $100$ & $90$ & $160$ & $5$ & $66$ & $7$ & $19$ & $6$ & $0.05$ & $6$ & $12$  \\
        \hline
    \end{tabular}    
    }
    \label{tab:dimensions}
\end{table}

All model inference is performed on a laptop with a 12$^\textrm{th}$ Gen Intel i9-12900H CPU, NVIDIA RTX 3070Ti laptop GPU, and 16GB of RAM. 
To minimize latency, we execute high- and low-level inference asynchronously (Appendix~\ref{sec:latency}) at a frequency of 10 and 50\,Hz, respectively. 
The outputs from the models are then fed to the Raspberry PI running our custom arm and quadruped drivers operating at approximately 500\,Hz. 
These drivers are responsible for communicating with the onboard controllers on the physical robot as well as caching fresh sensor data.
The camera operates at a fixed 60\,FPS with a resolution of $424\times240$ for minimal latency. These images are then resized to $160\times90$ before being passed to the model.
A schematic of our overall asynchronous system architecture is in \Figure~\ref{fig:system_architecture} in the appendix.

Given our hardware setup, the dimensionality values from Section~\ref{sec:architecture} are summarized in Table~\ref{tab:dimensions}.

\subsection{Ablations and Baselines}\label{sec:baselines}

We compare \name{} against the following baselines, where the first four are ablations and the last is human teleoperation as a reference:
\begin{itemize}
    \item \NoArmRetract: Remove the arm retract (raise) reward in RL training (Section~\ref{sec:task_decompositions}).
    \item \NoPerturb: Remove arm control randomization, arm mount perturbation, and object perturbation (Appendix~\ref{app:dynamics_gap}).
    \item \NoVisualAug: Remove random background objects and spatial visual augmentations (Appendix~\ref{app:visual_gap}).
    \item \DistillationOnly: Train student without the RL loss, keeping only the distillation and representation loss (Section~\ref{sec:student}).
    \item \HumanTeleop: An expert human teleoperator that provides locomotion commands and delta EE pose to the robot via a joystick. Delta EE is handled via IK. The teleoperator shares the same observation space as the policy, 
    \emph{i.e.,} observing solely through the wrist camera stream,
    and is allowed to practice on the system for an hour.
    While not fully autonomous, human teleoperation provides a valuable reference point to compare our model against. See Appendix~\ref{app:teleop} for more details.
\end{itemize}

{\flushleft These ablation baselines are chosen due to their significant impact to the overall system.} 
 
Some of the other techniques mentioned before (Table~\ref{tab:sim2real_techniques} and Section~\ref{sec:architecture}), \emph{e.g.} arm PID control, stationary manipulation, and rotational search are essential for both safe and successful deployment, and are applied to all the methods, thus do not appear in Table~\ref{tab:real_world_experimental_results}.\footnote{We also tried another baseline \NoDistillation{}, which trains the student policy directly via RL without distillation from the teacher policy (\emph{i.e.,} RL for visuomotor policy learning from scratch). 
This baseline cannot learn to solve the task at all in simulation, because of the compounded difficulties of visual representation learning, behavior learning, and long-horizon exploration. It is excluded from the results.}

\subsection{Metrics and Evaluation Protocol}

To accurately evaluate policy performance, we run the entire training pipeline across three different random seeds for all methods, training from scratch the low-level, teacher, and student policies. 
Note that high level policies of all different methods for one particular random seed share the same low-level policy trained using that seed.
We report the following two key metrics when deploying the final student  policy (together with the low-level policy used in training):
\begin{enumerate}
    \item \textbf{Cumulative Subtask Success Rate}:
    The success rate up to each subtask from the beginning of the episode.
    \item \textbf{Episode Time}: The time spent completing the full task from the beginning. We apply a time limit of $t_{\rm max}\!=\!90$\,s. For failed episodes, we use $t_{\rm max}$ as the episode time.
\end{enumerate}

\begin{figure*}[t]
    \centering
    \begin{overpic}[width=18cm]{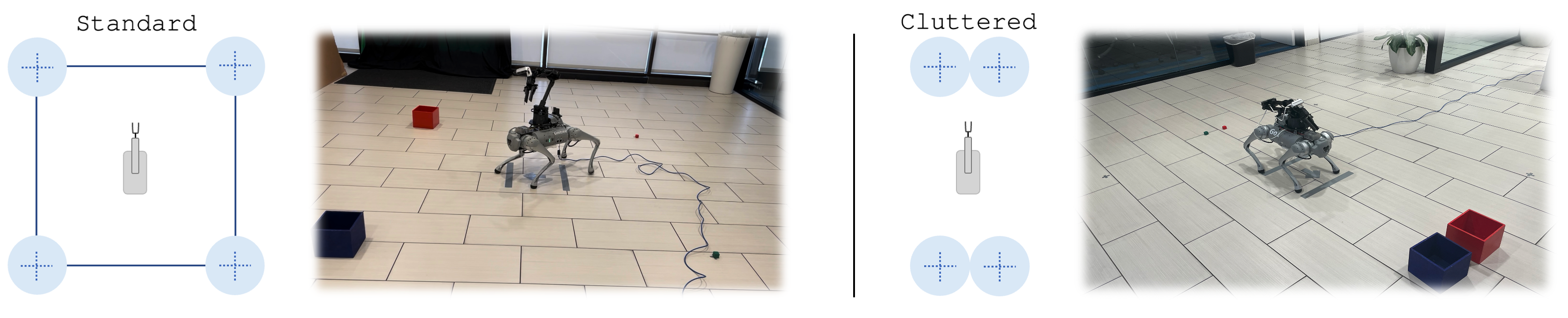}
    \end{overpic}
    \vspace{-0.1in}
    \caption{\textbf{Scene Spatial Layouts in Lobby Environment.} 
    {\ttfamily Standard}: we put the robot in the center of a $2{\rm m}\! \times\! 2{\rm m}$ square space and objects including objects to be grasped and containers to drop the graspable objects into on the four corners. The instructions and object placements vary across episodes.
    During training, the spatial layout used for the objects are randomize based on the {{\ttfamily Standard}} setting, with each position sampled from a circular region centered around each corner with a radius of $0.5$m.
    {\ttfamily Cluttered}: objects are placed close to each other in front and behind the robot, leading to a spatial layout that is out of the training distribution (OOD). The instructions and object placements vary across episodes.
    }
    \label{fig:spatial_layout}
\end{figure*}

{\flushleft For the main evaluation, we use the {\ttfamily Standard} object spatial layout in an indoor {\small \textsf{Lobby}} scene, as shown in \Figure~\ref{fig:spatial_layout}, to maintain repeatability across all methods and seeds.}
For task objects, we use cubes and baskets of different colors.
For each method of each random seed, we roll out the policy for 20 episodes with varying object colors and positions according to a consistent evaluation protocol. This amounts to 360 real-world episodes in Table~\ref{tab:real_world_experimental_results}.
More details on the physical objects and protocol can be found in Appendices~\ref{app:physical_objects} and \ref{app:eval_protocol}, respectively.
In addition to the {\ttfamily Standard} spatial setup in {\small \textsf{Lobby}}, we also evaluate the robot in several other scene and layout combinations in subsequent experiments, with another 40 real-world episodes (\Figure~\ref{fig:scene_variation_results}), leading to a total of 400 real-world episodes of experiments.

\begin{table*}[t]
\renewcommand{\arraystretch}{1.2}
\centering
\caption{Real-world task success rates and episode time (mean$\pm$stddev over 3 random seeds), with 20 episodes per seed.}
\resizebox{\textwidth}{!}{
\begin{tabular}{l|c| D{,}{\pm}{-1} D{,}{\pm}{-1} D{,}{\pm}{-1} D{,}{\pm}{-1}|D{,}{\pm}{-1}}
\toprule
\multirow{2}{*}{\textbf{Method}} 
& \multirow{2}{*}{\textbf{Autonomous}}
& \multicolumn{4}{c|}{\textbf{Cumulative Success Rate} [\%] $\uparrow$} 
& \multicolumn{1}{c}{\multirow{2}{*}{\textbf{Episode Time} [s] $\downarrow$}} \\

&  & \multicolumn{1}{c}{\texttt{Search+MoveTo}} 
& \multicolumn{1}{c}{\texttt{Grasp}} 
& \multicolumn{1}{c}{\texttt{Search+MoveTo(WObj)}} 
& \multicolumn{1}{c|}{\texttt{DropInto} (Full Task)} 
& \\
\midrule
\textbf{\NoArmRetract} & \tikzcmark & \bkcolor{100.0},\bkcolor{0.0} & 58.3,51.1 & 28.3,40.7 & 5.0,8.7 & 87.3,4.8 \\
\textbf{\NoPerturb} & \tikzcmark & \bkcolor{100.0},\bkcolor{0.0} & 53.3,46.5 & 43.3,37.5 & 43.3,37.5 & 62.4,23.9 \\
\textbf{\NoVisualAug} & \tikzcmark  & 98.3,2.4 & 88.3,4.7 & 63.3,26.6 & 56.6,22.5 & 52.9,18.7 \\
\textbf{\DistillationOnly} & \tikzcmark & 93.3,4.7 & 71.1,33.2 & 60.0,38.9 & 50.0,35.5 & 61.2,25.5 \\
\cmidrule(lr){1-7}
\textbf{\HumanTeleop}  & \tikzxmark  & \bkcolor{100.0},\bkcolor{0.0} & \bkcolor{96.7},\bkcolor{2.9} & 86.7,7.6 & 75.0,5.0 & 65.5,3.6 \\
\cmidrule(lr){1-7}
\textbf{\name{} (ours)} & \tikzcmark & \bkcolor{100.0},\bkcolor{0.0} & \bkcolor{96.7},\bkcolor{5.8} & \bkcolor{96.7},\bkcolor{5.8} & \bkcolor{78.3},\bkcolor{5.8} & \bkcolor{43.8},\bkcolor{6.0} \\
\bottomrule
\end{tabular}
}
\label{tab:real_world_experimental_results}
\end{table*}

\subsection{Results and Analysis}\label{sec:main_results}

Table~\ref{tab:real_world_experimental_results} presents the main results, comparing \name{} against baselines under the {\ttfamily Standard} layout in the {\small \textsf{Lobby}} scene. 
As the ablations show, all the ablating factors  contribute to the efficient and successful completion of the final task. 

\begin{figure}[t]
    \centering
    \begin{overpic}[width=9cm]{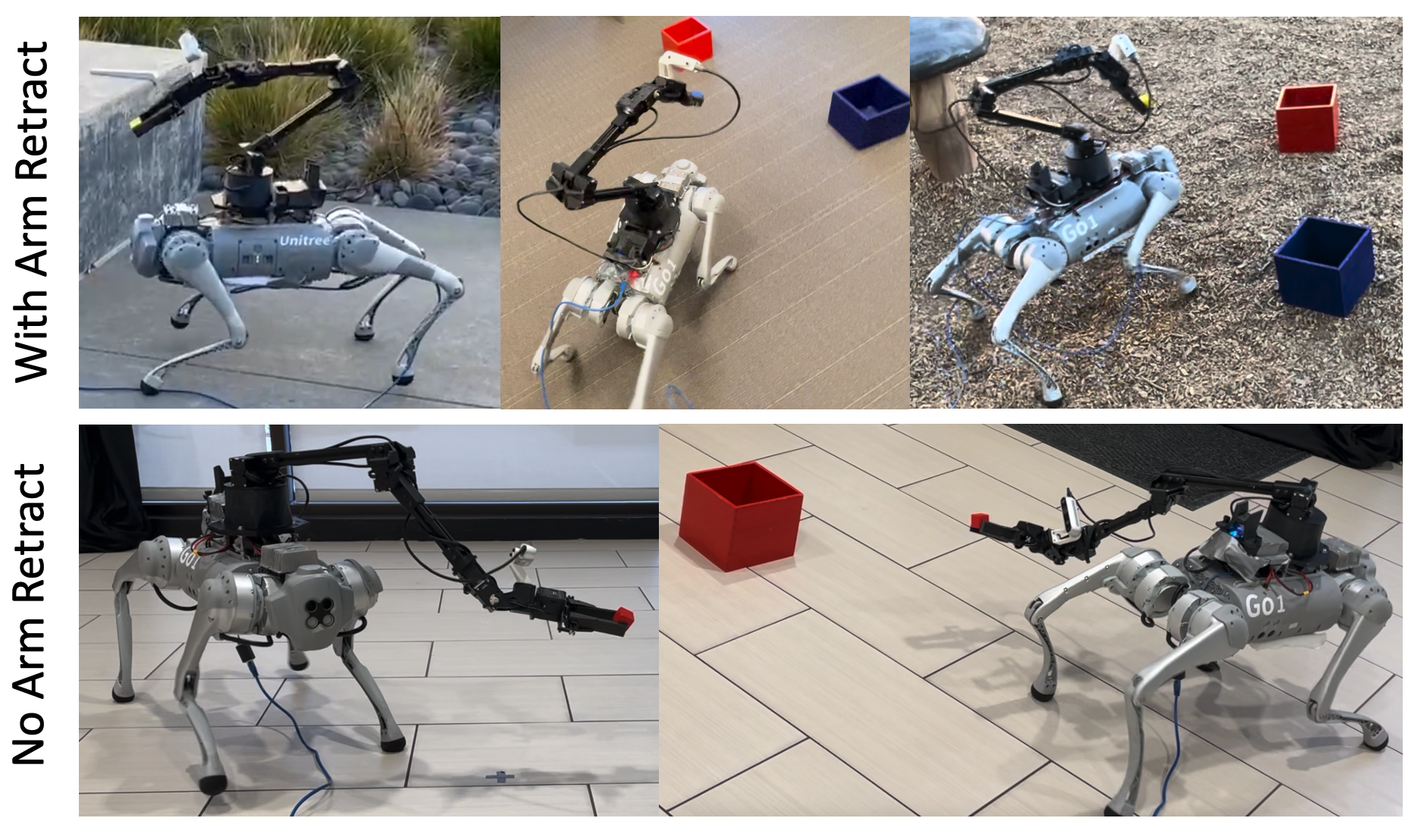}
    \end{overpic}
    \vspace{-0.15in}
    \caption{\textbf{Policy Behaviors With and Without Arm Retract.}
    The arm controlled by the policy trained without the arm-retract reward is over-extended, leading to more arm shake, cube drops and a higher chance of search failures.
    }
    \vspace{-0.1in}
    \label{fig:arm_retract_behaviors}
\end{figure}

\begin{figure*}[t]
    \centering
    \resizebox{\textwidth}{!}{
    \begin{overpic}{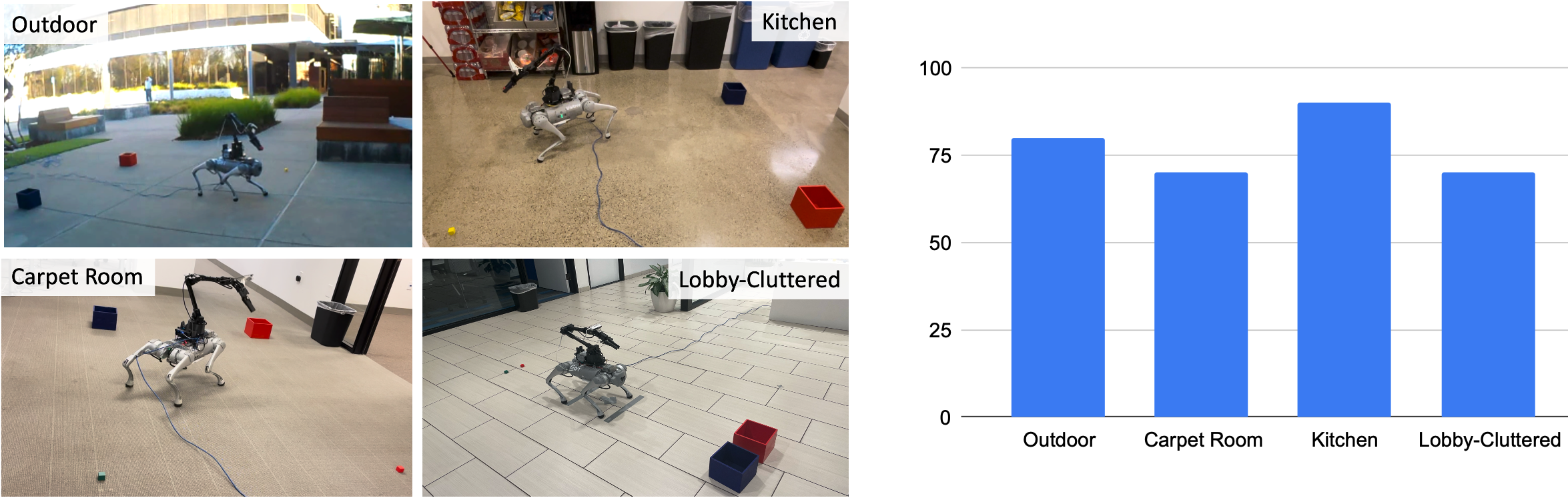}
    \end{overpic}
    }
    \vspace{-0.15in}
    \caption{\textbf{The \emph{same} \name{} Policy across Scene Variations.} Left: images showing the differences for evaluation. Right: the average full task success rate across 10 real-world deployment episodes for each of the scenes shown on the left.}
    \label{fig:scene_variation_results}
\end{figure*}

In particular, \NoArmRetract{} performs fairly well until \Grasp{}, when it only successfully locates 
the container and transports the cube to the container half of the times after a successful grasp.
There are several reasons for the failures. 
For one, without the arm retract reward acting as a soft constraint on the arm pose, the arm can remain at over-extended, close to to the ground.  This is very close to the physical torque limits of the shoulder motors responsible for raising the arm, leading to more arm shake, which in turn can result in the grasped cube slipping out of the fingers, or even motor failures. 
Secondly, right after grasp, the gripper and the camera are still relatively close to the ground, resulting in a low vantage camera view, and a higher chance of search failures. 
\Figure~\ref{fig:arm_retract_behaviors} shows the behaviors of the policies trained with and without the arm retract reward.

Another baseline, \NoPerturb{}, has the lowest grasp success. 
This is because without perturbations of any kind (object, arm mount, or control), the policy tends to remember a fixed trajectory of grasping, which is not robust against the variations in the location of the object relative to the robot. In real deployments, the quadruped can stop a bit farther away from the cube, and the policy would move the gripper fingers toward the cube without actually reaching.
Lack of variation in object relative positions is likely because during simulation training, the policy tends to stop just at the success boundary of the \MoveTo{} subtask, so that the \Grasp{} subtask usually starts with very similar locations of the target object relative to the robot, while in the real world such clear cut positioning is rarely seen.

Success of \NoVisualAug{} suffers most during the \Search{} subtasks where the model sometimes gets distracted, due to false positive detections of the target in the background. 
There's a particularly large drop of roughly 25\% when searching for the basket.
This indicates that visual augmentations are important for improving the robustness of the robot's vision module.

\DistillationOnly{} simply distills from the behavior of the highly successful teacher policy without using any task reward during student training. It can still achieve a relatively high success rate,
but, its performance is not consistent across random seeds.

In comparison, our full method \textbf{\name{}} uses distillation-guided RL, resulting in consistently high success across all seeds. 
It shows that distillation loss alone is not enough for robust policy learning. 
Task completion is also faster with the RL loss optimizing the policy further.

Overall, \textbf{\name{}} achieves the highest subtask and full task success rates among all the methods including \HumanTeleop, and is also the most efficient, taking on average 43.8 seconds to complete the full task, which is about 1.5$\times$ the speed of expert human teleoperation.

We provide visualizations of policy behavior and prediction outputs in Appendix~\ref{app:visualization}.

\subsection{Generalization to Different Real-world Scenes}
In addition to the standard scene used in Table~\ref{tab:real_world_experimental_results}, we further run real-world evaluations of \name{} 
under more scene variations (left side of \Figure~\ref{fig:scene_variation_results}): {\small\textsf{Outdoor}}, {\small\textsf{Carpet Room}}, {\small\textsf{Kitchen}}, and {\small\textsf{Lobby-Cluttered}}.
For the first three scenes, we use the same {\texttt{Standard}} spatial layout as before.
For the {\small\textsf{Lobby-Cluttered}} scene, we use the OOD \texttt{Cluttered} spatial layout (right side of \Figure~\ref{fig:spatial_layout}).
We conduct ten trials for each environment using the same \name{} policy.

All results for these four additional scenes are summarized on the right side of \Figure~\ref{fig:scene_variation_results}.
For the three novel scenes, we observe that success rates are all fairly close to the results from Table~\ref{tab:real_world_experimental_results} ($\sim$78\%).
For the {\small \textsf{Lobby-Cluttered}} scene, the success rate is just a bit lower than that of the \texttt{Standard} layout due to being out of distribution.
This shows that the model trained fully in simulation can \emph{zero-shot} adapt to a wide range of real world settings, with vastly different lighting conditions, backgrounds, and floor types and textures.

\begin{figure*}[t]
    \centering
    \resizebox{\textwidth}{!}{
    \begin{overpic}{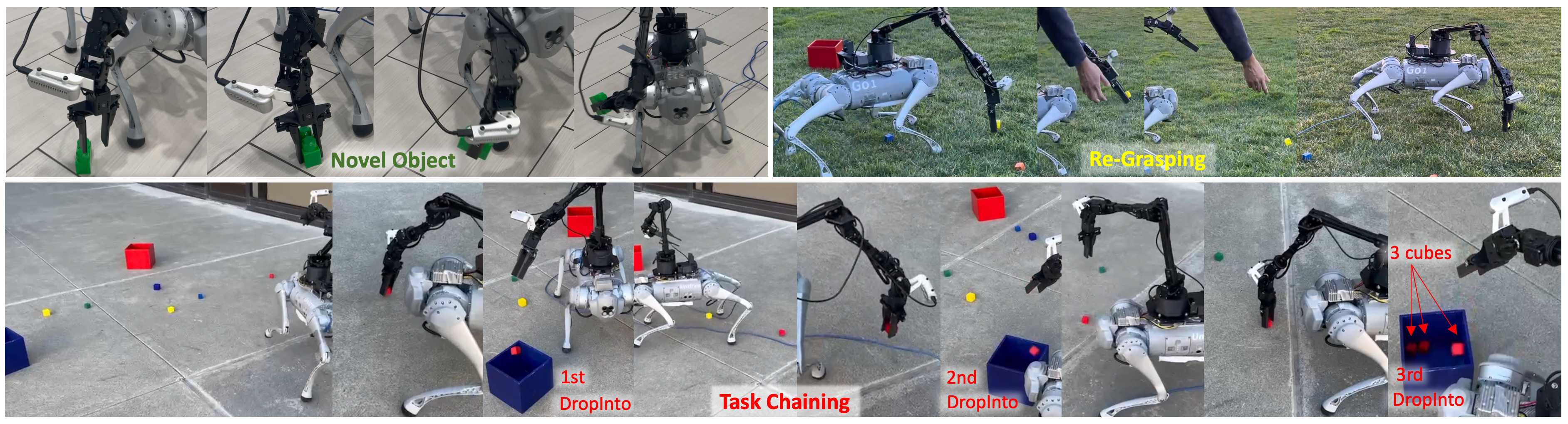}
    \end{overpic}
    }
    \vspace{-0.1in}
    \caption{\textbf{Generalization Behaviors of \name{} Policy.}
    {\small \textsf{Novel Object}}: grasping an object with a novel shape that is out of the training distribution.
    {\small \textsf{Re-Grasping}}: a human interrupts the task progress by
    removing the cube from the gripper and toss it to the ground. The \name{} policy will re-grasp the cube. 
    {\small \textsf{Task Chaining}}: executing the full task multiple times (3 times in this example) consecutively non-stop. Each time a cube is dropped into the basket, another target cube is tossed onto the ground. 
    }
    \vspace{-0.1in}
    \label{fig:generalization}
\end{figure*}

Furthermore, our qualitative evaluation showed that the robot also generalizes well to scenarios where:
\begin{enumerate}
\item task objects are randomly scattered on the ground (last two rows in \Figure~\ref{fig:slim_in_real}),
\item distractor objects are present (last row in \Figure~\ref{fig:slim_in_real}),
\item novel task object shapes (\Figure~\ref{fig:generalization} {\small \textsf{Novel Object}}), 
\item a human interrupts the task progress (\Figure~\ref{fig:generalization} {\small \textsf{Re-Grasping}}),
\item and executing the full task multiple times consecutively non-stop (\Figure~\ref{fig:generalization} {\small \textsf{Task Chaining}}).\end{enumerate}
We refer the reader to our demo videos in the \demo{} for a more intuitive viewing experience of these emergent behaviors.

\subsection{Sim \emph{vs.} Real Performance}

\begin{figure}[h]
    \centering
    \begin{overpic}[width=9cm]{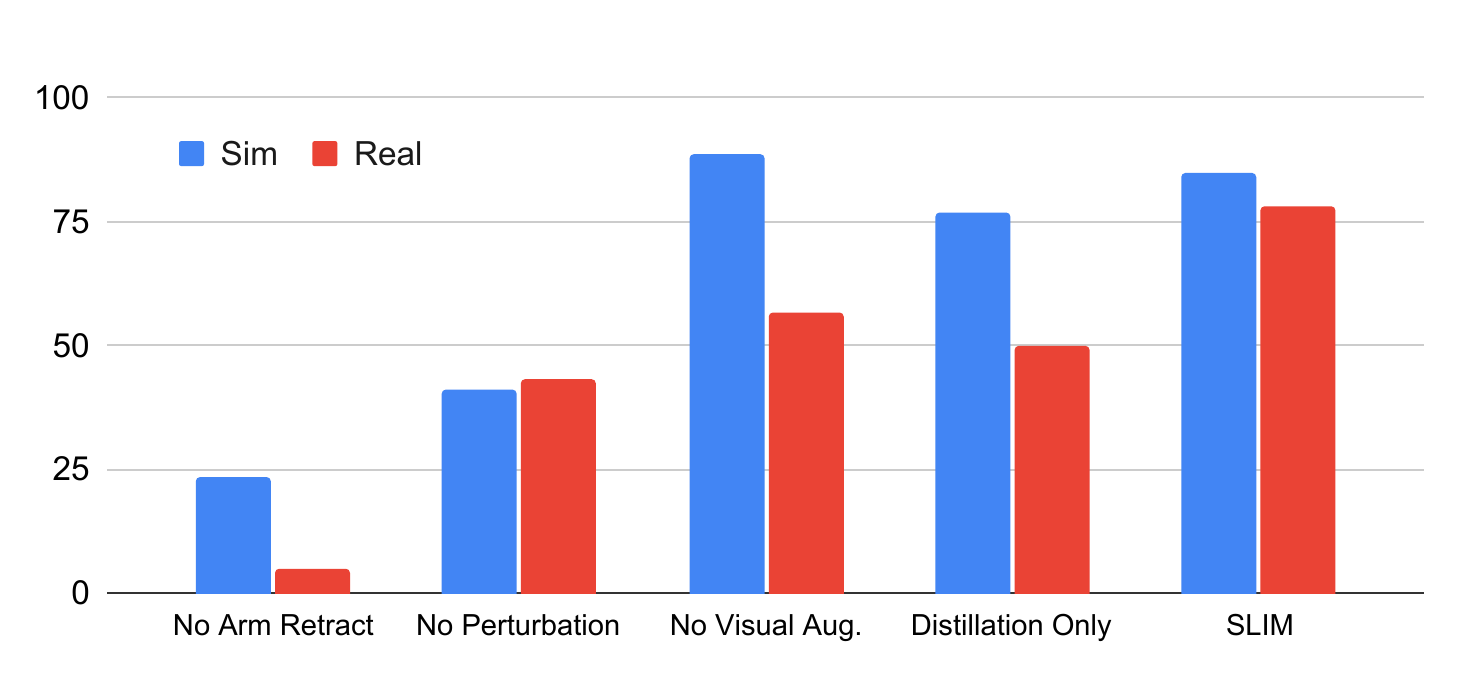}
    \end{overpic}
    \vspace{-0.15in}
    \caption{\textbf{Full Task Sim and Real Success Rate Comparison.} All success rates are the averaged success rate across three random seeds.}
    \vspace{-0.1in}
    \label{fig:sim_and_real_comparison}
\end{figure}

Different methods or even the same method with different seeds can exhibit very different behaviors, mostly due to the variance in the RL training of the teacher policies.  Additionally, because of the \simtoreal{} gap, real deployment can differ from simulation as well.
To better understand how each technique impacts final success, we showcase the performance of all methods in both simulation and real in \Figure~\ref{fig:sim_and_real_comparison}.
The sim success rate for each method denotes the average full task success rate over 100 episodes run in simulation, averaged across the three random seeds.  The simulation environment follows the same randomization setting as training, with the policy rollout being fully greedy.
The real success rate denotes the full task success rates from Table~\ref{tab:real_world_experimental_results}.

Without arm retract (\NoArmRetract{}), searching for the basket becomes more difficult in simulation as well as in real, because after grasping, the camera still points down to the ground, but searching for the basket requires many steps of arm raise. Without explicit rewards encouraging arm raise, it can create an exploration bottleneck for RL. 
Real-world performance suffers more, because moving with an out-stretched arm causes large sim-to-real gaps: the tip of the arm tends to shake more due to motor backlash, and the shoulder motors operate close to their physical torque limits, leading to more motor overloading.
For \NoPerturb{}, while one might expect perturbations to complicate RL learning, their inclusion results in much higher performance even in simulation. Without perturbations, the policy seems to converge to suboptimal solutions maybe due to a lack of exposure to a sufficiently large range of setups.
Without visual augmentations (\NoVisualAug{}), the simulation task is easier to be solved, and thus simulation success is artificially high, but real-world performance suffers when compared to the full method. RL training loss in the full \name{} method increases success in simulation (over \DistillationOnly{}) and achieves the highest success rate in real.

\subsection{Failure Modes}
We also recorded failures for \name{} and report the most common cases here.  In one case, 
the quadruped stopped a bit late during grasping and kicked the cube away.  In another case, the low-level policy shook and moved backward, causing the cube to be just out of reach. \DropInto{} sometimes failed because of either early dropping right outside of the basket, or the quadruped kicking the basket away, right before dropping, when backing towards the basket.  In rare cases where the gripper held only the top part of the cube, the gripper fingers completely blocked the cube from the camera view, and the robot got stuck hovering over the basket without dropping the cube.

\section{Limitations}\label{sec:limitations}
Although not the main focus of this work, the current form of low-level policy introduces several limitations.
One limitation is the applicable type of terrains.
Currently the low-level policy is only trained on flat terrains in simulation.
Although it shows the potential to work across different types of ground (\emph{e.g.,} concrete, tile floor, carpet, mulch, lawn, \emph{etc.}), it does not push the low-level policy to its full potential.
We can potentially leverage the recent advances in locomotion policy training~\cite{robot_parkour, extreme_parkour} to further expand the abilities and applicability of \name{} across more diverse types of terrains, \emph{e.g.,} stairs, ramp, \emph{etc.}

Another low-level related limitation is that the current form of low-level policy is not really a wholebody policy, being non-adaptive to target pose the arm is trying to reach. Although its current form is sufficient for the task considered in this work, by upgrading it to a wholebody policy~\cite{deep_wholebody}, we can further enlarge its potential workspace, due to a better coordination between the quadruped and the arm.

As a pure \simtoreal{} approach that is trained from scratch, we did not leverage any pre-trained foundation models. As a result, we did not address a large amount of visual concepts and language variations,
but instead focused on the right amount of variations that are both practical and representative enough.

There are also a few limitations on the robot system side.
Currently for visual perception, we only use the RGB image from the wrist-mounted camera, which is a minimal setting.
To further improve the robot's perception and therefore the overall ability in terms of navigation and obstacle avoidance, it might be necessary to incorporate additional sensors for perception such as depth, stereo, lidar etc. In addition, this can also provide valuable information for the robot to improve its longer-term memory via SLAM-like mechanisms~\cite{nis_slam, neural_slam}.

\section{Conclusion, Discussion and Future Work}
We present a complete legged mobile manipulation system for solving language-instructed long-horizon tasks.
The policy is trained via RL in sim plus \simtoreal{} transfer, with a progressive policy expansion-based teacher policy for solving the long-horizon task, followed by a distillation-guided RL trained student visuomotor policy. We further identify and design a suite of crucial techniques to reduce the \simtoreal{} gap.
Experimental results compared with a number of baseline methods verify the effectiveness of the proposed \name{} system.
Moreover, real-world testing of \name{} across different scenes and spatial layouts
shows that the robot can generalize well with emergent robust 
behaviors.

For future work, we plan to address the limitations discussed in Section~\ref{sec:limitations}, including 
further expanding the applicability of \name{} across more diverse scenarios
by improving locomotion policy training~\cite{robot_parkour, extreme_parkour, deep_wholebody},
extending the visual and language diversity, and
equipping the robot with additional sensors for SLAM-like ability~\cite{nis_slam, neural_slam}.

\bibliographystyle{plainnat}
\bibliography{ref}

\begin{thebibliography}{69}
\providecommand{\natexlab}[1]{#1}
\providecommand{\url}[1]{\texttt{#1}}
\expandafter\ifx\csname urlstyle\endcsname\relax
  \providecommand{\doi}[1]{doi: #1}\else
  \providecommand{\doi}{doi: \begingroup \urlstyle{rm}\Url}\fi

\bibitem[Agarwal et~al.(2022)Agarwal, Kumar, Malik, and Pathak]{agarwal2022ego}
Ananye Agarwal, Ashish Kumar, Jitendra Malik, and Deepak Pathak.
\newblock Legged locomotion in challenging terrains using egocentric vision.
\newblock In \emph{Conference on Robot Learning}, 2022.

\bibitem[Agrawal et~al.(2022)Agrawal, Chen, Rai, and
  Sreenath]{agrawal2022vision}
Ayush Agrawal, Shuxiao Chen, Akshara Rai, and Koushil Sreenath.
\newblock Vision-aided dynamic quadrupedal locomotion on discrete terrain using
  motion libraries.
\newblock In \emph{IEEE International Conference on Robotics and Automation},
  2022.

\bibitem[Aljalbout et~al.(2024)Aljalbout, Frank, Karl, and van~der
  Smagt]{aljalbout2024action_space}
Elie Aljalbout, Felix Frank, Maximilian Karl, and Patrick van~der Smagt.
\newblock On the role of the action space in robot manipulation learning and
  sim-to-real transfer.
\newblock \emph{IEEE Robotics and Automation Letters}, 9\penalty0 (6):\penalty0
  5895–5902, June 2024.

\bibitem[Black et~al.(2024)Black, Brown, Driess, Esmail, Equi, Finn, Fusai,
  Groom, Hausman, Ichter, Jakubczak, Jones, Ke, Levine, Li-Bell, Mothukuri,
  Nair, Pertsch, Shi, Tanner, Vuong, Walling, Wang, and Zhilinsky]{pi0}
Kevin Black, Noah Brown, Danny Driess, Adnan Esmail, Michael Equi, Chelsea
  Finn, Niccolo Fusai, Lachy Groom, Karol Hausman, Brian Ichter, Szymon
  Jakubczak, Tim Jones, Liyiming Ke, Sergey Levine, Adrian Li-Bell, Mohith
  Mothukuri, Suraj Nair, Karl Pertsch, Lucy~Xiaoyang Shi, James Tanner, Quan
  Vuong, Anna Walling, Haohuan Wang, and Ury Zhilinsky.
\newblock $\pi_0$: A vision-language-action flow model for general robot
  control.
\newblock \emph{CoRR}, arXiv:2410.24164, 2024.

\bibitem[Brohan et~al.(2023)Brohan, Brown, Carbajal, Chebotar, Dabis, Finn,
  Gopalakrishnan, Hausman, Herzog, Hsu, Ibarz, Ichter, Irpan, Jackson,
  Jesmonth, Joshi, Julian, Kalashnikov, Kuang, Leal, Lee, Levine, Lu, Malla,
  Manjunath, Mordatch, Nachum, Parada, Peralta, Perez, Pertsch, Quiambao, Rao,
  Ryoo, Salazar, Sanketi, Sayed, Singh, Sontakke, Stone, Tan, Tran, Vanhoucke,
  Vega, Vuong, Xia, Xiao, Xu, Xu, Yu, and Zitkovich]{rt1}
Anthony Brohan, Noah Brown, Justice Carbajal, Yevgen Chebotar, Joseph Dabis,
  Chelsea Finn, Keerthana Gopalakrishnan, Karol Hausman, Alex Herzog, Jasmine
  Hsu, Julian Ibarz, Brian Ichter, Alex Irpan, Tomas Jackson, Sally Jesmonth,
  Nikhil~J Joshi, Ryan Julian, Dmitry Kalashnikov, Yuheng Kuang, Isabel Leal,
  Kuang-Huei Lee, Sergey Levine, Yao Lu, Utsav Malla, Deeksha Manjunath, Igor
  Mordatch, Ofir Nachum, Carolina Parada, Jodilyn Peralta, Emily Perez, Karl
  Pertsch, Jornell Quiambao, Kanishka Rao, Michael Ryoo, Grecia Salazar, Pannag
  Sanketi, Kevin Sayed, Jaspiar Singh, Sumedh Sontakke, Austin Stone, Clayton
  Tan, Huong Tran, Vincent Vanhoucke, Steve Vega, Quan Vuong, Fei Xia, Ted
  Xiao, Peng Xu, Sichun Xu, Tianhe Yu, and Brianna Zitkovich.
\newblock {RT-1}: Robotics transformer for real-world control at scale.
\newblock \emph{CoRR}, arXiv:2212.06817, 2023.

\bibitem[Caluwaerts et~al.(2023)Caluwaerts, Iscen, Kew, Yu, Zhang, Freeman,
  Lee, Lee, Saliceti, Zhuang, Batchelor, Bohez, Casarini, Chen, Cortes,
  Coumans, Dostmohamed, Dulac-Arnold, Escontrela, Frey, Hafner, Jain, Jyenis,
  Kuang, Lee, Luu, Nachum, Oslund, Powell, Reyes, Romano, Sadeghi, Sloat,
  Tabanpour, Zheng, Neunert, Hadsell, Heess, Nori, Seto, Parada, Sindhwani,
  Vanhoucke, and Tan]{caluwaerts2023barkour}
Ken Caluwaerts, Atil Iscen, J.~Chase Kew, Wenhao Yu, Tingnan Zhang, Daniel
  Freeman, Kuang-Huei Lee, Lisa Lee, Stefano Saliceti, Vincent Zhuang, Nathan
  Batchelor, Steven Bohez, Federico Casarini, Jose~Enrique Chen, Omar Cortes,
  Erwin Coumans, Adil Dostmohamed, Gabriel Dulac-Arnold, Alejandro Escontrela,
  Erik Frey, Roland Hafner, Deepali Jain, Bauyrjan Jyenis, Yuheng Kuang, Edward
  Lee, Linda Luu, Ofir Nachum, Ken Oslund, Jason Powell, Diego Reyes, Francesco
  Romano, Feresteh Sadeghi, Ron Sloat, Baruch Tabanpour, Daniel Zheng, Michael
  Neunert, Raia Hadsell, Nicolas Heess, Francesco Nori, Jeff Seto, Carolina
  Parada, Vikas Sindhwani, Vincent Vanhoucke, and Jie Tan.
\newblock Barkour: Benchmarking animal-level agility with quadruped robots.
\newblock \emph{CoRR}, arXiv:2305.14654, 2023.

\bibitem[Chaplot et~al.(2020)Chaplot, Gandhi, Gupta, Gupta, and
  Salakhutdinov]{neural_slam}
Devendra~Singh Chaplot, Dhiraj Gandhi, Saurabh Gupta, Abhinav Gupta, and Ruslan
  Salakhutdinov.
\newblock Learning to explore using active neural {SLAM}.
\newblock In \emph{International Conference on Learning Representations}, 2020.

\bibitem[Cheng et~al.(2023)Cheng, Shi, Agarwal, and Pathak]{extreme_parkour}
Xuxin Cheng, Kexin Shi, Ananye Agarwal, and Deepak Pathak.
\newblock Extreme parkour with legged robots.
\newblock \emph{CoRR}, arXiv:2309.14341, 2023.

\bibitem[Dario~Bellicoso et~al.(2017)Dario~Bellicoso, Jenelten, Fankhauser,
  Gehring, Hwangbo, and Hutter]{dynamic_quad_locomotion}
C.~Dario~Bellicoso, Fabian Jenelten, Péter Fankhauser, Christian Gehring,
  Jemin Hwangbo, and Marco Hutter.
\newblock Dynamic locomotion and whole-body control for quadrupedal robots.
\newblock In \emph{International Conference on Intelligent Robots and Systems},
  2017.

\bibitem[Downs et~al.(2022)Downs, Francis, Koenig, Kinman, Hickman, Reymann,
  McHugh, and Vanhoucke]{downs2022google}
Laura Downs, Anthony Francis, Nate Koenig, Brandon Kinman, Ryan Hickman, Krista
  Reymann, Thomas~B McHugh, and Vincent Vanhoucke.
\newblock Google scanned objects: A high-quality dataset of {3D} scanned
  household items.
\newblock In \emph{IEEE International Conference on Robotics and Automation},
  2022.

\bibitem[Duan et~al.(2024)Duan, Zhuang, Zhao, and
  Schwertfeger]{playfuldoggybot}
Xin Duan, Ziwen Zhuang, Hang Zhao, and Soeren Schwertfeger.
\newblock Playful {DoggyBot}: Learning agile and precise quadrupedal
  locomotion.
\newblock \emph{CoRR}, arXiv:2409.19920, 2024.

\bibitem[Fan et~al.(2021)Fan, Wang, Huang, Yu, Fei-Fei, Zhu, and
  Anandkumar]{fan2021secant}
Linxi Fan, Guanzhi Wang, De-An Huang, Zhiding Yu, Li~Fei-Fei, Yuke Zhu, and
  Anima Anandkumar.
\newblock Secant: Self-expert cloning for zero-shot generalization of visual
  policies.
\newblock In \emph{International Conference on Machine Learning}, 2021.

\bibitem[Fankhauser et~al.(2018)Fankhauser, Bjelonic, Dario~Bellicoso, Miki,
  and Hutter]{robust_quad_walking}
Peter Fankhauser, Marko Bjelonic, C.~Dario~Bellicoso, Takahiro Miki, and Marco
  Hutter.
\newblock Robust rough-terrain locomotion with a quadrupedal robot.
\newblock In \emph{IEEE International Conference on Robotics and Automation},
  2018.

\bibitem[Feng et~al.(2022)Feng, Zhang, Li, Peng, Basireddy, Yue, Song, Yang,
  Liu, Sreenath, and Levine]{feng2022genloco}
Gilbert Feng, Hongbo Zhang, Zhongyu Li, Xue~Bin Peng, Bhuvan Basireddy, Linzhu
  Yue, Xhitao Song, Lizhi Yang, Yunhui Liu, Koushil Sreenath, and Sergey
  Levine.
\newblock {GenLoco}: Generalized locomotion controllers for quadrupedal robots.
\newblock In \emph{Conference on Robot Learning}, 2022.

\bibitem[Fu et~al.(2021)Fu, Kumar, Malik, and Pathak]{fu2021minimizing}
Zipeng Fu, Ashish Kumar, Jitendra Malik, and Deepak Pathak.
\newblock Minimizing energy consumption leads to the emergence of gaits in
  legged robots.
\newblock In \emph{Conference on Robot Learning}, 2021.

\bibitem[Fu et~al.(2022{\natexlab{a}})Fu, Cheng, and Pathak]{deep_wholebody}
Zipeng Fu, Xuxin Cheng, and Deepak Pathak.
\newblock Deep whole-body control: Learning a unified policy for manipulation
  and locomotion.
\newblock In \emph{Conference on Robot Learning}, 2022{\natexlab{a}}.

\bibitem[Fu et~al.(2022{\natexlab{b}})Fu, Kumar, Agarwal, Qi, Malik, and
  Pathak]{fu2021coupling}
Zipeng Fu, Ashish Kumar, Ananye Agarwal, Haozhi Qi, Jitendra Malik, and Deepak
  Pathak.
\newblock Coupling vision and proprioception for navigation of legged robots.
\newblock In \emph{IEEE Conference on Computer Vision and Pattern Recognition},
  2022{\natexlab{b}}.

\bibitem[Fu et~al.(2024)Fu, Zhao, and Finn]{mobilealoha}
Zipeng Fu, Tony~Z. Zhao, and Chelsea Finn.
\newblock Mobile aloha: Learning bimanual mobile manipulation with low-cost
  whole-body teleoperation.
\newblock In \emph{{Conference on Robot Learning (CoRL)}}, 2024.

\bibitem[Gehring et~al.(2013)Gehring, Coros, Hutter, Bloesch, Hoepflinger, and
  Siegwart]{control_dynamic_gaits}
Christian Gehring, Stelian Coros, Marco Hutter, Michael Bloesch, Markus~A.
  Hoepflinger, and Roland Siegwart.
\newblock Control of dynamic gaits for a quadrupedal robot.
\newblock In \emph{IEEE International Conference on Robotics and Automation},
  2013.

\bibitem[Ha et~al.(2024)Ha, Gao, Fu, Tan, and Song]{ha2024umionlegs}
Huy Ha, Yihuai Gao, Zipeng Fu, Jie Tan, and Shuran Song.
\newblock {UMI}-on-legs: Making manipulation policies mobile with a
  manipulation-centric whole-body controller.
\newblock In \emph{Conference on Robot Learning}, 2024.

\bibitem[Haarnoja et~al.(2018)Haarnoja, Zhou, Abbeel, and Levine]{sac}
Tuomas Haarnoja, Aurick Zhou, Pieter Abbeel, and Sergey Levine.
\newblock Soft actor-critic: Off-policy maximum entropy deep reinforcement
  learning with a stochastic actor.
\newblock In \emph{International Conference on Machine Learning}, 2018.

\bibitem[Hafner et~al.(2022)Hafner, Lee, Fischer, and Abbeel]{hafner2022deep}
Danijar Hafner, Kuang-Huei Lee, Ian Fischer, and Pieter Abbeel.
\newblock Deep hierarchical planning from pixels.
\newblock In \emph{Advances in Neural Information Processing Systems}, 2022.

\bibitem[He et~al.(2016)He, Zhang, Ren, and Sun]{he2016residual}
Kaiming He, Xiangyu Zhang, Shaoqing Ren, and Jian Sun.
\newblock {Deep Residual Learning for Image Recognition}.
\newblock In \emph{IEEE Conference on Computer Vision and Pattern Recognition},
  2016.

\bibitem[Huang et~al.(2024)Huang, Loquercio, Kumar, Thakkar, Goldberg, and
  Malik]{huang2024manipulatortail}
Huang Huang, Antonio Loquercio, Ashish Kumar, Neerja Thakkar, Ken Goldberg, and
  Jitendra Malik.
\newblock Manipulator as a tail: Promoting dynamic stability for legged
  locomotion.
\newblock In \emph{IEEE International Conference on Robotics and Automation},
  2024.

\bibitem[Ibrahim et~al.(2020)Ibrahim, Liu, Khan, Yang, Adeli, and
  Yang]{depth_processing}
Mostafa~Mahmoud Ibrahim, Qiong Liu, Rizwan Khan, Jingyu Yang, Ehsan Adeli, and
  You Yang.
\newblock Depth map artefacts reduction: a review.
\newblock \emph{IET Image Processing}, 14:\penalty0 2630--2644, 2020.

\bibitem[Imai et~al.(2022)Imai, Zhang, Zhang, Kierebinski, Yang, Qin, and
  Wang]{imai2022}
Chieko~Sarah Imai, Minghao Zhang, Yuchen Zhang, Marcin Kierebinski, Ruihan
  Yang, Yuzhe Qin, and Xiaolong Wang.
\newblock Vision-guided quadrupedal locomotion in the wild with multi-modal
  delay randomization.
\newblock In \emph{IEEE/RSJ International Conference on Intelligent Robots and
  Systems}, 2022.

\bibitem[Iqbal et~al.(2020)Iqbal, Gao, and Gu]{stable_quad_control}
Amir Iqbal, Yuan Gao, and Yan Gu.
\newblock Provably stabilizing controllers for quadrupedal robot locomotion on
  dynamic rigid platforms.
\newblock \emph{IEEE/ASME Transactions on Mechatronics}, 25\penalty0
  (4):\penalty0 2035--2044, 2020.

\bibitem[James et~al.(2019)James, Wohlhart, Kalakrishnan, Kalashnikov, Irpan,
  Ibarz, Levine, Hadsell, and Bousmalis]{james2019sim}
Stephen James, Paul Wohlhart, Mrinal Kalakrishnan, Dmitry Kalashnikov, Alex
  Irpan, Julian Ibarz, Sergey Levine, Raia Hadsell, and Konstantinos Bousmalis.
\newblock Sim-to-real via sim-to-sim: Data-efficient robotic grasping via
  randomized-to-canonical adaptation networks.
\newblock In \emph{IEEE Conference on Computer Vision and Pattern Recognition},
  2019.

\bibitem[Jeon et~al.(2023)Jeon, Jung, Choi, Kim, and Hwangbo]{wholebody}
Seunghun Jeon, Moonkyu Jung, Suyoung Choi, Beomjoon Kim, and Jemin Hwangbo.
\newblock Learning whole-body manipulation for quadrupedal robot.
\newblock \emph{CoRR}, arXiv:2308.16820, 2023.

\bibitem[Ji et~al.(2023)Ji, Margolis, and Agrawal]{ji2023dribble}
Yandong Ji, Gabriel~B Margolis, and Pulkit Agrawal.
\newblock {DribbleBot}: Dynamic legged manipulation in the wild.
\newblock In \emph{IEEE International Conference on Robotics and Automation},
  2023.

\bibitem[Kumar et~al.(2021)Kumar, Fu, Pathak, and Malik]{kumar2021rma}
Ashish Kumar, Zipeng Fu, Deepak Pathak, and Jitendra Malik.
\newblock {RMA}: Rapid motor adaptation for legged robots.
\newblock In \emph{Robotics: Science and Systems}, 2021.

\bibitem[Laskey et~al.(2017)Laskey, Lee, Fox, Dragan, and Goldberg]{dart}
Michael Laskey, Jonathan Lee, Roy Fox, Anca~D. Dragan, and Ken Goldberg.
\newblock {DART:} noise injection for robust imitation learning.
\newblock In \emph{Conference on Robot Learning}, 2017.

\bibitem[Lin et~al.(2024)Lin, Liu, Yang, Niu, Yu, Zhang, Tan, Boots, and
  Zhao]{locoman}
Changyi Lin, Xingyu Liu, Yuxiang Yang, Yaru Niu, Wenhao Yu, Tingnan Zhang, Jie
  Tan, Byron Boots, and Ding Zhao.
\newblock {LocoMan}: Advancing versatile quadrupedal dexterity with lightweight
  loco-manipulators.
\newblock \emph{CoRR}, arXiv:2403.18197, 2024.

\bibitem[Liu et~al.(2024)Liu, Chen, Cheng, Ji, Yang, and Wang]{VBC}
Minghuan Liu, Zixuan Chen, Xuxin Cheng, Yandong Ji, Ruihan Yang, and Xiaolong
  Wang.
\newblock Visual whole-body control for legged loco-manipulation.
\newblock In \emph{Conference on Robot Learning}, 2024.

\bibitem[Long et~al.(2024)Long, Wang, Li, Cao, Gao, and Pang]{HIM}
Junfeng Long, ZiRui Wang, Quanyi Li, Liu Cao, Jiawei Gao, and Jiangmiao Pang.
\newblock The {HIM} solution for legged locomotion: Minimal sensors, efficient
  learning, and substantial agility.
\newblock In \emph{International Conference on Learning Representations}, 2024.

\bibitem[Lyle et~al.(2022)Lyle, Rowland, and Dabney]{capacity_loss}
Clare Lyle, Mark Rowland, and Will Dabney.
\newblock Understanding and preventing capacity loss in reinforcement learning.
\newblock In \emph{International Conference on Learning Representations}, 2022.

\bibitem[Lyle et~al.(2023)Lyle, Zheng, Nikishin, Avila~Pires, Pascanu, and
  Dabney]{understanding_plasticity}
Clare Lyle, Zeyu Zheng, Evgenii Nikishin, Bernardo Avila~Pires, Razvan Pascanu,
  and Will Dabney.
\newblock Understanding plasticity in neural networks.
\newblock In \emph{International Conference on Machine Learning}, 2023.

\bibitem[Ma et~al.(2022)Ma, Farshidian, Miki, Lee, and Hutter]{ma2022mpc}
Yuntao Ma, Farbod Farshidian, Takahiro Miki, Joonho Lee, and Marco Hutter.
\newblock Combining learning-based locomotion policy with model-based
  manipulation for legged mobile manipulators.
\newblock \emph{IEEE Robotics and Automation Letters}, 7\penalty0 (2):\penalty0
  2377--2384, 2022.

\bibitem[Margolis et~al.(2022)Margolis, Yang, Paigwar, Chen, and
  Agrawal]{margolisyang2022rapid}
Gabriel Margolis, Ge~Yang, Kartik Paigwar, Tao Chen, and Pulkit Agrawal.
\newblock Rapid locomotion via reinforcement learning.
\newblock In \emph{Robotics: Science and Systems}, 2022.

\bibitem[Margolis and Agrawal(2022)]{walk_these_ways}
Gabriel~B Margolis and Pulkit Agrawal.
\newblock Walk these ways: Tuning robot control for generalization with
  multiplicity of behavior.
\newblock In \emph{Conference on Robot Learning}, 2022.

\bibitem[Margolis et~al.(2023)Margolis, Fu, Ji, and
  Agrawal]{margolis2023active}
Gabriel~B Margolis, Xiang Fu, Yandong Ji, and Pulkit Agrawal.
\newblock Learning to see physical properties with active sensing motor
  policies.
\newblock In \emph{Conference on Robot Learning}, 2023.

\bibitem[Pan et~al.(2024)Pan, Ben, Yuan, Jiang, Ji, Pang, Liu, and
  Xu]{pan2024roboduet}
Guoping Pan, Qingwei Ben, Zhecheng Yuan, Guangqi Jiang, Yandong Ji, Jiangmiao
  Pang, Houde Liu, and Huazhe Xu.
\newblock {RoboDuet}: A framework affording mobile-manipulation and
  cross-embodiment.
\newblock \emph{CoRR}, arXiv:2403.17367, 2024.

\bibitem[Perez et~al.(2018)Perez, Strub, de~Vries, Dumoulin, and
  Courville]{film}
Ethan Perez, Florian Strub, Harm de~Vries, Vincent Dumoulin, and Aaron~C.
  Courville.
\newblock {FiLM}: Visual reasoning with a general conditioning layer.
\newblock In \emph{AAAI Conference on Artificial Intelligence}, 2018.

\bibitem[Qiu et~al.(2024)Qiu, Hu, Yang, Song, Fu, Ye, Mu, Yang, Atanasov,
  Scherer, and Wang]{GEFF}
Ri-Zhao Qiu, Yafei Hu, Ge~Yang, Yuchen Song, Yang Fu, Jianglong Ye, Jiteng Mu,
  Ruihan Yang, Nikolay Atanasov, Sebastian Scherer, and Xiaolong Wang.
\newblock Learning generalizable feature fields for mobile manipulation.
\newblock \emph{CoRR}, arXiv:2403.07563, 2024.

\bibitem[Qureshi et~al.(2024)Qureshi, Garg, Yandun, Held, Kantor, and
  Silwal]{qureshi2024splatsimzeroshotsim2realtransfer}
Mohammad~Nomaan Qureshi, Sparsh Garg, Francisco Yandun, David Held, George
  Kantor, and Abhisesh Silwal.
\newblock {SplatSim}: Zero-shot sim2real transfer of rgb manipulation policies
  using gaussian splatting.
\newblock \emph{CoRR}, arXiv:2409.10161, 2024.

\bibitem[Ronneberger et~al.(2015)Ronneberger, Fischer, and Brox]{unet}
Olaf Ronneberger, Philipp Fischer, and Thomas Brox.
\newblock {U-Net}: Convolutional networks for biomedical image segmentation.
\newblock In \emph{Medical Image Computing and Computer-Assisted Intervention},
  2015.

\bibitem[Ross and Bagnell(2010)]{imitation_learning_drift}
Stephane Ross and Drew Bagnell.
\newblock Efficient reductions for imitation learning.
\newblock In \emph{International Conference on Artificial Intelligence and
  Statistics}, 2010.

\bibitem[Rudin et~al.(2021)Rudin, Hoeller, Reist, and Hutter]{walk_in_minutes}
Nikita Rudin, David Hoeller, Philipp Reist, and Marco Hutter.
\newblock Learning to walk in minutes using massively parallel deep
  reinforcement learning.
\newblock In \emph{Conference on Robot Learning}, 2021.

\bibitem[Schulman et~al.(2017)Schulman, Wolski, Dhariwal, Radford, and
  Klimov]{ppo}
John Schulman, Filip Wolski, Prafulla Dhariwal, Alec Radford, and Oleg Klimov.
\newblock Proximal policy optimization algorithms.
\newblock \emph{CoRR}, arXiv:1707.06347, 2017.

\bibitem[Sleiman et~al.(2023)Sleiman, Farshidian, and
  Hutter]{sleiman2023wholebody_control}
Jean-Pierre Sleiman, Farbod Farshidian, and Marco Hutter.
\newblock Versatile multicontact planning and control for legged
  loco-manipulation.
\newblock \emph{Science Robotics}, 8\penalty0 (81):\penalty0 eadg5014, 2023.

\bibitem[Sun et~al.(2022)Sun, Zhang, Xu, and Tomizuka]{PaCo}
Lingfeng Sun, Haichao Zhang, Wei Xu, and Masayoshi Tomizuka.
\newblock {PaCo}: Parameter-compositional multi-task reinforcement learning.
\newblock In \emph{Advances in Neural Information Processing Systems}, 2022.

\bibitem[Tobin et~al.(2017)Tobin, Fong, Ray, Schneider, Zaremba, and
  Abbeel]{tobin2017}
Josh Tobin, Rachel Fong, Alex Ray, Jonas Schneider, Wojciech Zaremba, and
  Pieter Abbeel.
\newblock Domain randomization for transferring deep neural networks from
  simulation to the real world.
\newblock In \emph{IEEE/RSJ International Conference on Intelligent Robots and
  Systems}, 2017.

\bibitem[Todorov et~al.(2012)Todorov, Erez, and Tassa]{mujoco}
Emanuel Todorov, Tom Erez, and Yuval Tassa.
\newblock {MuJoCo}: A physics engine for model-based control.
\newblock In \emph{International Conference on Intelligent Robots and Systems},
  2012.

\bibitem[Ulyanov et~al.(2016)Ulyanov, Vedaldi, and Lempitsky]{instancenorm}
Dmitry Ulyanov, Andrea Vedaldi, and Victor~S. Lempitsky.
\newblock Instance normalization: The missing ingredient for fast stylization.
\newblock \emph{CoRR}, arXiv:1607.08022, 2016.

\bibitem[Vaswani et~al.(2017)Vaswani, Shazeer, Parmar, Uszkoreit, Jones, Gomez,
  Kaiser, and Polosukhin]{transformer}
Ashish Vaswani, Noam Shazeer, Niki Parmar, Jakob Uszkoreit, Llion Jones,
  Aidan~N Gomez, {\L}ukasz Kaiser, and Illia Polosukhin.
\newblock Attention is all you need.
\newblock In \emph{Advances in Neural Information Processing Systems}, 2017.

\bibitem[Wu et~al.(2024{\natexlab{a}})Wu, Chong, Holmberg, Prasad, Gao, Khatib,
  Song, Rusinkiewicz, and Bohg]{wu2024tidybot}
Jimmy Wu, William Chong, Robert Holmberg, Aaditya Prasad, Yihuai Gao, Oussama
  Khatib, Shuran Song, Szymon Rusinkiewicz, and Jeannette Bohg.
\newblock Tidybot++: An open-source holonomic mobile manipulator for robot
  learning.
\newblock In \emph{Conference on Robot Learning}, 2024{\natexlab{a}}.

\bibitem[Wu et~al.(2024{\natexlab{b}})Wu, Fu, Cheng, Wang, and
  Finn]{wu2024helpfuldoggybot}
Qi~Wu, Zipeng Fu, Xuxin Cheng, Xiaolong Wang, and Chelsea Finn.
\newblock {Helpful} {DoggyBot}: Open-world object fetching using legged robots
  and vision-language models.
\newblock \emph{CoRR}, arXiv:2410.00231, 2024{\natexlab{b}}.

\bibitem[Yang et~al.(2024)Yang, Chen, Ma, Zheng, Chen, Nguyen, and
  Wang]{yang2023generalized}
Ruihan Yang, Zhuoqun Chen, Jianhan Ma, Chongyi Zheng, Yiyu Chen, Quan Nguyen,
  and Xiaolong Wang.
\newblock Generalized animal imitator: Agile locomotion with versatile motion
  prior.
\newblock In \emph{Conference on Robot Learning}, 2024.

\bibitem[Yarats et~al.(2021)Yarats, Kostrikov, and Fergus]{drqv1}
Denis Yarats, Ilya Kostrikov, and Rob Fergus.
\newblock Image augmentation is all you need: Regularizing deep reinforcement
  learning from pixels.
\newblock In \emph{International Conference on Learning Representations}, 2021.

\bibitem[Yarats et~al.(2022)Yarats, Fergus, Lazaric, and Pinto]{drqv2}
Denis Yarats, Rob Fergus, Alessandro Lazaric, and Lerrel Pinto.
\newblock Mastering visual continuous control: Improved data-augmented
  reinforcement learning.
\newblock In \emph{International Conference on Learning Representations}, 2022.

\bibitem[Yokoyama et~al.(2023)Yokoyama, Clegg, Truong, Undersander, Yang,
  Arnaud, Ha, Batra, and Rai]{ASC}
Naoki Yokoyama, Alexander~William Clegg, Joanne Truong, Eric Undersander, Jimmy
  Yang, Sergio Arnaud, Sehoon Ha, Dhruv Batra, and Akshara Rai.
\newblock {{ASC}: Adaptive Skill Coordination for Robotic Mobile Manipulation}.
\newblock \emph{IEEE Robotics and Automation Letters}, 2023.

\bibitem[Yu et~al.(2024)Yu, Yang, Choi, Ravan, Leonard, and
  Isola]{yu2024learning}
Alan Yu, Ge~Yang, Ran Choi, Yajvan Ravan, John Leonard, and Phillip Isola.
\newblock Learning visual parkour from generated images.
\newblock In \emph{Conference on Robot Learning}, 2024.

\bibitem[Yu et~al.(2022)Yu, Zhang, and Xu]{yu2022you}
Haonan Yu, Haichao Zhang, and Wei Xu.
\newblock Do you need the entropy reward (in practice)?
\newblock \emph{CoRR}, arXiv:2201.12434, 2022.

\bibitem[Yu et~al.(2019)Yu, Quillen, He, Julian, Hausman, Finn, and
  Levine]{metaworld}
Tianhe Yu, Deirdre Quillen, Zhanpeng He, Ryan Julian, Karol Hausman, Chelsea
  Finn, and Sergey Levine.
\newblock Meta-world: A benchmark and evaluation for multi-task and meta
  reinforcement learning.
\newblock In \emph{Conference on Robot Learning}, 2019.

\bibitem[Zhai et~al.(2024)Zhai, Huang, Hu, Li, Bao, and Zhang]{nis_slam}
Hongjia Zhai, Gan Huang, Qirui Hu, Guanglin Li, Hujun Bao, and Guofeng Zhang.
\newblock {NIS-SLAM}: Neural implicit semantic {RGB-D} {SLAM} for 3d consistent
  scene understanding.
\newblock \emph{IEEE Transactions on Visualization and Computer Graphics},
  pages 1--11, 2024.

\bibitem[Zhang et~al.(2023)Zhang, Xu, and Yu]{pex}
Haichao Zhang, Wei Xu, and Haonan Yu.
\newblock Policy expansion for bridging offline-to-online reinforcement
  learning.
\newblock In \emph{International Conference on Learning Representations}, 2023.

\bibitem[Zhang et~al.(2024{\natexlab{a}})Zhang, Gireesh, Wang, Fang, Xu, and
  Chen]{GAMMA}
Jiazhao Zhang, Nandiraju Gireesh, Jilong Wang, Xiaomeng Fang, Chaoyi Xu, and
  Weiguang Chen.
\newblock {GAMMA}: Graspability-aware mobile manipulation policy learning based
  on online grasping pose fusion.
\newblock In \emph{IEEE International Conference on Robotics and Automation},
  2024{\natexlab{a}}.

\bibitem[Zhang et~al.(2024{\natexlab{b}})Zhang, Ma, Miki, and
  Hutter]{zhang2024learning}
Mike Zhang, Yuntao Ma, Takahiro Miki, and Marco Hutter.
\newblock Learning to open and traverse doors with a legged manipulator.
\newblock In \emph{Conference on Robot Learning}, 2024{\natexlab{b}}.

\bibitem[Zhuang et~al.(2023)Zhuang, Fu, Wang, Atkeson, Schwertfeger, Finn, and
  Zhao]{robot_parkour}
Ziwen Zhuang, Zipeng Fu, Jianren Wang, Christopher~G Atkeson, S{\"o}ren
  Schwertfeger, Chelsea Finn, and Hang Zhao.
\newblock Robot parkour learning.
\newblock In \emph{Conference on Robot Learning}, 2023.

\end{thebibliography}

\clearpage
\appendices  
\renewcommand{\thefigure}{A.\arabic{figure}}

\section*{Table of Contents for Appendix}
\startcontents[sections]
\printcontents[sections]{l}{1}{\setcounter{tocdepth}{2}}








\section{RL Formulation}

\subsection{Subtask Boundaries, Rewards and Behavior Priors}\label{appendix:high_level_reward}

\bl{Subtask Boundaries.}
Since we decompose the full long task into a sequence of subtasks,
in practice, we need to decide the task boundaries in order to properly divide the full task.
In the sequential decompositional case as described in Section~\ref{sec:task_decompositions}, the task boundaries $\tau^i |\tau^{i+1}$ between two adjacent subtasks ($\tau^i \rightarrow\tau^{i+1}$) can be defined by the success condition $\mathcal{S}(\tau^i)$ of the subtask $\tau^i$ (i.e., $\tau^i |\tau^{i+1}\equiv \mathcal{S}(\tau^i)$), since the success of a subtask $\tau^i$ naturally leads to the subsequent subtask $\tau^{i+1}$.

We provide the success conditions of each subtask (therefore the subtask boundaries) below:
\begin{enumerate}
\item $\mathcal{S}$(\Search{}): succeeds when the target object is in the view of the wrist-mounted RGB camera;
\item $\mathcal{S}$(\MoveTo{}): succeeds when the target object enters the workspace of the robot within its arm reach;
\item $\mathcal{S}$(\Grasp{}): succeeds when the target object is in the gripper and lifted 5cm above the ground for at least two timesteps;
\item $\mathcal{S}$(\SearchWithObj{}): succeeds when the target object is grasped and the target basket enters the view of the wrist-mounted camera; 
\item $\mathcal{S}$(\MoveToWithObj{}): succeeds when the target object is grasped and the target basket enters a workspace of the robot within its arm reach;
\item $\mathcal{S}$(\MoveGripperToWithObj{}): succeeds when the target object is grasped and the gripper has moved to the top of the target basket;
\item $\mathcal{S}$(\DropInto{}): succeeds when the target object enters the target basket.
\end{enumerate}

\bl{Rewards.} 
We summarize the description of the subtask rewards in Table~\ref{tab:high_level_rewards_and_behavior_prior}.
The \textit{task} reward represents the sparse success ($+1$) or failure ($-1$) reward.
For timeout, we use a reward of $0$.
Apart from the sparse task reward, we also incorporate some shaping rewards, as shown in Table~\ref{tab:high_level_rewards_and_behavior_prior}.
Most of the shaping rewards are delta-distance based, encouraging two
positions of interest to get closer to each other.
Arm retract reward is also an instance of delta-distance based shaping reward, where the distance is measured between the current arm-joint pose (excluding gripper) and the target neutral arm pose (see \Figure~\ref{fig:robot_system} for a reference of the neutral arm pose). Intuitively, the arm retract reward encourages the robot's arm to rise up after grasping the object and move close to its neutral position.
Alignment reward is also an instance of delta-distance based shaping reward, where the distance is defined based on the cosine between the left and right finger-tip-to-object vectors.
Keep-grasping reward is a contact based reward issuing a penalty of $-0.1$ if the contact between gripper fingers and the object is lost.

\begin{table*}[h]
\centering
\caption{Rewards and Behavior Prior for Subtasks}
\begin{tabularx}{17cm}{l l Y}
\toprule
Subtask & Reward & Behavior Prior \\
\midrule
\Search{}   & task & rotational \\
\MoveTo{}    & task + distance & -- \\
\Grasp{}   & task + distance + alignment & stationary \\
\SearchWithObj{}   & task + arm-retract & rotational \\
\MoveToWithObj{}  & task + distance + arm-retract + keep-grasping & -- \\
\MoveGripperToWithObj{}  & task + distance + keep-grasping & stationary\\
\DropInto{}  & task + distance & stationary \\
\bottomrule
\end{tabularx}
\label{tab:high_level_rewards_and_behavior_prior}
\end{table*}

\bl{Behavior Priors.}
Another benefit of task decomposition is that we can easily incorporate different behavior priors to the policy.
We use two types of behavior priors in this work:
\begin{itemize}
    \item stationary manipulation: where both the forward and angular velocity commands are set to zero ($\mathbf c=0$) during manipulation related tasks (\Grasp{}, \MoveGripperToWithObj{}, \DropInto{}), so that manipulation policy can be carried out accurately and safely.
    \item rotational search: the forward velocity command ($\mathbf c[0]=0$) to the quadruped is set to zero, avoiding the quadruped from wandering off the work area during search subtasks (\Search{}, \SearchWithObj{}). 
\end{itemize}

In practice, both priors are implemented in a similar way. For any subtask with either prior, we initialize the teacher's output action distribution to have a zero mean for the corresponding action dimensions (\emph{e.g.}, the forward velocity dimension for rotational prior and both the forward and angular velocity dimensions for stationary prior).
The reason for this design is to accommodate the KL-based distillation step in student learning by always ensuring a valid action distribution, so that subtasks with the priors can compute the KL loss in a consistent way with those without the priors.

\subsection{Hierarchical RL and High-Level RL Training}
As typical in standard hierarchical RL structures, the high-level and low-level policies operate at different frequencies.
We refer to one high-level inference step as a re-planning step.
The high-level re-planning period is $5$, meaning the high-level policy (teacher or student) outputs an action once every $5$ low-level steps, and the same high-level output is used across these $5$ low-level step in the re-planning period for the low-level policy.

Following this structure and given a frozen low-level policy, we are interested in training the high-level (teacher/student) policy.
Because of this, RL training essentially happens at the high-level.
Therefore, the rewards at the low-level steps within a re-planning period are accumulated as the step reward for that high-level step, i.e.,
\begin{equation}
r_{\rm high\_level\_step} = \sum_{i=1}^5 r_{\rm low\_level\_step}^i,
\end{equation}
and this accumulated reward $r_{\rm high\_level\_step}$ is used as the per-step reward signal for high-level RL training.

\subsection{Low-Level Task and Reward}\label{app:low_level}
\bl{Low-Level Task.}
Details of low-level training task, commands, and corresponding rewards are summarized in Table~\ref{tab:low-level_task}.

\bl{Quadraped Gait Control.}
Motivated by \cite{walk_these_ways}, we also introduce gait alignment rewards that incentivize the robot to follow predefined gait patterns. 
The gait patterns are defined by a phase offset vector of length three, where each entry represents the phase difference of each of the three legs (excluding the first one) relative to the first leg (fore left). The range of each entry is $[-0.5, 0.5]$, where $0$ indicates a synchronous movement with the first leg and $-0.5$ or $0.5$ indicates a half cycle phase difference. The phase of a leg is evaluated independently for each leg, based on the leg's schedule at current time. The schedule is a number in $[0, 1]$. For gait alignment rewards design, schedule $1$ represents the ``swing" phase where smaller movements in the X-Y plane will be penalized; schedule $0$ represents the ``contact" phase where lower contact force will be penalized; for any schedule between $(0, 1)$, both aspects will be penalized with weights modulated by the schedule value.

For training of the gait following ability, we expand the input command $\mathbf{c}$ with randomly sampled phase offset vector $\mathbf{g}_{\rm offset}$ and gait cycle vector $\mathbf{g}_{\rm cycle}$. 
$\mathbf{g}_{\rm offset}$ is sampled from five predefined gait patterns, as summarized in Table~\ref{tab:gait_pattern}. $\mathbf{g}_{\rm cycle}$ is a two-dimensional vector, with each dimension representing the frequency and the stance ratio respectively. The frequency takes values in $[2.0, 4.0]$, indicating the number of cycles that each leg goes through per second. The stance ratio takes values in $(0.4, 0.6)$, indicating the portion of time within a cycle when each foot is in contact with the floor.

After low-level training, we fix the gait pattern to be ``trot", i.e., $\mathbf{g}_{\rm offset}\!=\![-0.5, -0.5, 0.0]$, and the gait cycle vector as $\mathbf{g}_{\rm cycle}\!=\![2.4, 0.5]$.  We have found that this gait is suitable for the mobile manipulation task tackled in this work in terms of movement efficiency and stability.
For the stationary task of low-level training, we always set $\mathbf{g}_{\rm cycle}\!=\![0, 1]$.

\begin{table}[t]
\centering
\caption{Low-Level Training Tasks and Rewards}
\resizebox{\columnwidth}{!}{
\begin{tabular}{ c|l|l }
\hline
Task/Command Name & \multicolumn{1}{c|}{Sample Range} & \multicolumn{1}{c}{Reward} \\
\hline
forward velocity & $(-0.5, 1.2)$ m/s & $e^{-\frac{|v_{\text{forward}}-v_{\text{target}}|}{\max(v_{\text{target}}, 0.5)}}$ \\ 
angular velocity & $(-1.0, 1.0)$ rad/s & $e^{-|v_{\text{yaw}}-v_{\text{target}}|}$ \\
stay stationary & with probability $0.1$ & 
$\begin{cases}
      -1 & \text{if any foot lifted}\\
      0 & \text{otherwise}
 \end{cases}  $ \\
balance and survival & 
    \begin{tabular}{@{}l@{}}
      tilt thresh: $0.6$\\
      height thresh: $0.26$
    \end{tabular}
    & 
    \begin{tabular}{@{}l@{}}
        $-(\mathbbm{1}_{\text{over-tilt}} + \mathbbm{1}_{\text{falling}})$ \\
        $-0.3\cdot \mathbbm{1}_{\text{killed}}$
    \end{tabular} \\
\hline
\end{tabular}
}
\label{tab:low-level_task}
\end{table}

\begin{table}[t]
\centering
\caption{Predefined Gait Patterns}
\begin{tabular}{ c|c }
\hline
Gait Names & Phase Offsets \\
\hline
pronk & $(0.0, 0.0, 0.0)$ \\
walk & $(-0.5, 0.25, -0.25)$ \\
trot & $(-0.5, -0.5, 0.0)$ \\
bound & $(0.0, 0.5, 0.5)$ \\
pace & $(-0.5, 0.0, -0.5)$ \\
\hline
\end{tabular}
\label{tab:gait_pattern}
\end{table}

\bl{Low-Level Domain Randomization.}
The domain randomization setting for low-level training is summarized in Table~\ref{tab:env_randomization}. 

\begin{table}[tb]
\centering
\caption{Low-Level Domain Randomization Parameters}
\resizebox{\columnwidth}{!}{
\begin{tabular}{ c|l }
\hline
Env Params & Sample Range \\
\hline
base payload & $(-0.5, 3)$ \\
torso mass center & $(-0.15, 0.15)$ for each dimension \\
feet friction & $(0.25, 1.75)$ \\
feet softness (solimp[2])\footnote{Only the third entry of the solimp parameter of Mujoco is randomized. It is roughly proportional to the size of the deformation of its feet when the robot stands still on the ground.} & $(0.005, 0.02)$ \\ 
griper payload & $(0, 0.1)$ \\
Kp of leg joints & $(0.7, 1.3)$ \\
actuator gain of arm joints & $(0.7, 1.3)$ \\
bumpiness of the ground & $(0, 0.15)$ for $90\%$, $0$ for $10\%$ \\
delay of sensors & $(u-0.002, u+0.002),\ u\in(0.003, 0.013)$ \\
\hline
\end{tabular}
}
\label{tab:env_randomization}
\end{table}

\section{Model Details}
\label{app:model}
\subsection{Teacher (High Level)}
The teacher network  $\Pi\!=\!\{\pi^k\}_{k=1}^K$ is a set of sub-policies (referred to as policy set), as shown in \Figure~\ref{fig:expert_nn}.
The effective size of the policy set is $K\!=\!8$ (including the \Idle{} subtask).

Each policy net $\pi^k$ (for subtask $k$) is implemented as as a 3 layer MLP, with the layer size as $512$ ($[512, 512, 512]$).
A squashed Normal projection layer (a layer for generating Normal distributions) is used as the output layer for each policy net following the standard practice~\cite{sac}, generating the action distribution conditioned on the current input.

The critic net for each subtask uses the same MLP structure and with the size of ($[512, 512, 512]$).
Frame stacking is used with a stacking size of $5$.

The re-planning period is $5$, meaning that the high-level policy predicts once every $5$ low-level steps.

\begin{figure}[t]
    \centering
    \includegraphics[width=0.95\columnwidth]{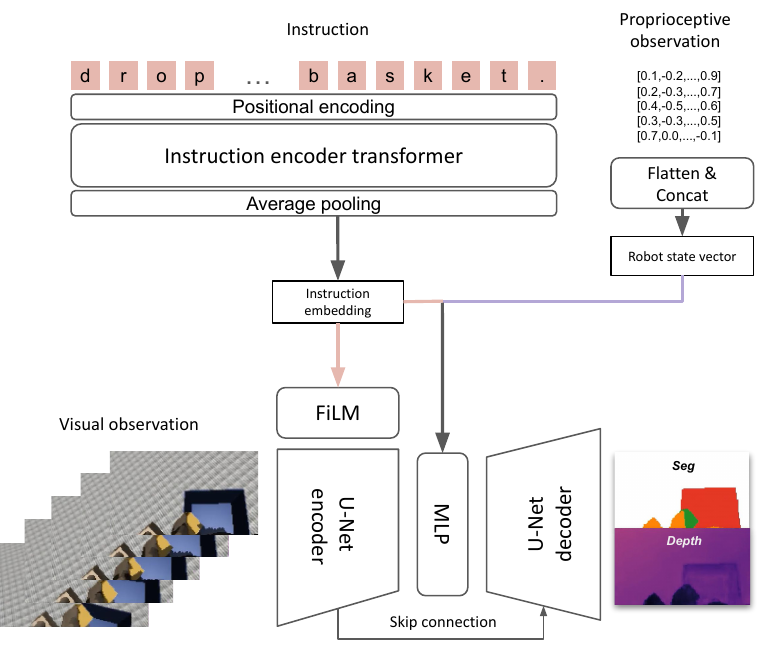}
    \caption{The architecture of the student's U-Net for predicting segmentation and depth maps.}
    \vspace{-0.2in}
    \label{fig:unet_diagram}
\end{figure}

\begin{figure}[t]
    \centering
    \includegraphics[width=0.88\columnwidth]{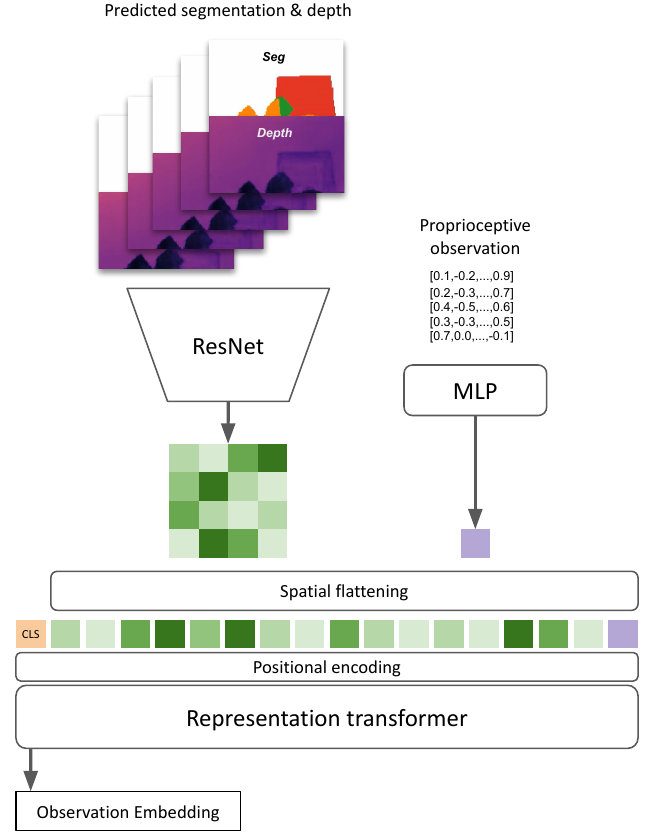}
    \caption{The architecture of the student's policy input encoder that generates a compact observation embedding for RL.}
    \label{fig:repr_diagram}
    \vspace{-0.2in}
\end{figure}

\subsection{Student (High Level)}
\bl{Maps Prediction U-Net.} An illustration of the U-Net~\citep{unet} for predicting segmentation and depth maps is shown in \Figure\ref{fig:unet_diagram}. 
The architecture details will be presented according to input modality below.
\begin{enumerate}
\item \textit{Language.} Due to the simplicity of our language instructions, we directly encode each instruction into a sequence of byte tokens.
We set the max sequence length to be 100, padding shorter sequences with zeros while truncating longer sequences beyond this limit.
Then each byte token is converted into an embedding vector of 128 dimensions by looking up a learnable embedding table.
The sequence of language embeddings, with learnable positional encodings, is fed to an instruction encoder transformer~\citep{transformer} with self-attention.
The transformer has 2 layers, each layer with 8 attention heads. 
Each attention head has a key dimensionality of 64 and a value dimensionality of 64. 
The MLP for generating the residual output has 128 hidden units.
Finally, the output embeddings are average pooled to generate a compact instruction-level embedding of 128 dimensions
\item \textit{Proprioceptive state.} We use a temporal stack of $N=5$. Different proprioceptive observations are simply flattened and concatenated together to produce a 1D robot state vector.
\item \textit{Vision.} We use a temporal stack of $N=5$ RGB images for vision. 
The images are stacked along the channel dimension and fed to a U-Net encoder containing 6 convolutional blocks.
Each block has a convolutional layer $(filters,kernel\_size,strides,padding)\triangleq(c,k,s,p)$, an instance norm layer~\citep{instancenorm}, a ReLU activation, and a 2D max pooling layer (only starting from the second block).
The convolutional layers are configured as $c=[32,64,64,128,128,128]$, and $k=3\times 3,s=1,p=1$ for all blocks.
We use FiLM~\citep{film} conditioned on the instruction embedding to modulate each intermediate convolutional output, before it is fed to the next convolutional layer.
After getting the output from the encoder, we concatenate it with the robot state vector and the instruction embedding, and feed it to the non-skip MLP.
The MLP has two hidden layers of 512 units.
Finally, the output from MLP is projected back to the output space of the encoder, and is fed to the U-Net decoder to produce the predicted map.
The map has $M+1$ channels, where $M=4$ is the predefined maximum number of segmentation classes, and the remaining one channel is for depth prediction.
\end{enumerate}

\noindent The maps prediction U-Net is trained by groundtruth segmentation and depth in simulation.

\bl{Policy Inputs Encoder.} An illustration of the policy input encoder network is shown in \Figure~\ref{fig:repr_diagram}.
The network is responsible to digest predicted segmentation\&depth maps and proprioceptive observations, and produce a compact representational embedding for downstream RL networks.
Again, we use a temporal stack of $N=5$ for both segmentation and depth map inputs.
After converting each segmentation map to a one-hot representation, we stack all maps along the channel dimension, resulting in an input of $5\times(4+1)=25$ channels.
The input is fed to a ResNet~\citep{he2016residual}  of 5 residual blocks, each block with 64 filters, a kernel size of $3\times 3$, and a stride of 2.
The output from the ResNet is spatially flattened, resulting in a sequence of visual tokens.
On the other hand, we use an MLP with one hidden layer of 256 units to project the concatenated proprioceptive observation into a token embedding of 256 dimensions.
This proprioceptive token is appended to the visual token sequence which is fed to a representation transformer with self-attention.
The transformer has 5 layers, each layer with 8 attention heads.
Each attention head has a key dimensionality of 128 and a value dimensionality of 128.
The MLP for generating the residual output has 256 hidden units.
Finally, we take the transformed output corresponding to a special learnable \texttt{CLS} token as the final output of the entire policy input encoder.

\bl{Critic and Actor Networks.} Both the critic and the actor have three hidden layers of $(512,512,512)$. The actor outputs a squashed Gaussian distribution as in SAC~\citep{sac}. Note that the RL and distillation gradients will be propagated all the way back to the policy input encoder for learning the representation.

\bl{Stationary Bit.} The teacher policy incorporates a stationary behavior prior into the locomotion command $\mathbf{c}$ for several subtasks (Table~\ref{tab:high_level_rewards_and_behavior_prior}).
With this prior, the corresponding locomotion command distribution always has a zero mean and a small std.
Taking $\text{argmax}$ of the distribution, the teacher policy generates exactly zero locomotion command and thus the quadruped can remain stationary.
However, if we directly distill this distribution into the student's counterpart, it is almost impossible to guarantee a zero mode, due to the continuous nature of the distribution.
Thus we employed a sationary bit trick to solve this issue.
Specifically, we create an extra binary action distribution for the student for predicting whether the quadruped should be stationary or not, which is determined by checking if the locomotion command of the teacher is exactly zero.
We then add a new prediction loss for learning this stationary bit, while masking out the KLD loss (only for the two locomotion command dimensions) if the stationary bit is true.
For student inference, whenever the stationary bit distribution outputs 1, the locomotion command $\mathbf{c}$ will be overwritten to be zero.

\subsection{Low Level}
\label{app:low-level}
For low-level training, we use PPO~\citep{ppo} with a regularized online adaptation module~\citep{deep_wholebody}.

\bl{Policy Input and Locomotion Command.}
As shown in \Figure~\ref{fig:framework}, the low-level locomotion policy
takes two types of inputs: 1) the standard proprioceptive state $\mathbf s_\textrm{leg}$ obtained from the environment, 2) the locomotion command $\mathbf{c}$ from an external command issuer (\emph{e.g.},
the second type of input is crucial to make the low-level policy useful in downstream tasks where locomotion is involved as a basic skill for exploration and learning.
During training, the command is re-sampled periodically within each episode and the low-level policy is trained to follow the command $\mathbf{c}$ via the command-following rewards detailed in Table~\ref{tab:low-level_task}.
The arm pose is also adjusted periodically within each episode, to make sure the low-level policy can follow locomotion commands while being robust against arm movements.
For the standard proprioceptive input to the low-level policy, while it is also possible to include those from the arm, empirically, we discovered that it might cause the quadruped to shake more often, and decide to only use $\mathbf s_\textrm{leg}$.  We also discovered that the joint velocities in the hardware are delayed more than joint position readings from the motor sensors, possibly due to temporal smoothing.  Since we do not know the exact algorithm that computes the joint velocities, we simply do not use them as input, and rely on feeding the neural networks the previous few frames for the policy to estimate the joint velocities.

After training, the low-level policy can be used to follow commands issued by an external command issuer. This can be either a higher level policy (as in \name{} or all the autonomous baselines), or human teleoperation (\HumanTeleop) in this work.

\bl{Actor and Value Networks.}
Both the actor and value networks have three hidden layers of size $(256, 256, 256)$. The actor network outputs a Beta distribution instead of the typical Gaussian. For numerical stability, we set the minimum value of the shape parameters of the Beta distribution to $1$.

\bl{Online Adaptation Module for \SimtoReal{} Transfer}.
Similar as in~\citep[\Figure 2]{deep_wholebody}, our online adaptation module consists of two encoders, the privileged encoder and the adaptation encoder. The privileged encoder takes privileged information as input and predicts a latent representation of environment extrinsics, while the adaptation encoder predicts the same latent representation depending only on robot sensory observations. Both encoders are trained to match each other and jointly with the RL objective. The privileged encoder network has two hidden layers of $(128, 64)$ and the adaptation encoder has hidden layers of $(256, 128)$.

\subsection{Model Parameter Count}
We summarize in Table~\ref{tab:param_count} the rough number of model parameters for different components of our system, including low level, teacher high level and student high level.

\begin{table}[h]
\centering
\caption{Number of Trainable Parameters.}
\resizebox{0.9\columnwidth}{!}{
\begin{tabular}{ c| c }
\hline
Component & Approx. Param. Count \\
\hline
Low Level & $3$M \\
Teacher High Level (all subtasks) & $10$M \\
Student High Level & $16$M \\
\hline
\end{tabular}
}
\label{tab:param_count}
\end{table}

Our deployed policy (``Low Level'' + ``Student High Level'') has a total parameter count of less than 19M. 

\section{Important Training Hyperparameters}
\label{app:hyperparams}
Below we list some important training hyperparameters of \name{}. 
For other hyperparameters, please refer to our code at \code{}.
All the modules (teacher, student and low-level) were trained with V100 or similar GPUs.

\bl{Teacher (High level).} For training, we used a batch size of $2000$, a learning rate of $1\times 10^{-4}$, gradient clipping of $0.1$ with the Adam optimizer, $120$ parallel environments, and training iterations of $4$M (corresponding to about $60$M environment steps). With four GPUs, the training of the teacher policy takes about 1 week.

\bl{Student (High level).} We used 60 parallel environments to collect rollout data for training the student, where the data was stored in to a replay buffer with a capacity of $600$k environment steps.
With four GPUs, the training of the student policy takes about 2 weeks, with a total of $1.3$M iterations.
This amounts to roughly $40$M environment steps in total.
We used a batch size of $64$, a learning rate of $1\times 10^{-4}$, gradient clipping of $100$, and the Adam optimizer.
For the mixed rollout schedule, we kept the probability $\beta$ of sampling from the student policy as $0.5$ until the $750$K-th iteration, and linearly increased $\beta$ so that it reaches $1.0$ at the $1.25$M-th iteration.
The distillation loss weight $\alpha$ in Equation~\ref{eq:student} was set to $0.01$.
The fixed modal dispersion $\sigma$ for distilling the teacher policy was $0.05$.

\bl{Low level.} We used 5000 parallel environments to roll out and train the low-level policy. With a single GPU, it takes about three days for training the low-level policy with a total of 4B environment steps. We used a batch size of 5000. The Adam optimizer had an initial learning rate of $2\times 10^{-4}$ and a step decay to its $1/5$ after $70\%$ of the total training progress.

\section{Simulator Choice}
All simulations were done in the MuJoCo simulator \citep{mujoco}.  At the beginning of the project, we evaluated several popular simulators at the time, including MuJoCo\footnote{\url{https://mujoco.org/}}, Isaac Gym\footnote{\url{https://developer.nvidia.com/isaac-gym}}, and PyBullet\footnote{\url{https://pybullet.org/wordpress/}}.   
We settled on MuJoCo due to its reasonably fast simulation (on CPU) and rendering (on GPU) at the same time. 

Besides camera observation rendering for student training, due to partial observability, we also rely on segmentation rendering to check which objects are in the camera view and use it as privileged information for the teacher, during both teacher and student training.  
Thus RGB and segmentation rendering efficiencies are critical to our simulator choice.

\section{\SimtoReal{} Techniques}
\label{app:sim2real}

\subsection{Dynamics Gap Reduction Techniques}
\label{app:dynamics_gap}
The dynamics \simtoreal{} gap is the mismatch between the transition function $P(\mathbf s^t | \mathbf s^{t-1}, \mathbf a^{t-1})$ of the simulation versus that of the real world.  It can be due to the mismatch in the physical properties of objects and motors, or by things that are not properly simulated, such as friction and backlash.
This difference in dynamics can cause policies trained purely in simulation to fail when deployed in the real world, especially if the task requires accurate motor control, such as grasping a small object.
We use the following key techniques to address the dynamics gap.

\bl{Arm PID Control to Minimize Tracking Errors.}
An arm joint tracking error $\mathbf e_\textrm{arm}$ is the positional difference between the arm command of the previous timestep and the joint positions of the next timestep,
\[
\mathbf e^t_\textrm{arm} =  \mathbf s^t_{\textrm{arm}} - \mathbf a^{t-1}_{\textrm{arm}}.
\]

Prior works \citep{deep_wholebody} have used PD control for the arm joints.  PD control needs to have large tracking errors, the difference between the target and the current joint positions, to compensate for lifting the arm and any grasped object against gravity.

Intuitively, we know that if the tracking error $\mathbf e_\textrm{arm}$ is equivalent between sim and real for all visited states, then there is essentially no dynamics gap.  
With this, one solution would be to match the tracking errors as closely as possible through accurate embodiment modeling with low-level PD control tuning.
In practice, achieving this alignment for the entire state distribution requires nontrivial work, given the nonlinearity and drifting of motor dynamics due to effects such as motor backlash, wear-and-tear, etc.

Rather than matching non-zero tracking errors, another approach is to simply achieve near-zero tracking errors in both sim and real. 
In other words, whatever arm command is outputted, the resulting arm position should be roughly achieved $\mathbf s^t_{\textrm{arm}} \approx \mathbf a^{t-1}_{\textrm{arm}}$.
In fact, prior studies have shown that tracking error minimization is one of the most reliable strategies for enabling successful \simtoreal{} transfer~\cite{aljalbout2024action_space}.
To do so, we use a PID joint position controller for our manipulator in both simulation and in real, with a small joint position change limit $z\!=\!0.05$ (Section~\ref{sec:obs_and_actions}) to ensure minimal tracking errors for a wide variety of pose configurations. Our tuned PID controller executes smoother and more stable trajectories than those from the PD controller, allowing safe and accurate deployment.

We stay with PD control for the quadruped motors as the hardware driver does not support PID.

\bl{Arm Control Perturbation.}
We add $\pm0.02$ noise to the arm joint position targets uniformly randomly, sampled every 50 high-level control steps or 5 seconds in real time.
This noise is applied to both teacher and student training.  The noise level is increased to $\pm0.05$ after 3M iterations for teacher training.
This is to make the learned policy to be robust \emph{w.r.t.} different sources of control noise arising when transferred to real.

\bl{Arm Mount Perturbation.} We randomly perturb the arm mount position and yaw at the beginning of each episode to make the learned policy robust to \simtoreal{} gaps that may arise from the actual mounting accuracy, torso height variations, \emph{etc}.
In this work, we sample perturbations for $x$, $y$, $z$ and yaw $\theta$ from the following ranges: $x: \pm1$cm, $y: \pm1$cm, $z: \pm1$cm, $\theta: \pm 2\times 10^{-3}$ rad.

\bl{Object Perturbations.} 
Due to the relatively deterministic nature of simulations (in contrast to the high-entropy real world), we observed that policies often converge to deterministic strategies. 
For example, for picking up an object, a policy may learn to always position itself so that the object has the same relative position to its base.
This deterministic ``memorizing" behavior can result in task failure if the conditions for success are even slightly off in real. 
Therefore, we enforce learning robust, reactive policies by perturbing objects of interest during various stages of task execution.
Empirically, this helps with both the exploration (as the probability of stuck in a local state is reduced) and the robustness of the policy (as the state coverage is enlarged).

We perturb the object within a circle with a radius of $r_{\rm perturb}$ centered around the current position of the object.
We randomly sample a position within this circle and check its eligibility.
If it touches or within the space of robot torso projected on to the ground plane, then this perturbation position is regarded as invalid. Otherwise it is regarded as valid.
If a valid perturbation is found within 10 trails, then we teleport the object to the new location as a form of object perturbation.
Otherwise, we keep the position of the object unchanged.
In this work, $r_{\rm perturb}\!=\!10$cm is used.

\bl{Stationary Manipulation.} We make the quadruped stand as still as possible when manipulating objects, \emph{i.e.}, during \emph{Grasp} or \emph{DropInto}.  This avoids the large \simtoreal{} gap of a walking and shaking quadruped when accurate manipulation is needed.  To achieve this, when training low-level policies to follow locomotion commands, we also encourage the quadruped to enter a stationary mode with a balanced stance when the speed command is close to zero.  Without stationary manipulation, not only will success rates suffer, but also it becomes unsafe to deploy, because the gripper fingers can hit the ground, damaging fingers, arm motors and the surrounding.

\bl{Reducing Quadruped Shaking.} When transitioning from locomotion to stationary, the quadruped with the arm on top can shake, sometimes quite violently, before settling.  This only happens in real deployment but rarely in simulation.  Our experiments seem to show that hiding the arm command and observations from the low-level locomotion policy during training and deployment reduces most of the shaking. For the experiments in this paper, we ran low-level training for ten random seeds, and chose three of the best ones (less shaking and more stable for the stationary pose) to continue high-level policy training and evaluation.

\bl{Dynamics \SimtoReal{} Gaps Unaddressed.} Unlike prior work, we did not explicitly model the motor dynamics, and thus the effects of backlash and motor operating near torque limit boundaries are not modeled.
For the leg motors of Go1, we follow existing work which uses P control plus damping to mimic the physical PD controller.  Damping is applied to the absolute joint velocity of the motor, while the D factor in PD control is applied to the difference between the targeted and the actual joint velocities.

\subsection{Visual Gap Reduction Techniques}
\label{app:visual_gap}

The second source of \simtoreal{} gap for image-conditioned policies is the visual gap.
This gap is caused by a visual distribution mismatch between simulated and real-world pixels.
Since we do not assume knowing the target scenes in advance, we have to ensure that the perception model is able to handle a wide range of visual scenes.
Accordingly, we apply various visual \simtoreal{} gap reduction techniques to student training, in addition to the visual information bottleneck in Section~\ref{sec:student}

\bl{Image Domain Randomization.}
Inspired by SECANT~\citep{fan2021secant}, we apply pixel-level perturbations to every RGB image to increase the robustness of the perception model and the chance of its successful \simtoreal{} transfer.
Specifically, we will randomly select a transformation from the list (\textit{Gaussian Noise}, \textit{SaltPepper Noise}, \textit{Speckle Noise}, \textit{Random Brightness}, \textit{Gaussian Blur}) and apply it to each RGB image. 
We choose not to introduce higher-level image randomization/generation~\citep{yu2024learning} because it might impact image semantics.

\bl{Random Textures.}
We spawn the robot on a flat ground plane and randomize the ground texture. 
The texture can be arbitrary and generally has no specific semantics, as long as they provide enough visual distraction to the student.
This distraction will force the student to learn to focus on the most important visual features that are related to the task at hand: objects with reasonable sizes and target colors.
When deployed in real, we hope that this visual focus will get transferred and help the robot locate objects.

\bl{Random Spatial Augmentation.}
To make the visual representation module robust, we incorporate spatial augmentations in the visual representation learning~\citep{drqv1,drqv2}.

\begin{figure}[t]
    \centering
    \resizebox{\columnwidth}{!}{
        \begin{tabular}{@{}c@{\hspace{1mm}}c@{}}
            \includegraphics[width=0.4\columnwidth]{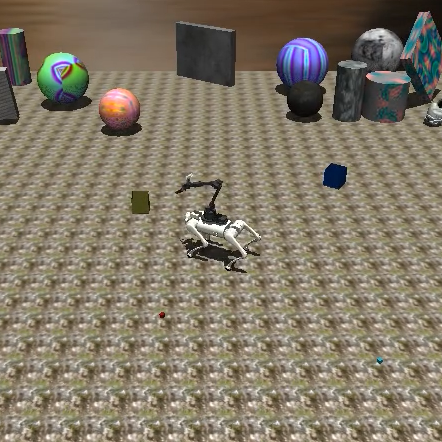}&
            \includegraphics[width=0.4\columnwidth]{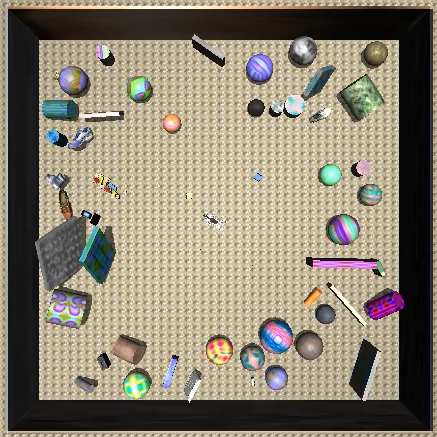}\\
        \end{tabular}
    }
    \vspace{-0.05in}
    \caption{\textbf{Example Training Scene with Randomized Objects.} Left: isometric view; Right: birds-eye view.
    }
    \vspace{-0.15in}
    \label{fig:rand_background_objs}
\end{figure}

\bl{Random Background Objects.}
To further increase the visual complexity of the surrounding environment, we randomly sample and spawn a set of background objects in the room around the robot's initial position (Figure~\ref{fig:rand_background_objs}).
There are two categories of background objects: 1) primitive shapes (cuboid, cylinder, and sphere) and 2) Google scanned objects~\citep{downs2022google}.
We make sure that these objects only serve the purpose of distraction and the student will not confuse them with task objects.
It could collide with an background object during training with some penalty reward. 
When spawning an object, we also randomize its size, orientation, and texture (only for primitive shapes).

\bl{No Camera Depth Data.} For our setting, language-guided visual recognition in the RGB space is the key to finish the task.
Without pretrained vision models, this \simtoreal{} transfer in RGB has to be supported by our method.
We choose not to use camera depth data because this will introduce an extra \simtoreal{} gap, which has to be reduced with denoising or hole-filling techniques~\citep{depth_processing}.
We observe that depth can be reasonably estimated from stacked RGB images, even with monocular vision.

\begin{figure}[t]
    \centering
    \includegraphics[width=0.90\columnwidth]{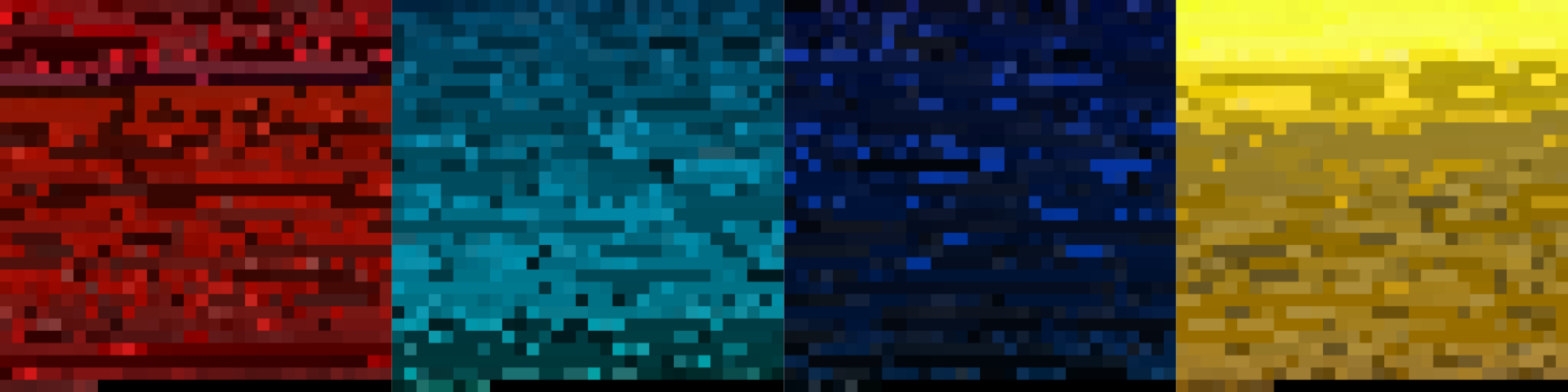}
    \caption{\textbf{Color samples.} Each color had 1000 randomly sampled pixels used to estimate the distribution in simulation. }
    \label{fig:color_calibration}
\end{figure}

\bl{Color Modeling.}
Visual \simtoreal{}, especially for colors, is notoriously difficult~\citep{qureshi2024splatsimzeroshotsim2realtransfer}.  
We simplify the problem by only using four distinct colors in our experiments, namely, red, green (more like cyan), blue and yellow.  
These colors are sampled from real objects in a handful of lighting conditions and are randomized in the HSV space during simulation training.
To obtain the randomization range, we randomly sample 1000 pixels for each color, as shown in \Figure~\ref{fig:color_calibration}.
Afterwards, we compute the means and standard deviations of the samples.
HSV values are then sampled during training from a uniform distribution with a matching mean and $\pm$ bounds that cover the the sampled lighting conditions.

\section{Robot Implementation Details}

\subsection{Control Latency Management}
\label{sec:latency}
The robot performs high level inference at 10\,HZ and low level inference at 50\,HZ.
In order to satisfy the time constraints, we measured the latency along the full control cycle, including motor sensor reading ($\sim$3\,ms), camera observation (30$\sim$40\,ms), network connection ($\sim$0.2\,ms with cable and $\sim$5\,ms on Wi-Fi), and model inference (20$\sim$40\,ms for the high level~\footnote{Later with TensorRT, we were able to reduce this latency to around 10\,ms. However, the experiments in this paper were conducted without this option back then.}, and $\sim$5\,ms for the low level).  
We implement the following methods to address such delays as much as possible, while properly modeling them in our simulator.

\begin{enumerate}
\item Asynchronous low-high inference: most of the model latency comes from the high-level inference, which takes around 20\,ms, but can sometimes peak to 40\,ms. This could delay the low-level policy (50\,HZ) considerably and make the robot falter, if the two levels are executed in a synchronous mode. Alternatively, in the asynchronous mode, the low-level policy will reuse the previous locomotion command if the new one has not arrived yet, and thus still maintain its 50\,HZ frequency. This leads to a more stable gait of the quadruped. 
\item Ethernet cable connection to the robot: Wi-Fi connection is more convenient than cable.  However, we find the Wi-Fi connection between Go1 and the laptop to be unstable, sometimes spiking to a latency of 100\,ms, which is unacceptable for low-level motor control.  It is possible to have a Wi-Fi solution with more powerful and stable hardware, or to run low-level inference on Go1 instead of over Wi-Fi, but we simply rely on cable connection in this work.
\item Sensor delay randomizations: the mean of motor sensor reading delays is sampled within $(3,13)$\,ms every episode, with $\pm$\,2ms variation sampled uniformly per control step. Camera observation delay is sampled from $(30,50)$\,ms uniformly per step with observations taken at 60\,HZ.  Low-level control delay is sampled from $(5,7)$\,ms uniformly per step. 
\item High-level inference delay: we trained policies without high-level inference delay, and compared the performance with adding additional test-time delay.
We did not observe any performance drop with the added delay, and simply train the high level without inference delay in this work.

\end{enumerate}

\begin{figure}[t]
    \centering
    \includegraphics[width=0.80\columnwidth]{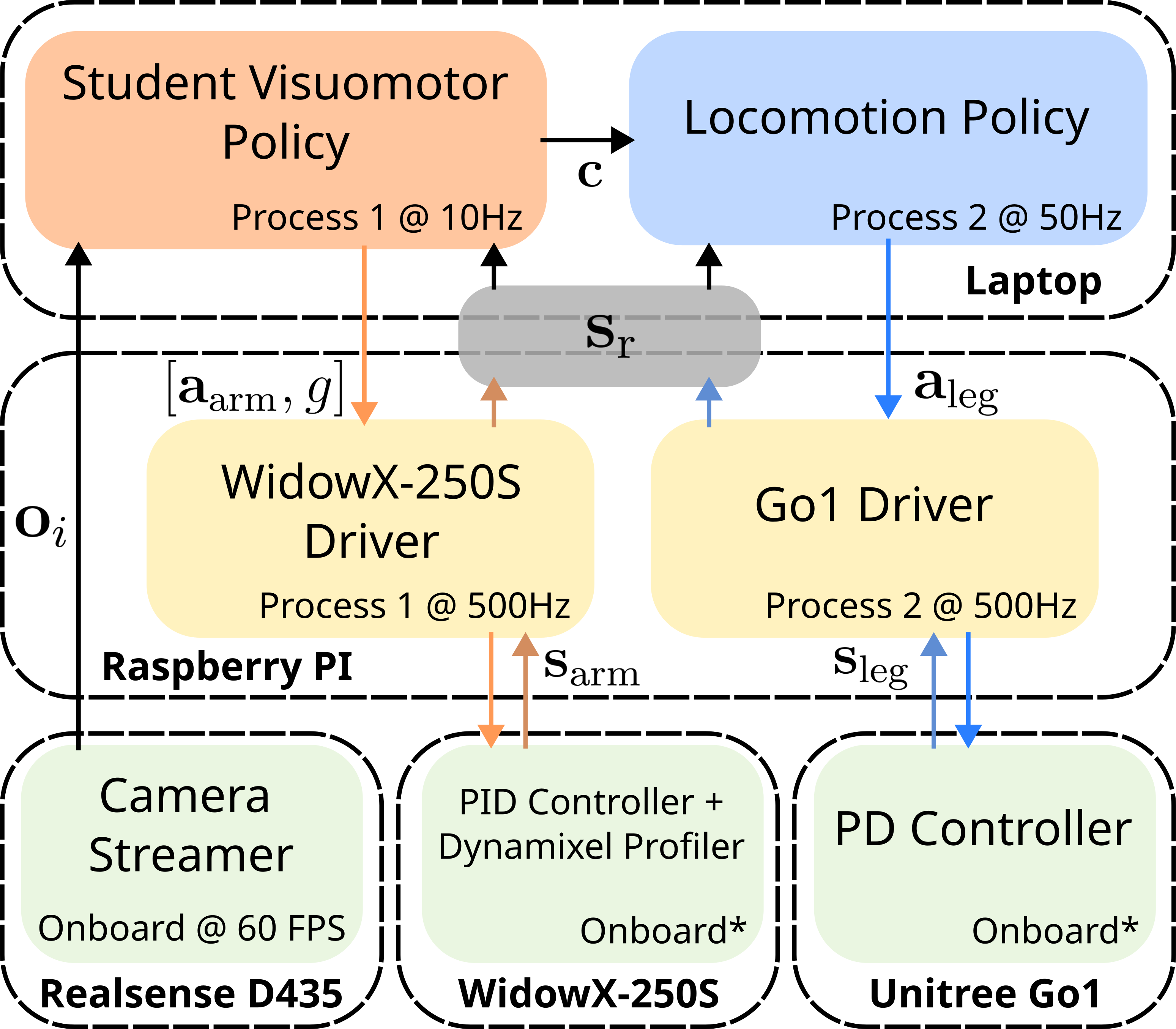}
    \caption{\textbf{Asynchronous System Architecture.} Note that ``Onboard*" refers to the low-level controller frequency being unknown due to the use of closed-source software. Furthermore, when using the WidowX's PID controller, we also take advantage of Dynamixel's onboard profiling for smooth control.}
    \label{fig:system_architecture}
\end{figure}

\subsection{Safety Measures for the Physical Robot}
To reduce damages to surroundings and the robot itself, we employed the following safety measures.

\begin{enumerate}
\item We first addressed all \simtoreal{} gaps of the arm alone by training and deploying a tabletop policy.
Similarly, we trained and deployed a standalone quadruped policy, without the mounted arm.
Only after the two standalone policies were transferred well, we started training and deploying the complete \name{} system.
This helped us avoid potential catastrophic failures from the first-time deployment of a complex quadruped manipulation system.
\item We used PID control for the arm (Section~\ref{app:dynamics_gap}), with a reasonably small delta action $z=0.05$, resulting in much smoother and more reliable arm movement.
This strategy reduced damages to the arm motor gears significantly.
\item The randomizations of motor strength, payload, center of mass, arm mounting position, \emph{etc.} (Section~\ref{app:dynamics_gap}), though making simulation training more difficult, actually allowed the policy to adapt to the real world more easily, resulting in very accurate grasping executions.
Thus the arm did not push into the ground or the target objects due to dynamics mismatch, avoiding motor damages.
\item As high-level inference can sometimes take over 40ms to finish, we adopted asynchronous low-high inference (\Figure~\ref{fig:system_architecture}) to make sure that the high-level inference did not prevent the robot from receiving closed-loop low-level commands.
Otherwise, the quadruped could topple over and damage components mounted on top.
\item We also enabled deployment termination with a single button press.
When latency spikes, especially when the deployment machine suddenly freezes or the network disconnects, our driver on Go1 will automatically stop the quadruped. 
This preventative stop was done using a scripted policy and it prevented damages to the hardware in most cases.
\item To prevent the robot from wandering off and bumping into surroundings, during training we incorporated a rotational search behavior as explained in Section~\ref{appendix:high_level_reward}.
This was done for all baseline methods and \name{}.
\item We attached a leash to the quadruped to prevent it falling down due to any reason.
Later when the policy became more reliable in real, we removed the leash.
\item Our 3D-printed camera mount (\Figure~\ref{fig:robot_system}) was designed to break easily, sacrificing itself to protect the camera in the event of a side fall.
\item During deployment, we frequently monitored the temperature of Go1's motors and took a break if necessary.
A very hot Go1 motor will result in unpredictable, dangerous locomotion behaviors.
\end{enumerate}

\section{Real-World Experiment Details}
\subsection{Containers and Graspable Objects}\label{app:physical_objects}
\begin{figure}[h]
    \centering
    \begin{overpic}[width=8cm]{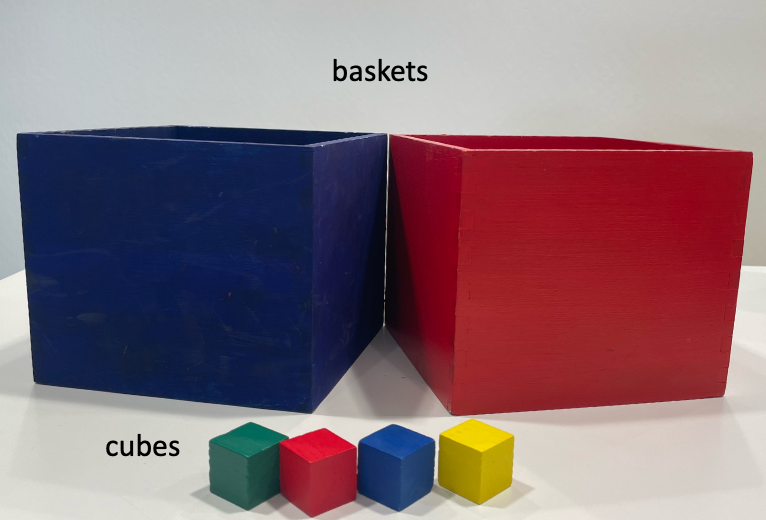}
    \end{overpic}
    \vspace{-0.05in}
    \caption{\textbf{Containers and Graspable Objects.}
    }
    \label{fig:containers_and_objects}
\end{figure}

We use cubes of different colors as graspable objects, and baskets of different colors as the target container for dropping cubes in, as show in \Figure~\ref{fig:containers_and_objects}.
The size of the basket is about $16{\rm cm}\!\times\!16{\rm cm}\!\times\!16{\rm cm}$.
The size of the cube is about $2.5{\rm cm}\!\times\!2.5{\rm cm}\!\times\!2.5{\rm cm}$ and the weight of the cube is about $9{\rm g}$.

\subsection{Real-world Evaluation Protocols} \label{app:eval_protocol}
We detail the protocols used for real-world quantitative evaluation in the main experiment (Table~\ref{tab:real_world_experimental_results}).

\begin{table}[t]
\centering
\caption{Scene Setups for Real-World Quantitative Evaluations. Legends: $\uparrow$ denotes the position robot with its orientation aligned with the direction of $\uparrow$. Uppercase letters denote baskets and lower case letters denote the objects to be grasped. The color of an object is denoted both by the background color and the character (\emph{e.g.,}  \fcolorbox{green}{green}{\textcolor{black}{g}} denotes the green cube and \fcolorbox{red}{red}{\textcolor{white}{R}} denotes the red basket.)}
\begin{tabularx}{7cm}{c|c}
\toprule
Scene setup  & Instruction: \emph{"Drop x into Y"} (x $\mapsto$ Y)  \\
\midrule
\multirow{4}{*}{
\begin{tikzpicture}
    \draw (0.2, 0.2) rectangle (1, 1);
    \node[fill=blue, text=white] at (0.2, 1) {B};
    \node[fill=red, text=white] at (1, 1) {R}; 
    \node[fill=green, text=black] at (0.2, 0.2) {g};
    \node[fill=red, text=white] at (1, 0.2) {r};
    \node at (0.6, 0.6) {$\uparrow$};
\end{tikzpicture}
}   & \fcolorbox{green}{green}{\textcolor{black}{g}}
    $\mapsto$ \fcolorbox{red}{red}{\textcolor{white}{R}}   \\
   & \fcolorbox{green}{green}{\textcolor{black}{g}}
    $\mapsto$ \fcolorbox{blue}{blue}{\textcolor{white}{B}}    \\
   & \fcolorbox{red}{red}{\textcolor{white}{r}}
    $\mapsto$ \fcolorbox{red}{red}{\textcolor{white}{R}}   \\
   & \fcolorbox{red}{red}{\textcolor{white}{r}}
    $\mapsto$ \fcolorbox{blue}{blue}{\textcolor{white}{B}}   \\
\hline
\multirow{4}{*}{
\begin{tikzpicture}
    \draw (0.2, 0.2) rectangle (1, 1);
    \node[fill=blue, text=white] at (0.2, 1) {B};
    \node[fill=red, text=white] at (1, 1) {R}; 
    \node[fill=red, text=white] at (0.2, 0.2) {r};
    \node[fill=green, text=black] at (1, 0.2) {g};
    \node at (0.6, 0.6) {$\uparrow$};
\end{tikzpicture}
}   & \fcolorbox{green}{green}{\textcolor{black}{g}}
    $\mapsto$ \fcolorbox{red}{red}{\textcolor{white}{R}}   \\
   & \fcolorbox{green}{green}{\textcolor{black}{g}}
    $\mapsto$ \fcolorbox{blue}{blue}{\textcolor{white}{B}}    \\
   & \fcolorbox{red}{red}{\textcolor{white}{r}}
    $\mapsto$ \fcolorbox{red}{red}{\textcolor{white}{R}}   \\
   & \fcolorbox{red}{red}{\textcolor{white}{r}}
    $\mapsto$ \fcolorbox{blue}{blue}{\textcolor{white}{B}}   \\
\hline
\multirow{4}{*}{
\begin{tikzpicture}
    \draw (0.2, 0.2) rectangle (1, 1);
    \node[fill=red, text=white] at (0.2, 1) {r};
    \node[fill=red, text=white] at (1, 1) {R}; 
    \node[fill=blue, text=white] at (0.2, 0.2) {B};
    \node[fill=green, text=black] at (1, 0.2) {g};
    \node at (0.6, 0.6) {$\uparrow$};
\end{tikzpicture}
}   & \fcolorbox{green}{green}{\textcolor{black}{g}}
    $\mapsto$ \fcolorbox{red}{red}{\textcolor{white}{R}}   \\
   & \fcolorbox{green}{green}{\textcolor{black}{g}}
    $\mapsto$ \fcolorbox{blue}{blue}{\textcolor{white}{B}}    \\
   & \fcolorbox{red}{red}{\textcolor{white}{r}}
    $\mapsto$ \fcolorbox{red}{red}{\textcolor{white}{R}}   \\
   & \fcolorbox{red}{red}{\textcolor{white}{r}}
    $\mapsto$ \fcolorbox{blue}{blue}{\textcolor{white}{B}}   \\
\hline
\multirow{4}{*}{
\begin{tikzpicture}
    \draw (0.2, 0.2) rectangle (1, 1);
    \node[fill=green, text=black] at (0.2, 1) {g};
    \node[fill=red, text=white] at (1, 1) {R}; 
    \node[fill=blue, text=white] at (0.2, 0.2) {b};
    \node[fill=blue, text=white] at (1, 0.2) {B};
    \node at (0.6, 0.6) {$\uparrow$};
\end{tikzpicture}
}   & \fcolorbox{green}{green}{\textcolor{black}{g}}
    $\mapsto$ \fcolorbox{red}{red}{\textcolor{white}{R}}   \\
   & \fcolorbox{green}{green}{\textcolor{black}{g}}
    $\mapsto$ \fcolorbox{blue}{blue}{\textcolor{white}{B}}    \\
   & \fcolorbox{blue}{blue}{\textcolor{white}{b}}
    $\mapsto$ \fcolorbox{red}{red}{\textcolor{white}{R}}   \\
   & \fcolorbox{blue}{blue}{\textcolor{white}{b}}
    $\mapsto$ \fcolorbox{blue}{blue}{\textcolor{white}{B}}   \\
\hline
\multirow{4}{*}{
\begin{tikzpicture}
    \draw (0.2, 0.2) rectangle (1, 1);
    \node[fill=yellow, text=black] at (0.2, 1) {y};
    \node[fill=red, text=white] at (1, 1) {r}; 
    \node[fill=red, text=white] at (0.2, 0.2) {R};
    \node[fill=blue, text=white] at (1, 0.2) {B};
    \node at (0.6, 0.6) {$\uparrow$};
\end{tikzpicture}
}   & \fcolorbox{yellow}{yellow}{\textcolor{black}{y}}
    $\mapsto$ \fcolorbox{red}{red}{\textcolor{white}{R}}   \\
   &  \fcolorbox{yellow}{yellow}{\textcolor{black}{y}}
    $\mapsto$ \fcolorbox{blue}{blue}{\textcolor{white}{B}}    \\
   & \fcolorbox{red}{red}{\textcolor{white}{r}}
    $\mapsto$ \fcolorbox{red}{red}{\textcolor{white}{R}}   \\
   & \fcolorbox{red}{red}{\textcolor{white}{r}}
    $\mapsto$ \fcolorbox{blue}{blue}{\textcolor{white}{B}}   \\
\hline
\end{tabularx}
\label{tab:experiment_sheet}
\end{table}

\begin{enumerate}
    \item For each method, we run 20 episodes with different initial scene states;
    \item We apply a time limit of 90 seconds for each evaluation episode. 
    \item To ensure the repeatability of the scene setup, we use the \texttt{Standard} spatial layout as shown in \Figure~\ref{fig:spatial_layout}, as it is easy to ensure accuracy of the scene setup in terms of spatial positions. In practice, 
    we mark the positions for the robot at the center of the square.
    Similarly, we mark the 4 corner positions on the lobby floor for placing the objects and baskets.
    The size of the square is about $2 \times 2$ meters.
    \item We follow a scene initialization procedure as detailed in Table~\ref{tab:experiment_sheet} across the 20 episodes to ensure good coverage of the positions of the objects, the target object and basket in the language instruction, and the colors of the objects.
\end{enumerate}

\section{Visualizations}\label{app:visualization}

\begin{figure*}[t]
    \centering
    \resizebox{\textwidth}{!}{
    \begin{overpic}[width=18cm]{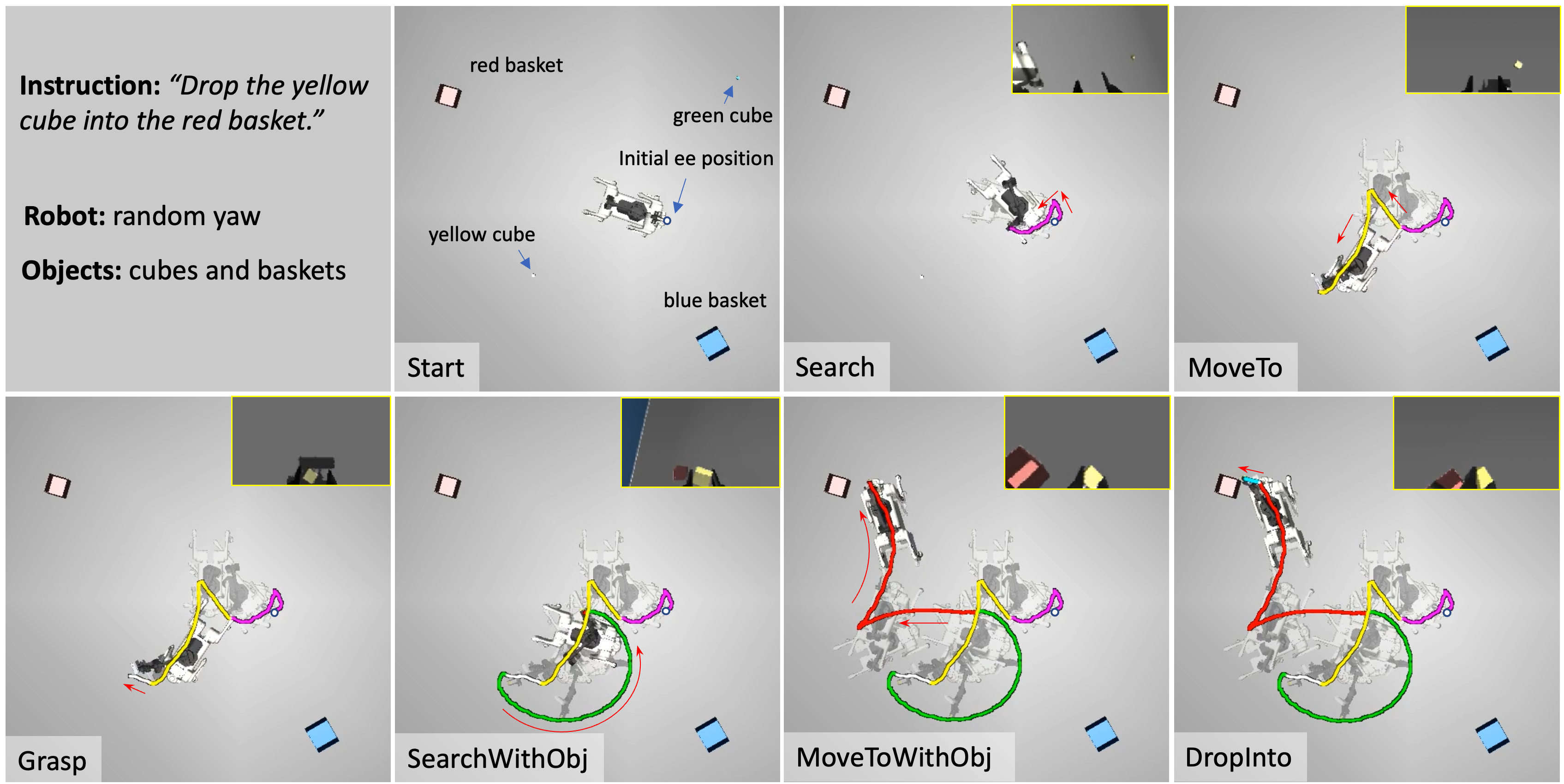}
    \end{overpic}
    }
    \vspace{-0.05in}
    \caption{\textbf{Policy Trajectory Visualization.} For visualization purpose, we removed background objects. We also visualize the EE trajectories. Trajectories corresponding to different subtasks are rendered with different colors. The ego-centric camera view is shown on the top-right of each image. The red colored arrows denote the direction of movement.}
    \label{fig:policy_visualization}
\end{figure*}

\begin{figure}[t]
    \centering
    \resizebox{\columnwidth}{!}{
    \begin{overpic}[width=10cm]{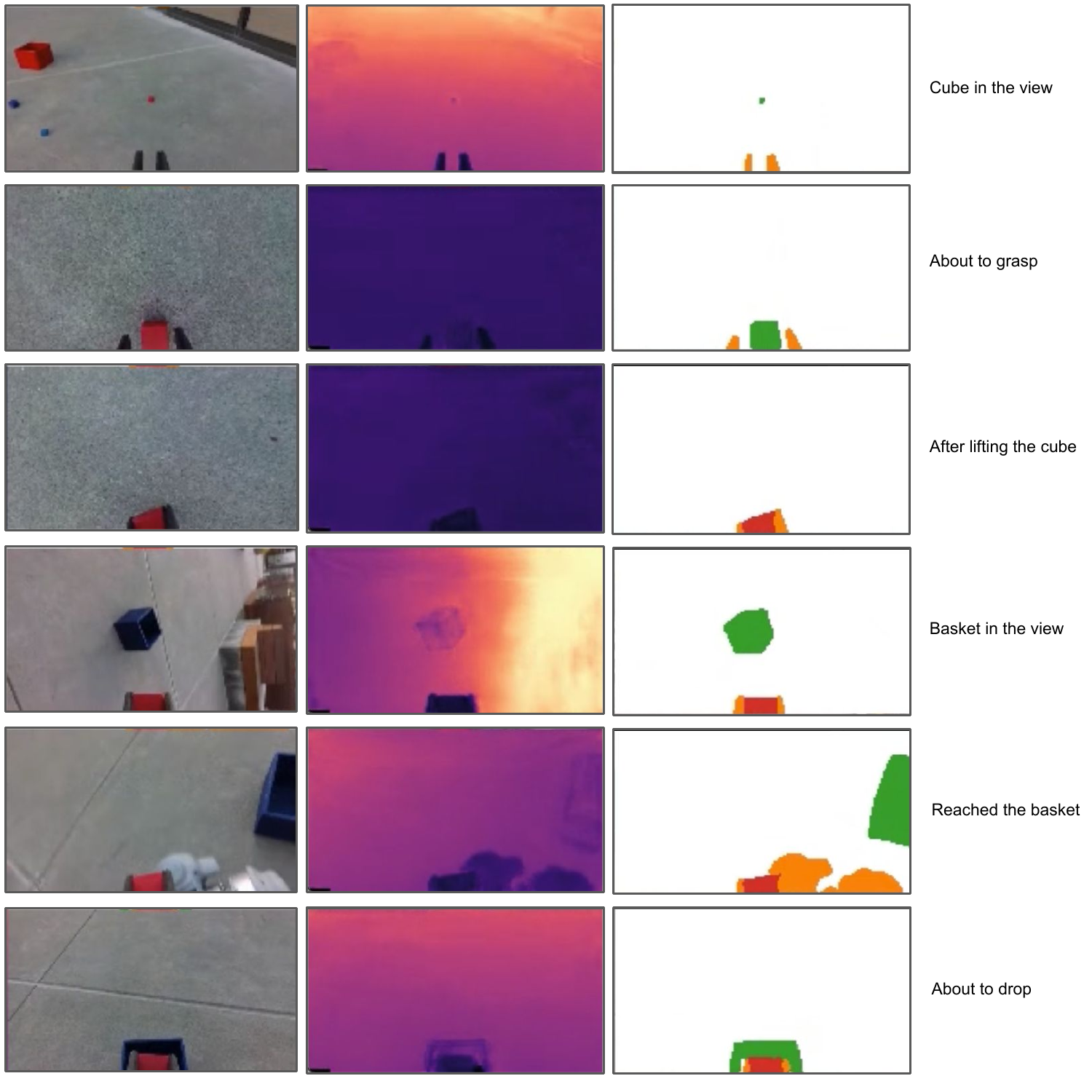}
    \end{overpic}
    }
    \vspace{-0.1in}
    \caption{\textbf{Maps prediction results for some key frames of a deployment session.}
    Time flows from top to bottom, and each row corresponds to one time step.
    From left to right, the columns represent the RGB frame, the predicted depth map, the predicted segmentation mask, and our added note for better interpretation of different stages.
    }    
    \vspace{-0.1in}
    \label{fig:maps_prediction_result}
\end{figure}

\subsection{Visualization of Policy Behavior}\label{app:policy_behavior}
A visualization of the policy behaviors in real is shown in \Figure~\ref{fig:slim_in_real}. We also provide demo videos in the \demo{}.
Here we further visualize the policy behavior in simulation, across the complete long-horizon task with the end-effector (EE) trajectory rendered. To do this, we run the policy in the simulator, and render the images from a fixed top-down perspective and overlay the EE trajectory. Each EE trajectory is color-coded so that the segments corresponding to a different subtasks are colored differently.
The behavior of the policy is shown in \Figure~\ref{fig:policy_visualization}. 
To make the visualization clear we did not add background objects for this episode.
The robot starts with a randomized position and yaw within a region 
({\footnotesize\textsf{Start}} phase).
A red cube and a green cube as well as a red basket and a blue basket are used.
Their positions are randomized based on the \texttt{Standard} layout and their yaw angles are also randomized.
Given an instruction (\emph{``Drop the yellow cube into the red basket."}), the robot will first search for the target graspable object (the yellow cube).
Note that during the {\footnotesize\textsf{Search}} phase, the robot actively moves both its body and arm in search of the target object ({\Figure~\ref{fig:policy_visualization}-{\footnotesize\textsf{Search}}).
Once the target object is in view, the robot will move towards it  (\Figure~\ref{fig:policy_visualization}-{\footnotesize\textsf{MoveTo}}).
Once the robot moves close to the target object, it will start to grasp (\Figure~\ref{fig:policy_visualization}-{\footnotesize\textsf{Grasp}}).
During this stage, the robot body remains stationary and the arm will move towards the target cube and grasp it.
After picking up the target cube, the robot will search for the target basket while keeping the cube in gripper.
As shown in \Figure~\ref{fig:policy_visualization}-{\footnotesize\textsf{SearchWithObj}}, the robot rotates around its current position in search of the target basket (as indicated by the circular EE trajectory). At the same time, the arm retracts from the initial out-reaching pose after grasping towards a neutral arm pose, as indicated by the shrinking radius of the circular EE trajectory in \Figure~\ref{fig:policy_visualization}-{\footnotesize\textsf{SearchWithObj}}.
Once the target basket is in view, the robot will move towards it (\Figure~\ref{fig:policy_visualization}-{\footnotesize\textsf{MoveToWithObj}})
and drop the cube in gripper into that basket (\Figure~\ref{fig:policy_visualization}-{\footnotesize\textsf{DropInto}}).

\subsection{Visualization of Predicted Segmentation and Depth Maps}

In \Figure~\ref{fig:maps_prediction_result}, we show the robot's prediction results for depth and object segmentation, during an arbitrary outdoor deployment session.
We emphasize again that the robot was deployed in this zero-shot scenario that was never modeled by our simulator,
and no model finetuning was performed.
Despite this, we can see that the prediction results are of high quality, given only a very short history ($0.5$ seconds) of a single wrist-mounted RGB camera and the robot proprioceptive state.
While searching for the target cube, the perception model had to locate it with just a handful of pixels (first row).
Right after the gripper holding and lifting the cube above the ground, the model correctly predicted that the target had switched from the cube to a basket (cube's green color turned red; third row).
Moreover, while searching for the target basket in the outdoor environment, the cluttered background did not confuse the depth and segmentation prediction (fourth row).
The perception model was also able to correctly predict the segmentation mask for the robot itself (gripper fingers and quadruped in orange; fifth row).
Overall, we find that in practice, this high-quality visual information bottleneck not only greatly improves the generalization ability of our RL policy, but also facilitates our debugging of the entire robotic system as it separates the errors of the perception model from those of the policy.

\section{Human Teleoperation}\label{app:teleop}
We list failure modes of the \HumanTeleop{} baseline to give a more complete picture of how difficult our task is, and to show that \name{}'s performance is non-trivial.
\begin{enumerate}
    \item The most common failure mode for teleoperation is task timeout.  
    This is sometimes due to not fully grasping the cube and the cube slipping out of the fingers during movement, leaving no time for searching and grasping again.  
    Failing to grasp fully is in turn due to either stopping a bit far from the cube, or incorrectly assessing the distance to the cube under monocular vision. 
    \item The arm can have a twisted pose due to IK near singularities.
    This pose either overloads the arm motors, or if not, flips the camera upside down, making the operator difficult to perceive the environment.
\end{enumerate}
Overall, human teleoperation is less accurate and slower than an RL policy, due to the low-level policy's shaking, the difficulty of coordinating the movement of the end effector of the arm with locomotion, and sometimes reduced accuracy due to human fatigue despite given at least one break every half hour.
Besides being less time efficient, human teleoperation seems to be less energy efficient than our RL policies as well.  
While the RL trained high-level policy can run smoothly on a particular low-level policy, human teleoperation on the same low-level policy sometimes triggers over-current protection of Go1. 
This often happens when an over-extended arm is raised, causing Go1 to lose control of its leg motors.  
In evaluation, we ignored this Go1 hardware failure, and let the human operator simply try from start again.

\section{Long-Horizon Task Learning}
\subsection{Long-Horizon Task with Bottleneck States}\label{app:bottlebeck_states}
The challenges of long-horizon task for exploration and learning have been discussed in Section~\ref{sec:teacher_learning}.
Here we want to further highlight that the existence of bottleneck states, which is orthogonal to long horizon, further exacerbates the difficulties.

\begin{figure}[h]
    \centering
    \begin{overpic}[width=\linewidth]{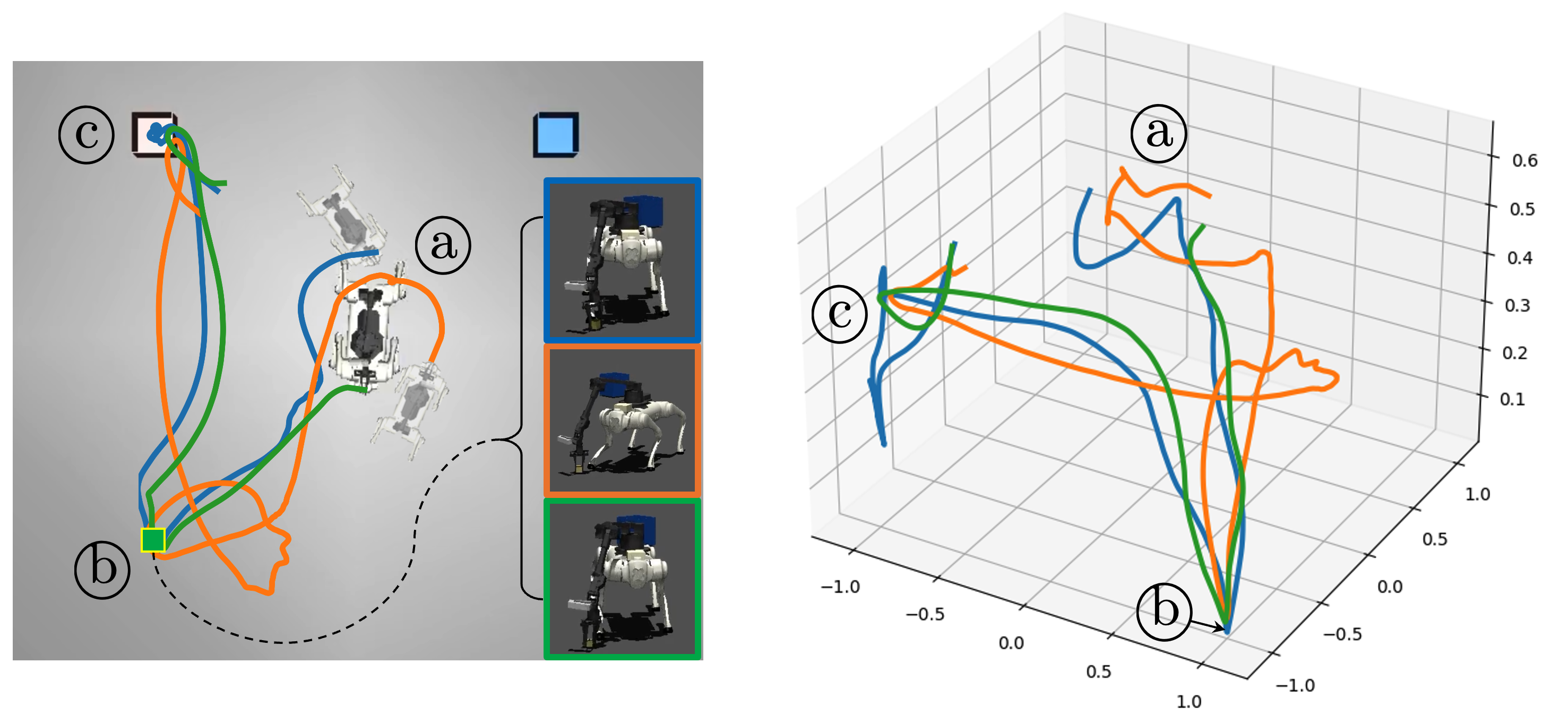}
    \end{overpic}
    \vspace{-0.2in}
    \caption{\textbf{Long-Horizon Task with Bottleneck States.}
    Left: Top-down view of the end effector trajectories. Right: The same set of trajectories in 3D space.
    {\circled{a}-\circled{c} denotes different stages of task execution:
	\circled{a}~beginning \circled{b}~grasping \circled{c}~task completion.}
    }
    \vspace{-0.1in}
    \label{fig:bottleneck_states}
\end{figure}

A bottleneck state is a milestone that must be accomplished before the robot reaching the final task success.
Therefore, given two tasks with roughly the same horizon length, the one with the presence of bottleneck states will typically be more difficult to learn than the one without.

To visualize the bottleneck states for the tasks considered in this work, we use spatial proximity of a key position (end effector / EE position) as a proxy of state proximity. Therefore, positions that are spatially close to each other can be roughly interpreted as with similar states in terms of accomplishing the mobile manipulation task. 
Under this setting, for the same task, if there is a common segment along the task progress with a compact spatial support, it can be regarded as a bottleneck state.

We visualize the trajectories from three episodes in \Figure~\ref{fig:bottleneck_states},
given a fixed object layout and different starting positions.
As we can see from the figure, although with quite different starting positions,
all the trajectories converge and intersect at the same location marked by \raisebox{.5pt}{\textcircled{\raisebox{-.9pt} {b}}}  in the figure, corresponding to the state for grasping the target object. 

To achieve the eventual full task success (dropping the target object into basket), a successful grasping of the target object has to be accomplished as a prerequisite.
Given that grasping small objects requires high precision control (as shown by the inserts in the left part of \Figure~\ref{fig:bottleneck_states}), 
it is a clear bottleneck state, increasing the difficulties of exploration and learning, especially for the part of state space that is only available after passing the bottleneck state. 

The existence of bottleneck states coupled with long task horizon poses a great challenge to a standard algorithm without progressive policy expansion, as shown by the results in the subsequent section.

\begin{figure}[h]
    \centering
    \includegraphics[width=1\columnwidth]{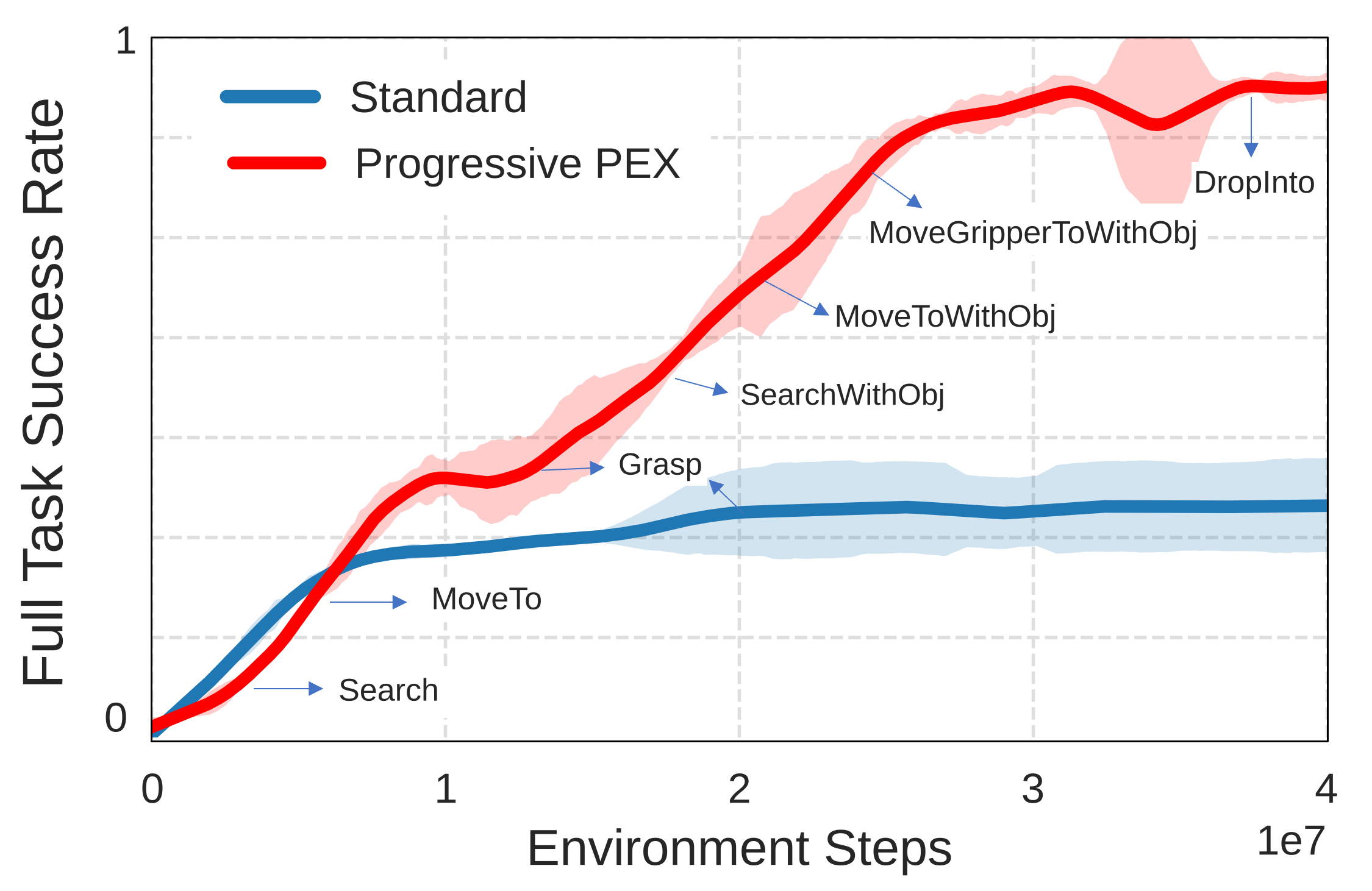}
     \vspace{-0.1in}
    \caption{\textbf{Standard Method v.s. Progressive PEX}. We compare the learning behaviors of the Standard method (without progressive policy expansion) and the Progressive PEX approach in terms of the full task success rate as training proceeds (the mean and standard deviation calculated across 3 seeds).
    We also provide rough annotations of the subtask stages that are under learning along the curve.
    }
    \label{fig:pex_vs_standard}
\end{figure}

\subsection{The Effectiveness of Progressive Policy Expansion}
\label{app:pex_results}
Section~\ref{sec:teacher_learning} introduces the Progressive Policy Expansion approach ({\footnotesize\textsf{Progressive PEX}}) for long-horizon task learning.
We compare the performance of {\footnotesize\textsf{Progressive PEX}} and that of the standard approach without progressive policy expansion ({\footnotesize\textsf{Standard}}) in \Figure~\ref{fig:pex_vs_standard}.
We show each policy's ability in solving the long-horizon task throughout the training process.

As can be observed from the figure, {\footnotesize\textsf{Standard}} approach starts with a similar trend of learning to solve the initial subtasks (\emph{e.g.}, {\footnotesize\textsf{Search}}, {\footnotesize\textsf{MoveTo}}) as {\footnotesize\textsf{Progressive PEX}}.
However, after that, {\footnotesize\textsf{Standard}} appears saturated in terms of the learning progress. More specifically, 
it cannot fully learn the  {\footnotesize\textsf{Grasp}} subtask and is not able to learn further beyond that, likely due to the inability of continual exploration and loss of plasticity~\citep{capacity_loss, understanding_plasticity}, especially in the presence of a long-horizon task with bottleneck states (Appendix~\ref{app:bottlebeck_states}).

In contrast, {\footnotesize\textsf{Progressive PEX}} can keep pushing the frontier on the long-horizon task progress, learning to solve the subsequent new subtasks that appear along the way towards solving the full task, as shown in \Figure~\ref{fig:pex_vs_standard}.

\end{document}